\pdfoutput=1
\documentclass[10pt,journal,compsoc]{IEEEtran}

\ifCLASSOPTIONcompsoc
  \usepackage[nocompress]{cite}
\else
  \usepackage{cite}
\fi
\ifCLASSINFOpdf
\else
\fi
\usepackage{hyperref}       
\usepackage{url}            
\usepackage{amsfonts}       
\usepackage{microtype}      
\usepackage{xcolor}         
\usepackage{algorithm}      
\usepackage{algorithmic}
\usepackage{amsmath}
\usepackage{xcolor}
\usepackage{graphicx}

\usepackage{multirow}
\usepackage{multicol}
\usepackage{bbding}         
\usepackage{makecell}
\usepackage{tabularx}       

\usepackage{amssymb}
\usepackage{pifont}
\newcommand{\cmark}{\ding{51}}%
\newcommand{\xmark}{\ding{55}}%

\begin{document}
%
\title{Uni-MoE: Scaling Unified Multimodal LLMs with Mixture of Experts}
%
%
%
%

\author{Yunxin Li, Shenyuan Jiang, Baotian Hu, Longyue Wang, Wanqi Zhong, Wenhan Luo, Lin Ma, Min Zhang
\IEEEcompsocitemizethanks{
\IEEEcompsocthanksitem Yunxin Li, Shenyuan Jiang, Baotian Hu, Wanqi Zhong, and Min Zhang are with the Department of Computer Science and Technology, Harbin Institute of Technology, Shenzhen, China. (e-mail: liyunxin987@163.com, hubaotian@hit.edu.cn, and zhangmin2021@hit.edu.cn)
\IEEEcompsocthanksitem Wenhan Luo is an Associate Professor at the Hong Kong University of Science and Technology, Hong Kong. (e-mail: whluo.china@gmail.com)
\IEEEcompsocthanksitem Lin Ma is a Researcher with Meituan, Beijing,
China. (e-mail: forest.linma@gmail.com)
\IEEEcompsocthanksitem Baotian Hu and Longyue Wang are the corresponding authors. (e-mail: hubaotian@hit.edu.cn and vincentwang0229@gmail.com.
\IEEEcompsocthanksitem Project Website: \url{https://uni-moe.github.io/}
}}

%
%

\markboth{Journal of \LaTeX\ Class Files,~Vol.~14, No.~8, August~2015}%
{Shell \MakeLowercase{\textit{et al.}}: Bare Advanced Demo of IEEEtran.cls for IEEE Computer Society Journals}
%



\IEEEtitleabstractindextext{%
\begin{abstract}
Recent advancements in Multimodal Large Language Models (MLLMs) underscore the significance of scalable models and data to boost performance, yet this often incurs substantial computational costs. Although the Mixture of Experts (MoE) architecture has been employed to efficiently scale large language and image-text models, these efforts typically involve fewer experts and limited modalities. To address this, our work presents the pioneering attempt to develop a unified MLLM with the MoE architecture, named \textbf{Uni-MoE} that can handle a wide array of modalities. Specifically, it features modality-specific encoders with connectors for a unified multimodal representation. We also implement a sparse MoE architecture within the LLMs to enable efficient training and inference through modality-level data parallelism and expert-level model parallelism. To enhance the multi-expert collaboration and generalization, we present a progressive training strategy: 1) Cross-modality alignment using various connectors with different cross-modality data, 2) Training modality-specific experts with cross-modality instruction data to activate experts' preferences, and 3) Tuning the Uni-MoE framework utilizing Low-Rank Adaptation (LoRA) on mixed multimodal instruction data. We evaluate the instruction-tuned Uni-MoE on a comprehensive set of multimodal datasets. The extensive experimental results demonstrate Uni-MoE's principal advantage of significantly reducing performance bias in handling mixed multimodal datasets, alongside improved multi-expert collaboration and generalization. Our findings highlight the substantial potential of MoE frameworks in advancing MLLMs and the code is available at \url{https://github.com/HITsz-TMG/UMOE-Scaling-Unified-Multimodal-LLMs}.
\end{abstract}


}

\maketitle

\IEEEdisplaynontitleabstractindextext

%
\IEEEpeerreviewmaketitle

\ifCLASSOPTIONcompsoc
\IEEEraisesectionheading{\section{Introduction}\label{sec:introduction}}
\else
\section{Introduction}
\label{sec:introduction}
\fi

%
%
%
%

\IEEEPARstart{R}{ecent} advancements in open-source Multimodal Large Language Models (MLLMs)~\cite{MultimodalLearning} such as InstructBLIP~\cite{dai2024instructblip} and LLaVA~\cite{liu2023improved} present notable successes in image-text understanding tasks ~\cite{seedBENCH,mmbench}. Additionally, there is a growing trend \cite{panagopoulou2023x,lyu2023macaw,moon2023anymal,li2023llamavid} toward building a unified MLLM that could comprehend more modalities such as video, audio, and speech, moving beyond the traditional image-text paradigm. To catch up with superior closed-source MLLMs like GPT-4V~\cite{gpt4} and Gemini~\cite{team2023gemini}, the main efforts of open-source community contain enlarging model sizes~\cite{chen2023internvl}, as seen with the expansion of vision foundation models to 6 billion parameters \cite{chen2023internvl} and the integration with 70B Large Language models (LLMs) \cite{laurenccon2024obelics,touvron2023llama}, and enhancing instruction tuning with diverse multimodal datasets \cite{liu2023improved,li2023lmeye,li2023comprehensive_gpt4v}.
These developments underscore the increasing ability of MLLMs to process and reason across multiple modalities, showing the importance of both model scalability and the expansion of multimodal instructional data.
However, scaling up model size usually incurs huge computational overhead in both the training and inference phases.

\begin{figure}[t]
    \centering
    \includegraphics[width=0.48\textwidth]{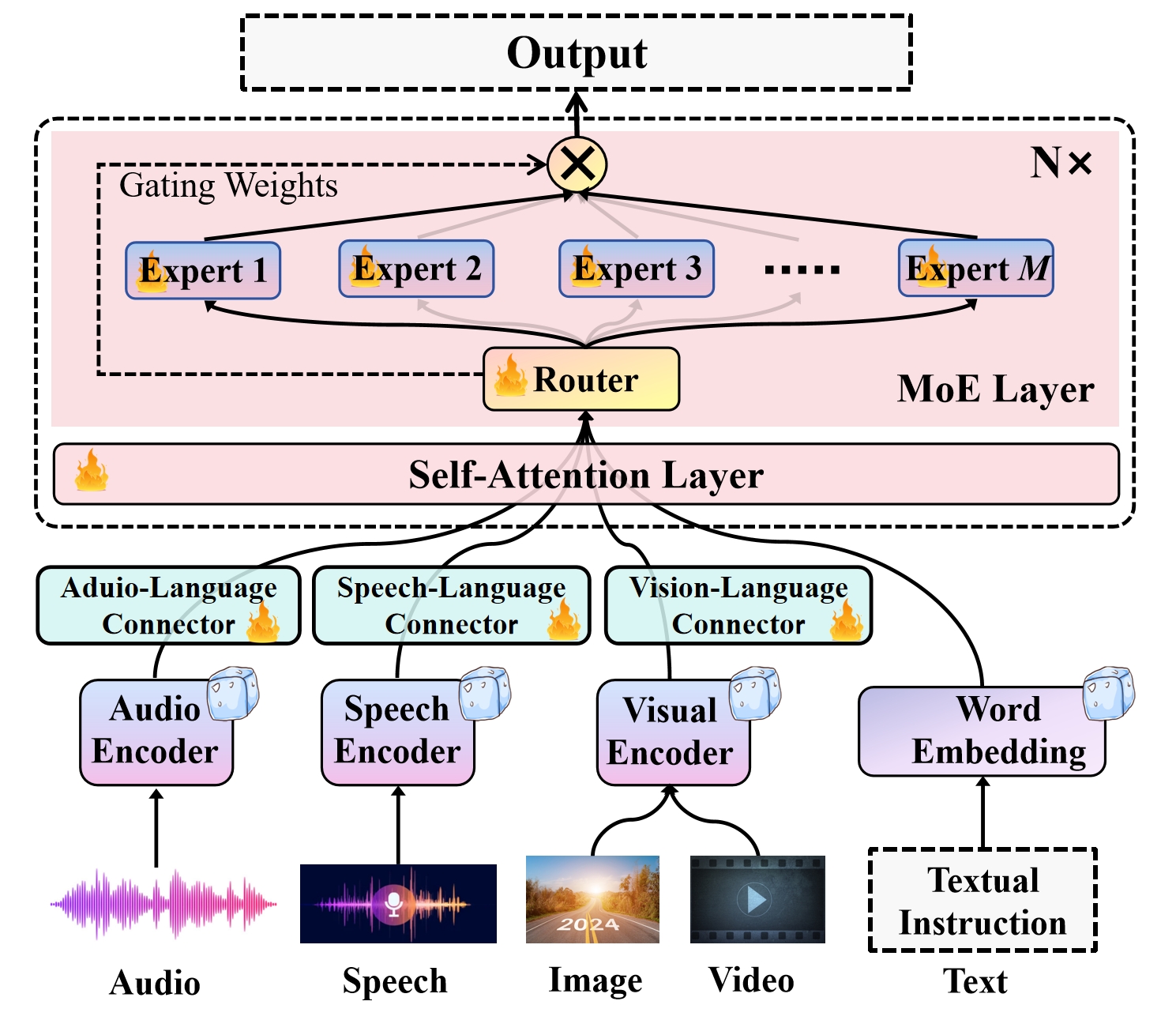}
    \caption{\textbf{An Illustration of Uni-MoE}. Compared to previous dense MLLMs, it employs the MoE architecture to build a unified MLLM that can handle various modalities. We use the sparse routing control with the lightweight finetuning method LoRA to activate different experts, aiming to reduce computational costs.}
    \label{fig:intro_case}
\end{figure}

To alleviate this issue, there has been a shift towards integrating the Mixture of Experts (MoE)~\cite{shazeer2017outrageously,6215056} architecture in large models to improve training and inference efficiency. Unlike conventional MLLMs or LLMs, where each input is processed by all model parameters, resulting in a dense computational approach, the MoE architecture only requires activating a subset of expert parameters for each input, as determined by an expert selector or router.
Consequently, the MoE approach emerges as a promising strategy for improving the efficiency of large models while reducing the need for extensive parameter activation.
For instance, Jiang \textit{et al.} \cite{jiang2024mixtral} introduces a sparse MoE-based language model named Mixtral-MoE 8x7B with each layer composed of 8 experts. It outperforms Llama2-70B in mathematics, code generation, and multilingual benchmarks while requiring less activated parameters. Lin \textit{et al.} \cite{moeLlava} developed an MoE-enhanced image-text MLLM, MoE-LLaVA, which utilizes about 3 billion activated parameters yet achieves comparable results to the dense 7B model on a variety of image-text understanding benchmarks.

While previous works have demonstrated the successful application of MoE in the construction of text-only and image-text large models, developing the MoE architecture to construct powerful unified MLLMs remains largely uncharted, e.g., scaling MLLMs to incorporate more than four experts and extending their application to modalities beyond just images and text.
In this work, we pioneer the exploration of scaling unified MLLMs with the MoE architecture and present an efficient MLLM named \textbf{Uni-MoE} that can leverage sparse MoE to adeptly manage and interpret multiple modalities. Specifically, as depicted in Figure~\ref{fig:intro_case}, we first use modality-specific encoders to obtain the encoding of different modalities and map them into the language representation space of LLMs by the designed various connectors. They contain a trainable transformer model with subsequent linear projection layers to distil and project the output representations of frozen encoders, respectively. Then, we introduce a sparse MoE layer within the internal blocks of dense LLM. Hence, each MoE-based block features a shared self-attention layer applicable across all modalities, diverse experts based on feed-forward networks (FFN), and a sparse router for allocating token-level expertise. In this way, Uni-MoE can understand multiple modalities such as audio, speech, image, video and text, and only need activate partial parameters during inference.

Additionally, to enhance the multi-expert collaboration and generalization of Uni-MoE, we develop a three-stage progressive training approach: Firstly, we use extensive image/speech/audio-to-language pairs to train the corresponding connector, respectively, realizing the unified modality representation in the language space of LLMs. Secondly, we separately train modality-specific experts employing cross-modality datasets to refine each expert’s proficiency within its respective domain. Thirdly, we integrate these trained experts into the MoE layer of the LLM and train the whole Uni-MoE framework with mixed multimodal instructional data. To further reduce the training cost, we employ the LoRA~\cite{lora} technique to fine-tune these pre-tuned experts and the self-attention layers. Through the above three-phase training approach, we gain an efficient and stable Uni-MoE that can adeptly manage and interpret multiple modalities. 


To verify the effectiveness of Uni-MoE, we compare it with various dense MLLMs across extensive benchmarks such as image-text, video, and audio/speech understanding datasets. Additionally, we also introduce a new benchmark named the English High School Listening Test to assess the model's performance in complex long speech understanding scenarios. The experimental results suggest that leveraging the MoE architecture in constructing multimodal models not only outperforms traditional dense model setups in demanding multimodal contexts such as videos and long speech but also enhances stability and robustness across different modalities. In addition, we also discovered that the integration of more modality information into Uni-MoE enhances the performance of single-modality tasks. For instance, incorporating more image-text data improves the output of video QA tasks.
Our contributions are as follows:

1) \textit{Framework.} We present Uni-MoE (Sec \ref{unimoe_architecture} and Table~\ref{architecture_moe}), the pioneering sparse MoE-based unified MLLM that integrates multiple modalities including video, images, text, audio, and speech. It is constructed using modality-specific encoders, other modality-to-language connectors, and an LLM equipped with the sparse MoE architecture. During training and inference, we enhance the scalability of the MoE architecture training with an increased number of experts and a diverse set of multimodal data, through the expert-level model parallel and modality-level data parallel\footnote{We have released the two distributed parallel training approaches.}.

2) \textit{Training Strategy}. We introduce a progressive training paradigm (Sec \ref{training_strategy} and Algorithm~\ref{algorithms_1}): an alignment stage for different modalities to language, training modality-specific experts, and undergoing unified MoE training with LoRA on mixed multimodal data. Our detailed experiments (Table~\ref{ablation_training}) reveal that pre-training experts on individual modalities significantly enhances the collaboration and generalization capabilities of multi-expert systems, surpassing the outcomes of standard MoE tuning where each expert possesses the same initial parameters.

3) \textit{Practice}. Uni-MoE consistently outperforms dense MLLMs on almost all evaluation benchmarks, showing its advantages of exceptional ability in handling complex out-domain tasks. We investigate the role of widely-used auxiliary balancing loss~\cite{fedus2022switch} in training the Sparse MoE-based MLLM with mixed modal data (Tables \ref{ablation_training} and \ref{more-experts-und}), finding that even without this auxiliary loss, Uni-MoE demonstrates superior multi-expert collaboration and generalization. Our results further reveal that as the number of experts and the routing search space expand, the benefits of auxiliary loss become pronounced.

\section{Related Work}

\begin{figure*}[t]
    \centering
    \includegraphics[width=0.98\textwidth]{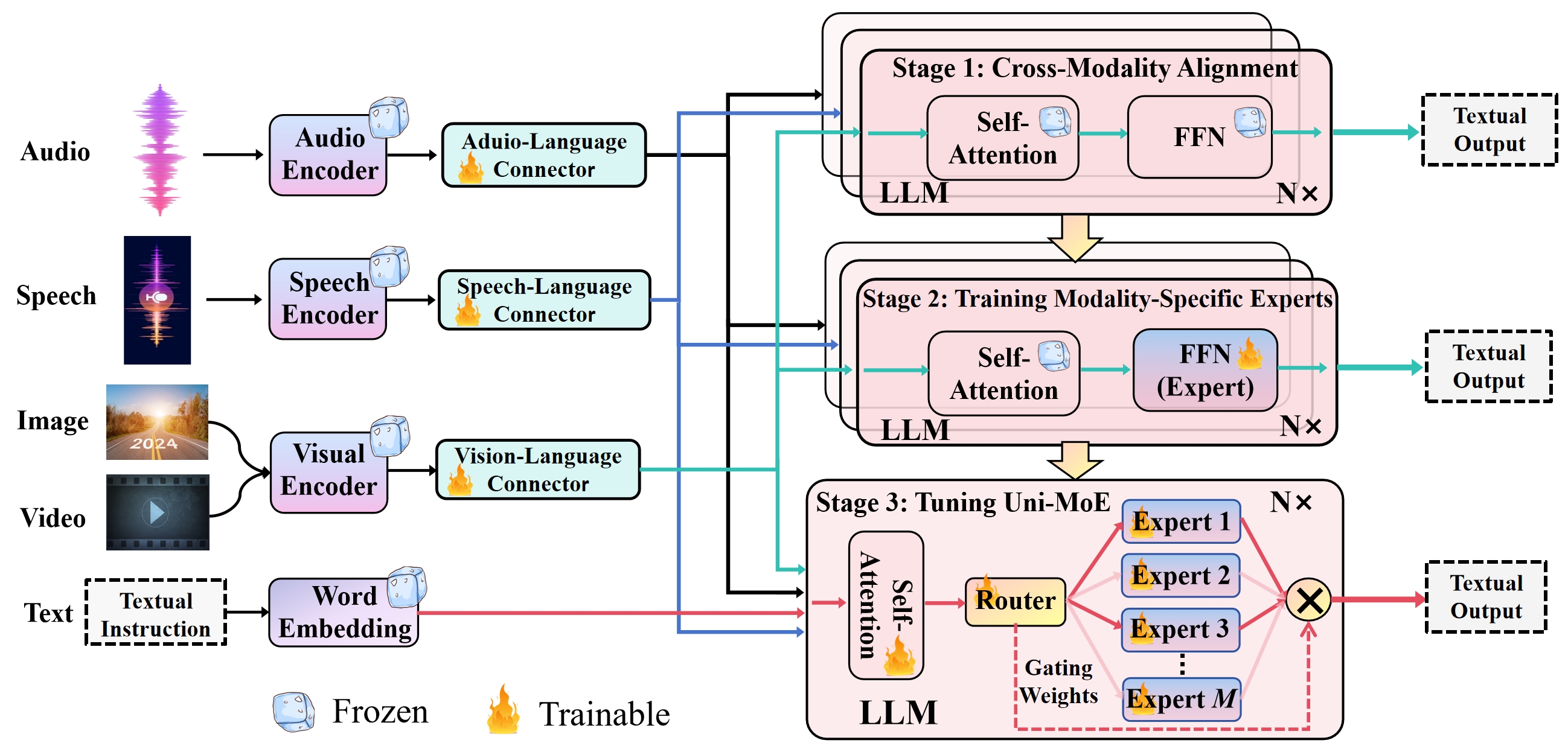}
    \caption{\textbf{Overview of Uni-MoE Training Methodology}. The progressive training stages contain: 1) Utilize pairs from different modalities and languages to train connectors that map these elements to a unified language space, establishing a foundation for multimodal understanding; 2) Develop modality-specific experts using cross-modal data to ensure deep understanding, preparing for a cohesive multi-expert model; 3) Incorporate multiple trained experts into LLMs and refine the unified multimodal model using the LoRA technique on mixed multimodal data.}
    \label{fig:model}
\end{figure*}

\subsection{Unified Multimodal Models}
The advent of the Transformer model, introduced by Vaswani \textit{et al.} \cite{vaswani2017attention}, marked a significant milestone in the field of deep learning, enabling the scalable integration of multiple modalities—including image, language, speech, audio, and video—into a unified representational space. This cross-modal interactive parallelism facilitates a range of generative and interpretive capabilities across different data types after training on extensive multimodal datasets. Recent advancements in generative multimodal large models have further expanded these capabilities~\cite{visionTrasfor,learningOpenVocab}, allowing for the perception and generation of diverse modalities~\cite{wu2023next,zhang2023meta,panagopoulou2023x,zhan2024anygpt}.
A notable example of this progress is ImageBind by Girdhar \textit{et al.} \cite{girdhar2023imagebind}, which pioneers in learning a joint embedding across six distinct modalities, thus opening the door to emergent applications such as cross-modal retrieval, modality composition through arithmetic operations, cross-modal detection, and generation, directly "out-of-the-box." Similarly,  Kirillov \textit{et al.} \cite{kirillov2023segment} introduce SAM, a foundational model for image segmentation capable of integrating insights from textual descriptions, visual cues, and semantic mappings, thereby demonstrating the versatility and robustness of multi-modal approaches in handling complex tasks.
In a parallel vein, Zhang \textit{et al.} \cite{zhang2023meta} put forward the Meta-Transformer model, characterized by a unified data Tokenizer, a shared encoder across modalities, and specialized heads for task-specific applications. This model stands out for its ability to conduct unified learning across an impressive array of 12 modalities using unpaired data, showcasing the potential for broad applicability across various domains and tasks~\cite{graphfusion,VisionLanguage}.
Building on these foundational works, our study aims to construct an efficient unified MLLM leveraging the strengths of large language models and dedicated modality connectors. 

\subsection{Multimodal Instruction Tuning for LLMs}
The recent surge in Natural Language Processing (NLP) research owes much to the evolution of Large Language Models (LLMs) such as Flan-T5 \cite{chung2022scaling}, GPT-4~\cite{gpt4} and LLAMA~\cite{touvron2023llama}. These developments have not only enhanced our understanding and capabilities within the realm of language processing but have also paved the way for the advent of MLLMs. The progression towards these advanced models has been facilitated by two key factors: the broadening of training datasets~\cite{liu2023improved} and significant improvements in model architecture and design, as highlighted by recent works~\cite{laurenccon2024obelics,moon2023anymal}.
A prominent area of research focuses on leveraging pre-trained LLMs to improve visual instruction tuning, a process that involves refining the models to better understand and execute visually grounded instructions. This field has seen the emergence of notable models like LLaVA~\cite{zhang2023llavar}, MiniGPT-4~\cite{zhu2023minigpt}, InstructBLIP~\cite{dai2024instructblip}, Qwen-VL~\cite{bai2023qwen}, and LMEye~\cite{li2023lmeye}. A common framework among these models involves integrating a pre-trained visual backbone~\cite{wu2023visual} for image and video interpretation, an LLM~\cite{gao2023llama} for comprehending textual instructions and generating appropriate responses, and a vision-language cross-modal connector. This connector plays a critical role in harmonizing the capabilities of the vision encoder with the language model, ensuring seamless interaction between the two modalities.
Efficient training paradigms~\cite{efficientlearning} with data management~\cite{dataManagement} also play an important role in constructing large models. It not only enhances the model's ability to process and understand multimodal inputs but also significantly improves its applicability across a wide range of tasks that require nuanced understanding and generation of responses based on both textual and visual cues. 

\subsection{Large Models with Mixture of Experts}
Due to the huge overhead caused by training and deploying large models, researchers have been committed to exploring utilizing mixture of experts (MoE)~\cite{eigen2013learning} architecture to improve the efficiency of large models. This is particularly evident in large models where MoE enables selective activation of different network components, optimizing computational resources and enhancing performance. Pioneering work in this space can be seen with models such as GShard~\cite{lepikhin2020gshard}, which introduced the use of MoE for massively multilingual language models, demonstrating considerable improvements in translation tasks. Switch Transformers~\cite{fedus2022switch} further refined this concept by scaling up the MoE approach, achieving unprecedented model sizes and training efficiency on visual understanding. Additionally, BASE Layers~\cite{lewis2021base} explored the integration of MoE into bidirectional encoder representations from transformers, further highlighting the potential for MoE to enhance language understanding tasks. The application of MoE in generative models has also been explored, with notable examples such as GLaM~\cite{du2022glam}, showcasing the capability of MoE to handle diverse and complex generative tasks. In this paper, we mainly explore constructing a unifying large multimodal model using the MoE architecture.
This body of work collectively underscores the versatility and efficiency gains MoE brings to the realm of large-scale model architectures, pointing towards a future where model size and performance can be enhanced without proportional increases in computational demands.

\section{Uni-MoE}

\subsection{Overview}
Motivated by the high training and inference costs brought by scaling multimodal large models towards GPT-4V and Gemini and the efficiency of the MoE structure, here, we explore the realization of an efficient and powerful unified MLLM, utilizing the MoE architecture.
Figure~\ref{fig:model} presents a schematic representation of the designed Uni-MoE, showcasing its comprehensive design that includes encoders for audio, speech, and visuals, along with respective modality connectors. These connectors serve to translate various modality inputs into a unified language space. We then integrate the MoE architecture within the core LLM blocks, which is crucial for boosting the efficiency of both training and inference processes owing to only activating partial parameters. This is achieved through the implementation of a sparse routing mechanism, which is shown in the bottom part of Figure~\ref{fig:model}. The whole training process of Uni-MoE is divided into three distinct phases: cross-modality alignment, training modality-specific experts, and tuning Uni-MoE using a diverse set of multimodal instruction datasets. In the ensuing subsections, we will delve into the intricate architecture and the progressive training methodology of Uni-MoE in detail.

\subsection{Architecture}
\label{unimoe_architecture}
\textbf{Connectors}. 
To facilitate the efficient transformation of diverse modal inputs into a linguistic format, our Uni-MoE model is built upon the pretrained visual-language framework LLaVA~\cite{liu2023llava}. This base model integrates the CLIP~\cite{clip} as the visual encoder, alongside a linear projection layer that converts image features into corresponding soft image tokens within the language domain of Vicuna-LLaMA~\cite{vicuna2023}. For processing video content, we select eight representative frames from each video and transform them into video tokens by employing average pooling to aggregate their frame-based (image) representations. 
In the audio domain, we enhance feature extraction through the deployment of two distinct encoders: the Whisper encoder, derived from the Whisper-small speech recognition model~\cite{whisper}, and the BEATs encoder~\cite{beats}, a sophisticated audio processing tool that generates bidirectional encoder representations from audio transformers. Following a strategy akin to the Q-former approach~\cite{li2023blip2}, we then distil fixed-length speech and audio feature vectors respectively and subsequently map them into soft audio and speech tokens via a linear projection layer. The specific workflow is given in:
    \begin{align}
         X_{Input} &= [I, V, A, S, T],\vspace{2.0pt}\\
         I &= \text{MLP}(\text{CLIP-V}(I)), \vspace{2.0pt} \\
         V &= \text{Mean}(\text{CLIP-V}([ I_1,..., I_8])), \vspace{2.0pt} \\
         A &= \text{Audio-Qformer}(\text{BEATs}(A)), \vspace{2.0pt} \\
         S &= \text{Speech-Qformer}(\text{Whisper}(S)), \vspace{2.0pt} \\
         T &= \text{Word-Embedding}(T),
    \end{align}
where $[I, V, A, S, T]$ represents the image, video, audio, speech, and Text, respectively.
``MLP'' is the learnable visual-language projection layer. \text{Audio/Speech-Qformer} is a four-layer transformer block, where we feed learnable fixed-length query vectors into them and the corresponding audio/speech hidden states as the key and value in the cross-attention layer. By doing such a four-layer calculation, the final outputs can be regarded as the representations of audio and speech inputs. 
Take the Audio-QFormer as an example, the detailed calculation progress is presented as the following equation:

\begin{align}
      X_Q^{A} &= (h_Q^{1}, ..., h_Q^{AM}),\vspace{2.0pt} \\
      h_B^{A} &= \text{BEATs}(A),\vspace{2.0pt} \\
      h_S^{A} &= \text{MSA}(\text{LN}(X_Q^{A})) + X_Q^{A}, \vspace{2.0pt} \\
      h_C^{A} &= \text{MCA}(\text{LN}(h_S^{A}), h_B^{A}) + h_S^{A}, \vspace{2.0pt} \\
      h_1^{A} &= \text{MLP}(h_C^{A}),
\end{align}

where $h_B^{A}$ is the top output of pretrained audio encoder BEATs and $h_S^{A}$ is the multi-head self-attention calculation for the fixed-length query vectors $X_Q^{A}$, where $AM$ refers to the total number of initial vectors. $h_C^{A}$ represents the output of the cross-attention module, which is used to distil the main content of input audio. After four layers of the same operation, we apply a learnable linear layer for projecting the last output into the representation space of LLM.

\noindent\textbf{Uni-MoE}. By the above connectors, we could obtain the encoding tokens of any modality. For any modality inputs, we concatenate the corresponding tokens into one sequence and feed it into the language model. We denote the image, video, text, audio, and speech embedding representations to $I=(I_1, ..., I_N)$, $V=(V_1, ..., V_N)$, $T=(T_1, ..., T_z)$, $A=(A_1, ..., A_k)$, and $S=(S_1, ..., S_L)$, respectively, where $N$, $z$, $k$, and $L$ refer to the encoding sequence length of different modalities: visuals, text, audio, and speech. Take understanding a video as an example, the calculation process of the $l$ block configured with MoE is as follows:
\begin{align}
        x_0 &= [V_1, ...., V_N; T_1,..., T_z; A_1, ..., A_k],\vspace{2.0pt}\\
         X_l^{s} &= \text{MSA}(\text{LN}(X_{l-1})) + X_{l-1}, \vspace{2.0pt} \\
         X_l^{M} &= \text{MoE}(\text{LN}(X_l^{s})) + X_l^{s}, \vspace{2.0pt} \\
         x_l &= \text{LN}(X_l^{M}),
         \end{align}
where ``MSA'' and ``LN'' refer to the multi-head self-attention and layer normalization. $X_{l-1}$ shows the output of \textit{l-1} th block. For the MoE layer, we adopt the sparse router to select the corresponding Top-$k$ experts at the token level. When we introduce a set of experts $E = (e_1, e_2, ..., e_M)$, the router is a linear function to predict the probability of each token being assigned to each expert. The outputs of selected experts will be added according to the gating weights. We formulate the whole calculation process as follows:
\begin{align}
        \mathcal{P}( X_l^{s})_i &=\frac{e^{f( X_l^{s} )_i}}{\sum_j^M e^{f( X_l^{s} )_j}}, \vspace{3.0pt}\\
         \operatorname{MoE}(X_l^{s}) &=\sum_{i=1}^k \mathcal{P}(X_l^{s})_i \cdot e(X_l^{s})_i,
\end{align}
where $f(x) = \mathbf{W} \cdot x$ is the linear router to produce  expert assignment probabilities and $\mathbf{W} \in \mathbf{R}^{d\times z}$. $d$ is the last dimension of hidden states and $M$ is the total number of experts.

\begin{algorithm}[t]
\footnotesize
\centering
\caption{\label{algorithms_1}Uni-MoE Optimization Framework}
    \begin{algorithmic}[1]
        \REQUIRE Set of modalities $M_I$, pretraining datasets $\mathcal{PD}_M$, instruction tuning datasets $\mathcal{D}_M$ with a set of templates $\Pi = \{I_{M_t} : M \in M_I, t \in \mathcal{T}\}$ for each task $t \in \mathcal{T}$
        \STATE Initialize text tokenizer $h$ and corresponding embedding layer $E$
        \STATE \textbf{\textcolor{blue}{Stage 1: Cross-Modality Alignment}}
        \FOR{each modality $M$ in $M_I$}
            \STATE Initialize modality-specific pre-trained encoder $Enc_M$
            \STATE Initialize Q-Former module $QF_A$ and $QF_S$ for audio and speech inputs and modality-specific linear projection layer $LP_M$ for each modality 
            \FOR{each step in a number of iterations}
                \STATE Sample $(x, y)$ from $\mathcal{PD}_M$
                \STATE $x_{M} \leftarrow Connector(x)$ \COMMENT{Modality Projection}
                \STATE Prediction $\leftarrow LLM(x_{M})$ \COMMENT{Get LLM’s prediction}
                \STATE  Loss $\leftarrow \mathcal{L}_{CE}(\text{Prediction}, h(y))$ \COMMENT{Calculate cross-entropy loss}
                \STATE $\theta \leftarrow \theta - \alpha \nabla_{\theta}\text{Loss}$ \COMMENT{\textit{Update Linear and audio/speech Q-Former parameters}}
            \ENDFOR
        \ENDFOR
        \STATE \textbf{\textcolor{blue}{Stage 2: Training Modality-Specific Experts}}
        \FOR{each modality $M$ in $M_I$}
            \STATE Copy corresponding weights from Stage 1.
            \FOR{each step in a number of iterations}
                \STATE Sample $(x, y)$ from $\mathcal{D}_M$
                \STATE $x_{M} \leftarrow Connector(x)$ \COMMENT{Modality Projection}
                \STATE Prediction $\leftarrow LLM(x_{M}, E(h(i_M)))$ \COMMENT{Get LLM’s prediction}
                \STATE  Loss $\leftarrow \mathcal{L}_{CE}(\text{Prediction}, h(y))$ \COMMENT{Calculate cross-entropy loss}
                \STATE $\theta \leftarrow \theta - \alpha \nabla_{\theta}\text{Loss}$ \COMMENT{\textit{Update Linear and MLP (in LLM) parameters}}
            \ENDFOR
        \ENDFOR
        \STATE \textbf{\textcolor{blue}{Stage 3: Tuning Uni-MoE}}
        \FOR{each step in a number of iterations}
            \STATE Sample $(x, y)$ from $\bigcup \mathcal{D}_M$
            \STATE $X'_M \leftarrow Connector(x)$ \COMMENT{Modality Projection}
            \STATE $x_{LLM} \leftarrow E(h(Context_{M}))\|X'_M \,||\, E(h(i_M))\,||\,E(h(x)))$ \COMMENT{Inputs Concatenation}
            \STATE Prediction $\leftarrow LLM(x_{LLM})$ \COMMENT{Get LLM’s prediction}
            \STATE Loss $\leftarrow \mathcal{L}_{CE}(\text{Prediction}, h(y))$ \COMMENT{Calculate cross-entropy loss}
            \STATE $\theta \leftarrow \theta - \alpha \nabla_{\theta}\text{Loss}$ \COMMENT{\textit{Update Linear and LoRA parameters}}
        \ENDFOR
    \end{algorithmic}
\end{algorithm}

To speed up the training process, we freeze all parameters of LLMs including added experts, applying Low-Rank Adaption~\cite{lora} (LoRA) to activate each expert. We denote the tokens of the input for one sequence inputted to the first expert in the MoE layer by $\mathbf{X}_{E_1}$. The computed process for this expert could be given in

\begin{align}
        \mathbf{h}_{e_1} &= e_1(\mathbf{X}_{E_1}), \vspace{3.0pt}\\
        \mathbf{h}_{e_1}^{LoRA} &= \text{LoRA-}e_1(\mathbf{X}_{E_1}), \vspace{3.0pt}\\
        \text{LoRA}(W_0) &:= W_0 X+\Delta W X=W_0 x + B A X, \vspace{3.0pt}\\
        \mathbf{h}_{e_1} &= \mathbf{h}_{e_1} + \mathbf{h}_{e_1}^{LoRA}\\
\end{align}
where $B, A$ are learnable parameters added for each pretrained linear weight $W_0$ of experts and the final output of this expert is $\mathbf{h}_{e_1}$. By adopting this method for each expert, the training process becomes efficient, as it does not require updating the overall parameters of experts.

\begin{table*}[t]
\renewcommand\arraystretch{1.20}
\scriptsize
\caption{\label{architecture_moe}
\textbf{Detailed Architecture of Uni-MoE and Comparison with Visual-Language MoE-LLaVA}. Portions of this table are reported by the MoE-LLaVA model~\cite{moeLlava}. "Width" represents the dimension of the hidden states. "FFN" denotes the dimension of the feed-forward network's intermediate layer. "FFN Factor" represents the quantity of linear layers in the FFN. "Activated" or "Total Param" refers to the activated or total number of parameters. "7B×4-Top2" denotes a dense foundation model with 7B parameters, which is designed to incorporate a total of four experts, with two of them being activated. "†" indicates all layers are equipped with the MoE layer.}
\centering
\begin{tabular}{l|ccc|cccccc|cc}
\hline
\textbf{Name} & \textbf{Experts} & \textbf{Top-k} & \makecell{\textbf{MoE}\\\textbf{Layers}} & \textbf{Embedding} & \textbf{Width} & \textbf{Layers} & \textbf{FFN} & \makecell{\textbf{FFN}\\\textbf{Factor}} & \textbf{Heads} & \makecell{\textbf{Activated}\\\textbf{Param}} & \makecell{\textbf{Total}\\\textbf{Param}}\\
\hline
{Phi2-2.7B\cite{phi}} & - & - & - & 51200 & 2560 & 32 & 10240 & 2 & 32 & 2.7B & 2.7B \\
{MoE-LLaVA-2.7B×4-Top2} & 4 & 2 & 16 & 51200 & 2560 & 32 & 10240 & 2 & 32 & 3.6B & 5.3B \\
{MoE-LLaVA-2.7B×4-Top2†} & 4 & 2 & 32 & 51200 & 2560 & 32 & 10240 & 2 & 32 & 4.5B & 7.8B \\
\hline
{OpenChat-7B~\cite{openchat}} & - & - & - & 32000 & 4096 & 32 & 14336 & 3 & 32 & 6.7B & 6.7B \\
{MoE-LLaVA-7B×4-Top2} & 4 & 2 & 16 & 32000 & 4096 & 32 & 14336 & 3 & 32 & 9.6B & 15.2B \\
{MoE-LLaVA-7B×4-Top2†} & 4 & 2 & 32 & 32000 & 4096 & 32 & 14336 & 3 & 32 & 12.4B & 23.7B \\
\hline
{Vicuna-7B~\cite{vicuna2023}} & - & - & - & 32000 & 4096 & 32 & 11008 & 3 & 32 & 6.7B & 6.7B \\
{Uni-MoE-7B×4-Top2} & 4 & 2 & 16 & 32000 & 4096 & 32 & 11008 & 3 & 32 & 8.9B & 13.2B \\
{Uni-MoE-7B×4-Top2†} & 4 & 2 & 32 & 32000 & 4096 & 32 & 11008 & 3 & 32 & 11.1B & 19.7B \\
{Uni-MoE-7B×8-Top2} & 8 & 2 & 16 & 32000 & 4096 & 32 & 11008 & 3 & 32 & 8.9B & 21.9B \\
{Uni-MoE-7B×8-Top2†} & 8 & 2 & 32 & 32000 & 4096 & 32 & 11008 & 3 & 32 & 11.1B & 37.0B \\
\hline
\end{tabular}
\end{table*}

\subsection{Training Strategy}
\label{training_strategy}

\begin{figure}[t]
    \centering
    \includegraphics[width=0.48\textwidth]{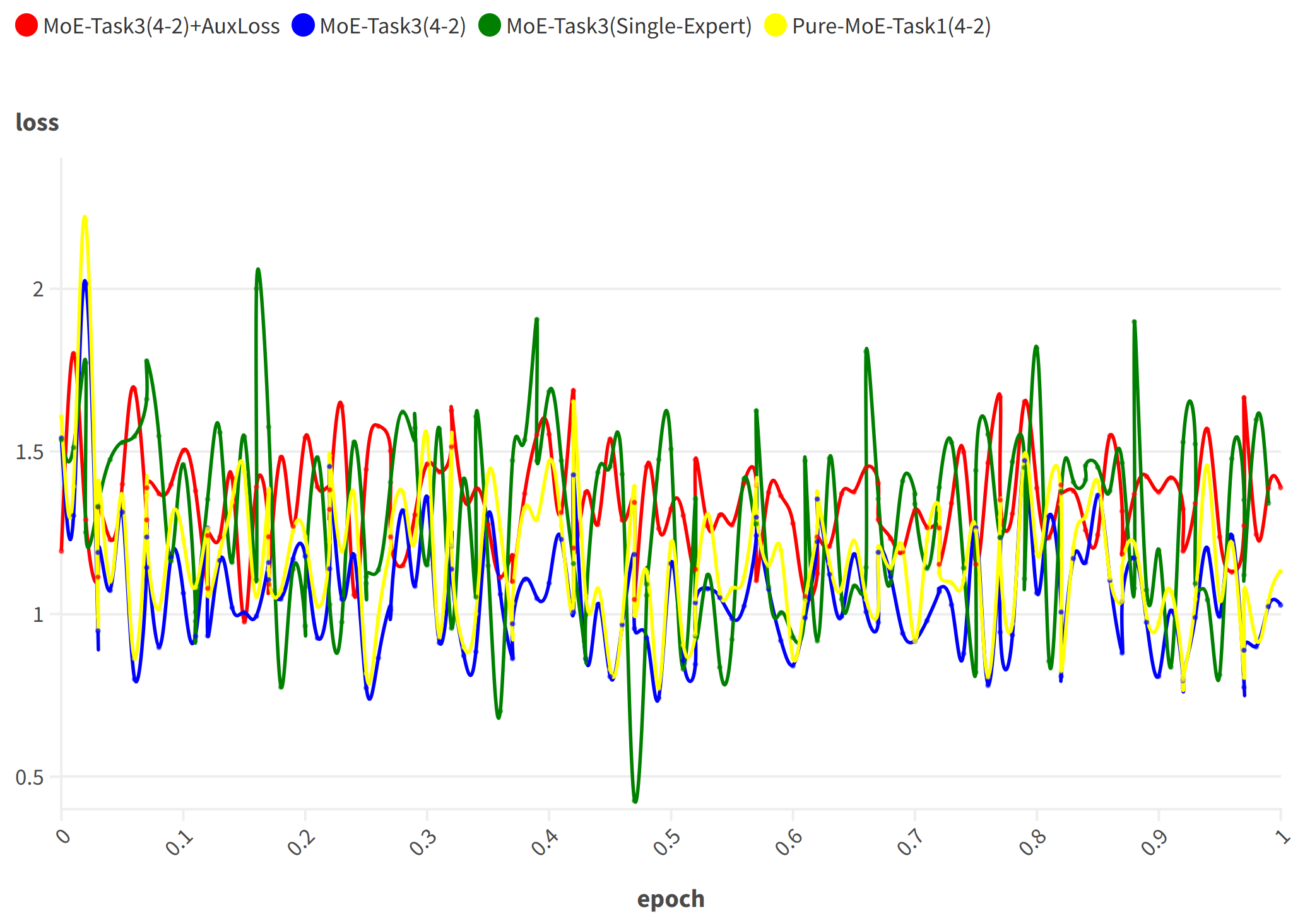}
    \caption{\textbf{Comparisons of loss curves under various MoE settings}. For MoE-TaskX (Z-Y), ``X'', ``Z'', and ``Y'' represent the specific training tasks presented in Table~\ref{dataset_tasks}, and this model contains ``Z'' experts in the MoE layer and selects ``Y'' experts for each token. ``AuxLoss'' indicates the widely-used auxiliary balancing loss~\cite{lepikhin2020gshard}, which is designed to encourage giving all experts the same importance. 
    Uni-MoE variant with MoE-Task3(4-2) (\textcolor{blue}{blue} line) achieves more stable coverage compared to model variants with fewer experts or identical expert settings. }
    \label{fig:loss_curve}
\end{figure}

To harness the distinct capabilities of various experts, enhance the efficacy of multi-expert collaboration, and bolster the overall framework's generalization, we introduce a progressive training strategy for the incremental development of Uni-MoE. Algorithm~\ref{algorithms_1} outlines the comprehensive three-stage training protocol designed to actualize the integrated MoE-based MLLM architecture. In the subsequent statements, we will delineate the objectives and elaborate on the specifics of each training stage.

\noindent\textbf{Stage 1: Cross-Modality Alignment}. 
In the initial stage, we aim to establish connectivity between different modalities and linguistics. We achieve this by constructing connectors that translate various modal data into soft tokens within the language space. The primary objective here is minimizing generative entropy loss. As illustrated in the upper section of Figure~\ref{fig:model}, the LLM is optimized to generate descriptions for inputs across modalities and only the connectors are subject to training. This approach ensures seamless integration of all modalities within a unified language framework, facilitating mutual comprehension by the LLM.

\noindent\textbf{Stage 2: Training Modality-Specific Experts}.
This stage concentrates on developing single-modality experts through dedicated training on specific cross-modal data. The goal is to refine each expert's proficiency within its respective domain, thereby enhancing the overall performance of the MoE system on diverse multimodal data. While maintaining generative entropy loss as the focal training metric, we tailor the FFN to align more closely with the characteristics of the targeted modality.

\noindent\textbf{Stage 3: Tuning Uni-MoE}.
The concluding stage involves integrating the expertly tuned weights from Stage 2 into the MoE layers. We then proceed to jointly fine-tune the MLLMs utilizing mixed multimodal instructional data. The progression of the training process, as reflected by the loss curves, is depicted in Figure~\ref{fig:loss_curve}. Comparative analysis between MoE configurations reveals that experts refined during Stage 2 achieve quicker convergence and display enhanced stability on mixed-modality datasets. Furthermore, in scenarios involving complex mixed-modal data, encompassing video, audio, images, and text, the model employing four experts demonstrates reduced loss variability and more consistent training performance than its two-expert counterpart. Uni-MoE shows better convergence compared to using auxiliary balancing loss, which caused the overall training loss to fluctuate and did not show obvious convergence.

\begin{table*}[t]
\renewcommand\arraystretch{1.30}
\centering
\caption{\label{dataset_tasks}
\textbf{All training tasks}.
``*'' refers to that the dataset we use is only a subset. ``MC'' represents the multi-choice setting. ``I-A'' means image-audio pairs, where the question is converted into the corresponding speech.  ``T-I'' shows the original text-image pairs. ``T-A'' indicates the contextual paragraph of the RACE dataset is transferred into the long speech. ``Pretraining task'' represents the tasks included in the previous training stage. ``Pure-Task'' indicates the MoE-based model contains identical experts and is trained with specific training data.}
\resizebox{\linewidth}{!}{
\begin{tabular}{l|ccccc}
\hline
\textbf{Training Tasks} & \textbf{Data Types} & \textbf{Data Size} & \textbf{Epochs} & \textbf{Trainable Modules} & \textbf{Pretraining tasks}\\
\hline
{Audio-Language Pretraining} & \makecell{WavCaps*,Audiocap*, \\ MELD, ClothoV1} & 194K & 2 & \makecell{Audio Q-former,\\Audio projection layer} & {-}\\
\hline
{Speech-Language Pretraining} & \makecell{Common Voice \\(Short Speech) } & 1.7M & 2 & \makecell{Speech Q-former,\\Speech projection layer} & {-}\\
\hline
{Single-Modality-Expert-Task1} & \makecell{LLaVA-Instruct-150K(I-A)} & 150K & 1 & \makecell{LoRA,\\speech \& image projection layer} & {Speech-pretrain-task}\\
\hline
{Single-Modality-Expert-Task2} & \makecell{LLaVA-Instruct-150K(T-I)} &150K & 1 & \makecell{LoRA,\\image projection layer} & {Speech-pretrain-task}\\
\hline
{Single-Modality-Expert-Task3} & \makecell{LLaVA-Instruct-150K(I-A)}  & 150K& 1 & \makecell{LoRA, Audio Q-former,\\speech \& image projection layer} & {Speech-pretrain-task}\\
\hline
{Single-Modality-Expert-Task4} & \makecell{LLaVA-Instruct-150K(I-A),\\RACE(T-A), LibriSpeech  } & 271K & 1 & \makecell{LoRA,\\speech \& image projection layer} & {Speech-pretrain-task}\\
\hline
{Single-Modality-Expert-Task5} & \makecell{LLaVA-Instruct-150K(T-I),\\RACE(T-A), LibriSpeech  } & 271K & 1 & \makecell{LoRA,\\speech \& image projection layer} & {Speech-pretrain-task}\\
\hline
{Single-Modality-Expert-Task6} & \makecell{LLaVA-Instruct-150K(I-A),\\LLaVA-Instruct-150K(T-I),\\RACE(T-A), LibriSpeech  } & 421K & 1 & \makecell{LoRA,\\speech \& image projection layer} & {Speech-pretrain-task}\\
\hline
{Single-Modality-Expert-Task7} & \makecell{RACE(T-A), LibriSpeech ,\\RACE(T-A)-MC} & 209K & 1 & \makecell{LoRA,\\speech projection layer} & {Speech-pretrain-task}\\
\hline
{Single-Modality-Expert-Task8} & \makecell{WavCaps*,Audiocap*,MELD\\ClothoAQA,ClothoV1} & 203K &  1 & \makecell{LoRA,\\audio projection layer} & {Audio-pretrain-task}\\
\hline
{MoE-Task1} & \makecell{LLaVA-Instruct-150K(I-A),\\LLaVA-Instruct-150K(T-I),\\RACE(T-A), LibriSpeech  \\RACE(T-A)-MC} & 509K &  3 & \makecell{LoRA, Router,\\speech \& image projection layer} & \makecell{LLaVA-v1.5-LoRA, \\ Single-Modality-Expert-Tasks 2 / 3 / 7}\\
\hline
{MoE-Task1-short-speech} & \makecell{LLaVA-Instruct-150K(I-A),\\LLaVA-Instruct-150K(T-I)} & 300K & 3 & \makecell{LoRA, Router,\\speech \& image projection layer} & \makecell{LLaVA-v1.5-LoRA, \\ Single-Modality-Expert-Tasks 2 / 3 / 7}\\
\hline
{MoE-Task2} & \makecell{Video-Instruct-Dataset,\\LLaVA-Instruct-150K(T-I),\\RACE(T-A), LibriSpeech  \\RACE(T-A)-MC} & 459K & 2 & \makecell{LoRA, Router,\\speech \& image projection layer} & \makecell{llava-v1.5-LoRA,\\Single-Modality-Expert-Tasks 2 / 3 / 7}\\
\hline
{MoE-Task3} & \makecell{Video-Instruct-Dataset,\\LLaVA-Instruct-150K(T-I),\\WavCaps*,Audiocap*,MELD\\ClothoAQA,ClothoV1} & 453K & 2 & \makecell{LoRA, Router,\\audio \& image projection layer} & \makecell{ LLaVA-v1.5-LoRA, \\ Single-Modality-Expert-Tasks 2 / 3 / 8}\\
\hline
{MoE-Task4} & \makecell{Video-Instruct-Dataset,\\ LLaVA-v1.5-665K(T-I)~\cite{liu2023improved},\\RACE(T-A), LibriSpeech  \\RACE(T-A)-MC} & 874K & 2 & \makecell{LoRA, Router,\\speech \& image projection layer} & \makecell{ LLaVA-v1.5-LoRA, \\ Single-Modality-Expert-Tasks 2 / 3 / 7}\\
\hline
{Pure-MoE-Task1} & \makecell{Video-Instruct-Dataset,\\LLaVA-Instruct-150K(T-I),\\WavCaps*,Audiocap*,MELD\\ClothoAQA,ClothoV1} & 453K & 2 & \makecell{LoRA, Router,\\audio \& image projection layer} & \makecell{ LLaVA-v1.5-LoRA}\\
\hline
{Pure-MoE-Task2} & \makecell{Video-Instruct-Dataset,\\LLaVA-Instruct-150K(T-I),\\WavCaps*,Audiocap*,MELD\\ClothoAQA,ClothoV1} & 453K & 2 & \makecell{LoRA, Router,\\audio \& image projection layer} & -\\
\hline
\end{tabular}}
\end{table*}

\section{Experiments}

\subsection{Uni-MoE Settings}
The Uni-MoE's architectural specifications are presented in Table~\ref{architecture_moe}, exhibited along with the recently proposed MoE-LLaVA. Our exploration is anchored on scaling the Uni-MoE model, which is based on the LLaMA-7B architecture. 
The design and optimization of Uni-MoE are guided by specialized training tasks listed in Table~\ref{dataset_tasks}. These tasks are instrumental in refining the capabilities of the MLP layers, thereby leveraging their specialized knowledge for enhanced model performance. We undertake eight single-modality expert tasks to elucidate the differential impacts of various training methodologies. Our comprehensive training approach uses the MoE framework to execute six diverse tasks spanning multiple modalities, including video, audio, speech, image, and text. This multi-faceted training task evaluates the Uni-MoE’s performance under varied MoE configurations, thus ensuring a robust and versatile modelling approach adaptable to diverse data types and applications.

\begin{table*}[t]
\renewcommand\arraystretch{1.10}
\centering
\caption{\label{automatic_speech_image}
\textbf{Experimental results on zero-shot speech-image and long speech understanding benchmarks}. The questions of A-OKVQA, OK-VQA, and VQAv2 are converted into speech by the Text-to-Speech technical. MMBench-Audio (Input Type: Text-Speech-Image) and RACE-Audio are used for held-out testing of long speech understanding and reasoning, where ``middle'' and ``high'' refer to the Chinese English examinations designed for middle school and high school students. ``EHSL'' indicates our collected English High School Listening Test (Long Speech) in real world.
Empty experimental results indicate that these model variants are not adequate for long-speech reasoning. 
\textit{``Uni-MoE'' in Tables 3-6 refers to Uni-MoE-7B×4-Top2† (speech) and Uni-MoE-7B×4-Top2 (audio)}.
}
\resizebox{0.95\linewidth}{!}{
\begin{tabular}{l|cccccccc}
\hline
\multirow{2}*{\textbf{Method}} & \multirow{2}*{\textbf{A-OKVQA}} & \multirow{2}*{\textbf{OK-VQA}} &\multirow{2}*{\textbf{VQAv2}}& \multirow{2}*{\textbf{MMBench-Audio}} & \multicolumn{2}{c}{\textbf{RACE-Audio}} & \multicolumn{2}{c}{\textbf{EHSL}} \\
\cline{6-9} 
&&&&&  middle & high & Long & Short\\
\hline
{Macaw-LLM~\cite{lyu2023macaw}} & 1.08\% & 1.06\% & 2.33\% & {1.86\%}  & 4.04\% & 3.00\% & 0.67\% & 2.00\%\\
{X-InstructBLIP~\cite{panagopoulou2023x}} & 0.91\% & 0.00\% & 26.47\% & {11.23\%} & 16.33\% & 18.88\% & 0.67\% & 2.00\% \\
\hline
{Single-Modality-Expert-Task1} & 18.38\% & 26.58\% & 32.60\% & -  & - & - & - & - \\
{Single-Modality-Expert-Task3} & 19.26\% & 27.68\% & \textbf{32.95\%} & -  & - & - & - & -\\
{Single-Modality-Expert-Task4} & 9.14\% & 12.26\% & 24.30\% & 45.96\%  & 29.46\% & 26.67\%  & 12.67\% & 6.00\%\\
{Single-Modality-Expert-Task5} & 4.56\% & 3.42\% & 5.90\% & 41.30\%  & 30.78\% & 24.90\% & 9.33\% & 12.00\%\\
{Single-Modality-Expert-Task6} & 11.40\% & 14.40\% & 25.11\% & 47.52\%  & 32.59\% & 29.02\% & 18.67\% & 8.00\% \\
{Single-Modality-Expert-Task7} & - & - & - & 47.83\% & 46.80\% & 48.74\% & 34.67\% & 46.00\%\\

\hline
{Uni-MoE w/ MoE-Task1 2 epoch} & 14.80\% & 19.98\% & 26.01\% & 44.41\%  & 47.08\% & 47.08\% & 41.33\% & 36.00\% \\
{Uni-MoE w/ MoE-Task1 3 epoch} & 13.58\% & 17.68\% & 20.14\% & 47.20\%  & \textbf{50.77\%} & 48.94\% & 32.67\% & 38.00\%\\ 
{Uni-MoE w/ MoE-Task1-short-speech  2 epoch} & 19.31\% & 26.46\% & 31.28\% & 37.58\% & - & - & - & - \\
{Uni-MoE w/ MoE-Task1-short-speech 3 epoch} & \textbf{20.01\%} & \textbf{29.04\%} & 30.86\% & 34.78\% & - & - & - & - \\
{Uni-MoE w/ MoE-Task2 2 epoch} & 7.76\% & 8.08\% & 12.68\% & \textbf{49.38\%}  & 49.65\% & \textbf{49.37\%}& \textbf{42.00\%} & \textbf{48.00\%}\\
\hline
\end{tabular}}

\end{table*}


\subsection{Datasets}

\textbf{Training Datasets}. To endow our model with speech recognition capabilities, we incorporate the \textbf{Common Voice} dataset~\cite{commonvoice} at the cross-modality alignment stage of training. This dataset comprises short speech clips, each less than 30 seconds in duration, with a cumulative total exceeding 1.7 million instances. Subsequently, we develop a tri-modal dataset derived from \textbf{LLaVA-Instruct-150K}~\cite{liu2023llava}, converting user queries into auditory format utilizing sophisticated Text-to-Speech (TTS) technologies from Microsoft Azure~\cite{tts}. Additionally, the original LLaVA-Instruct-150K dataset is employed in various training tasks to facilitate comparative analyses. To enhance the model's proficiency in understanding extended speech sequences, we integrate and concatenate audio files from the \textbf{LibriSpeech} dataset~\cite{LibriSpeech} with short speeches of 30 seconds into longer sound files, each extending up to two minutes in duration. Additionally, we convert the \textbf{RACE} dataset~\cite{race}, a comprehensive reading comprehension collection derived from English examinations in China, from its original textual format into long audio files. These transformed audio files are subsequently presented to the model, enabling it to interpret lengthy speech inputs and accurately determine appropriate responses to questions. For the task of audio captioning, the model is subjected to both cross-modality aligning and single-modality expert training processes utilizing an identical dataset collection. The \textbf{WavCaps} dataset~\cite{wavcaps}, which constitutes a ChatGPT-assisted, weakly-labelled audio captioning collection segmented into four subsets (AudioSet SL subset, SoundBible, FreeSound, and BBC Sound Effects), is employed partially within our optimization framework. Additionally, the \textbf{AudioCaps} dataset~\cite{audiocaps}, a comprehensive corpus comprising approximately 46,000 pairs of audio clips and their corresponding human-generated textual descriptions, is also selectively utilized during the training regimen. The \textbf{Clotho} and \textbf{ClothoAQA} datasets~\cite{clotho, clothoaqa} are utilized to bolster the model's proficiency in audio-related question-answering tasks. Additionally, \textbf{MELD}~\cite{meld}, a dataset designated for audio emotion detection, is employed to augment the diversity of audio-relevant tasks within our framework. In the realm of video-related tasks, which serve as a pivotal component in enhancing models' visual comprehension, the \textbf{Video-Instruct-Dataset} from Video-ChatGPT~\cite{videochatgpt}, encompassing 100,000 video-text instructional pairs, is leveraged as the training corpus to advance our models' performance in scenarios involving video content.

\noindent\textbf{Evaluation Datasets.} Our models are assessed across various benchmarks, reflecting the diversity of specialized tasks they are designed to perform. To evaluate our models' proficiency in short speech recognition and image understanding, we employ modified versions of the \textbf{A-OKVQA}~\cite{aokvqa}, \textbf{OKVQA}~\cite{okvqa}, and \textbf{VQAv2}~\cite{vqa2} benchmarks, utilizing speech synthesis technologies TTS to convert questions into human speeches. To design tasks with long speech, we leverage the image-text reasoning dataset \textbf{MMBench}~\cite{mmbench} and utilize TTS to transform contextual hints into long audios, called \textbf{MMBench-Audio}, along with the speech version of the \textbf{RACE} evaluation set, named \textbf{RACE-Audio}, to assess our models rigorously. More precisely, the proficiency of our models is also evaluated by their performances on the \textbf{English listening part of the Chinese College Entrance Examination}, to check their practical real-world speech recognition capabilities. \textit{This compact dataset comprises 150 questions related to long audio segments with an average length of 109 seconds, and 50 questions about short audio segments with an average length of 14 seconds.} The format of these materials aligns with that of the RACE-Audio evaluation dataset. Additionally, the environmental audio understanding ability is evaluated by utilizing the test sets of \textbf{Clotho V1/2} and \textbf{ClothoAQA}. In the context of video-related tasks, the performance of Uni-MoE is gauged using benchmarks from \textbf{ActivityNet-QA}~\cite{activitynet} and \textbf{MSVD-QA}\cite{ msvd}, which are instrumental in assessing video comprehension and interaction capabilities.

\subsection{Evaluation Metrics}
For the visual question-answering benchmarks including A-OKVQA, OK-VQA, and VQA2, we adopt the Exact Match (EM) metric to evaluate if the model's predicted answers align perfectly with the ground truths. For multi-choice formats such as MMBench, RACE, and the English Listening Test, we apply strict accuracy metrics to assess the correctness of chosen responses. In the audio quality assessment for ClothoAQA, the EM accuracy metric is utilized to gauge the precision of responses. For assessing the relevance and quality of generated captions in Clotho, we employ the CIDEr metric. Additionally, for video-related benchmarks like ActivityNet and MSVD-QA, we implement a comprehensive accuracy verification procedure inspired by the methodology used in Video-ChatGPT~\cite{videochatgpt}.

\subsection{Implementation Details}

We employ the AdamW~\cite{kingma2014adam} optimizer in conjunction with a cosine learning rate scheduler to train our models in all stages. In the cross-modality alignment phase, we utilize 2 A100 GPUs to separately process 1.7 million short-speech data and 194K short audio captioning data with a global batch size of 32 and a base learning rate of 2e-5. Notably, during this stage, only the Qudio/Speech Q-former and projection layer are trained. Subsequently, we train our model to handle specific tasks using various combinations of data on the same devices, employing a global batch size of 16 and a base learning rate of 4e-5. During fine-tuning audio connectors with audio captioning data, we apply 4e-4 as the LoRA learning rate and 3e-5 as the learning rate of the audio projection layer. In our implementation, we set the LoRA rank to 64 and alpha to 16, and only apply it to tune the MLP in LLMs. Transitioning to the MoE training stage, we utilize a single, eight, or sixteen (two nodes) A800 GPU to train our models. We set the rank to 8 and alpha to 16 in this phase, maintaining a learning rate of 4e-5 for both LoRA parameters and projection layer parameters. Notably, we implement data parallelism and expert parallelism for mixed multi-modal data, as well as multi-nodes and multi-GPUs parallel training for larger models (with more than 8 experts).

\subsection{Overall Performance}
We evaluate the performance of various model variants across a diverse set of benchmarks, encompassing five speech-related, three audio-understanding, and two challenging video-understanding tasks. We compare the results against two foundational unified multimodal models: \textbf{Macaw-LLM}~\cite{lyu2023macaw}, which represents a pioneering effort in multi-modal language modeling, integrating image, video, audio, and text data; and \textbf{X-InstructBLIP}~\cite{panagopoulou2023x}, noted for its simplicity and effectiveness in handling multiple modalities simultaneously. Both models serve as strong baselines due to their innovative multimodal integration approaches.

\begin{table}[t]
\renewcommand\arraystretch{1.15}
\centering
\caption{\label{audio_understanding}
\textbf{Experimental Results on audio understanding benchmarks (Audio-Text)}. These audio understanding tasks focus on environmental audio and we adopt the CIDEr metric, following X-InstructBLIP~\cite{panagopoulou2023x}. }
\resizebox{0.95\linewidth}{!}{
\begin{tabular}{l|ccc}
\hline
\textbf{Method} & \textbf{ClothoAQA} & \textbf{ClothoV1} & \textbf{ClothoV2}\\
\hline
MWAFM~\cite{MWAFM} & 22.2\% & - & - \\
Kim et al.~\cite{kim} & - & - & 19.2\% \\
X-InstructBLIP~\cite{panagopoulou2023x} & 20.69\% & 17.8\% & 17.3\% \\
{Macaw-LLM}~\cite{lyu2023macaw} & 16.8\% & 0.2\% & 0.2\% \\
\hline
{Single-Modality-Expert-Task8} & 32.3\% & 22.3\% & 21.6\% \\
\hline
{Uni-MoE w/ MoE-Task3 1 epoch} & \textbf{32.6\%} & \textbf{25.0\%} & 24.7\% \\
{Uni-MoE w/ MoE-Task3 2 epoch} & 31.1\% & 24.8\% & \textbf{25.1\%} \\

\hline
\end{tabular}}

\end{table}

\begin{table}[t]
\renewcommand\arraystretch{1.15}
\centering
\caption{\label{image-text-und}
\textbf{Model performance on image-text understanding and reasoning benchmarks}. Uni-MoE still achieves better performance compared to models with the single expert configuration.}
\resizebox{1.0\linewidth}{!}{
\begin{tabular}{l|cccc}
\hline
\textbf{Method} & \textbf{A-OKVQA} & \textbf{OK-VQA} & \textbf{VQAv2} & \textbf{MMBench} \\
\hline
Macaw-LLM~\cite{lyu2023macaw}  & 1.90\% & 5.70\% & 20.73\% & 3.84\%  \\
X-InstructBLIP~\cite{panagopoulou2023x} & 21.52\% & 30.61\% & 37.77\% & 8.96\% \\
\hline
{Single-Modality-Expert-Task2} & 67.07\% & 62.91\% & 75.18\% & 71.26\%  \\
{Single-Modality-Expert-Task5} & 58.86\% & 56.01\% & 67.35\% & 65.80\%  \\
{Single-Modality-Expert-Task6} & 58.69\% & 57.77\% & 68.74\% & 67.53\%  \\
\hline

{Uni-MoE w/ MoE-Task1 2 epoch} & 61.22\% & 57.63\% & 68.42\% & 68.15\%  \\
{Uni-MoE w/ MoE-Task1 3 epoch} & 59.91\% & 57.07\% & 68.05\% & 67.28\%  \\ 
{Uni-MoE w/ MoE-Task2 1 epoch} & 66.20\% & 63.02\% & 74.41\% & 71.17\%  \\
{Uni-MoE w/ MoE-Task2 2 epoch} & 65.07\% & 62.10\% & 73.87\% & 70.50\%  \\
{Uni-MoE w/ MoE-Task3 1 epoch} & 65.07\% & 61.38\% & 73.66\% & 69.52\%  \\
{Uni-MoE w/ MoE-Task3 2 epoch} & 64.28\% & 61.96\% & 73.87\% & 69.82\%  \\
\hline
\end{tabular}}

\end{table}
\textbf{Speech-Image and Speech-Text Understanding: } 
We adopt the evaluation of manually constructed datasets of three modalities: text, image, and speech.  The speech recognition ability is shown and compared in Table~\ref{automatic_speech_image}. Firstly, we observe that previous baselines have inferior performances on speech understanding, which further affects their performances on image-speech reasoning.
Additionally, introducing automatically generated audio-image datasets based on text-image instruction data improves the overall performance of short speech and image understanding, e.g., the model trained with Single-Modality-Expert-Task6 significantly outperforms that with Task5 (without introducing audio-image training data) in the audio-image setting of A-OKVQA, OK-VQ, and VQA2. The comparative model performances trained with Single-Modality-Expert and MoE tasks show that \textit{1) Introducing the MoE architecture can improve the generalization of MLLMs on unseen audio-image reasoning tasks such as MoE-Task2 vs. Single-Modality-Expert-Task5 on A-OKVQA, OK-VQA, VQAv2, and MMBench-Audio; 2) Uni-MoE achieves better performance on challenging long speech understanding and reasoning tasks compared to dense models, e.g., Uni-MoE significantly suppresses other model variants on English High School Listening Test; 3) When the instruction tuning data mixes long and short speech, the performance of Uni-MoE is more stable compared to single-expert model, e.g., Single-Modality-Expert-Task4 vs. MoE-Task1 in Table~\ref{automatic_speech_image}. As the training epochs increase, it focuses on improving the performance of challenging long-speech understanding tasks.}

\begin{figure*}[t]
\centering
\resizebox{0.92\linewidth}{!}{
\includegraphics[width=1.0\linewidth]{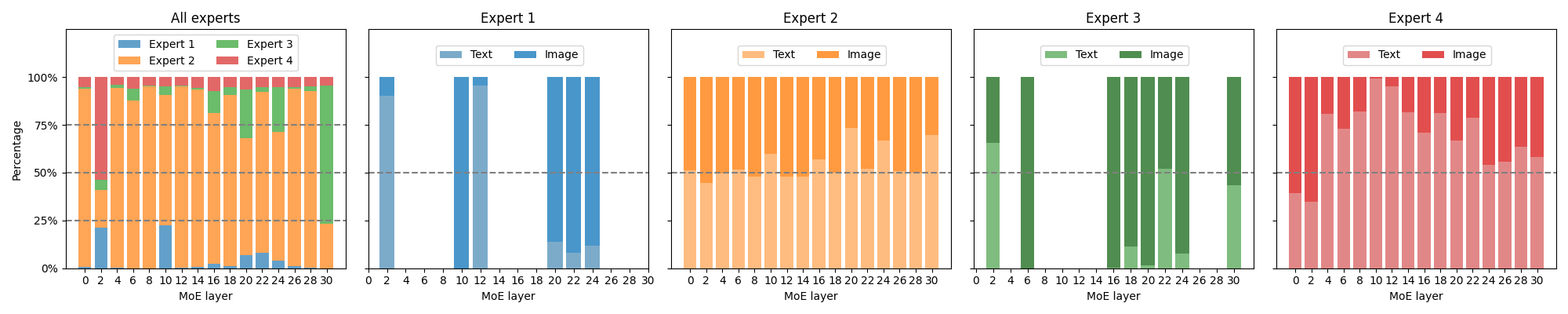}}
\resizebox{0.92\linewidth}{!}{
\includegraphics[width=1.0\linewidth]{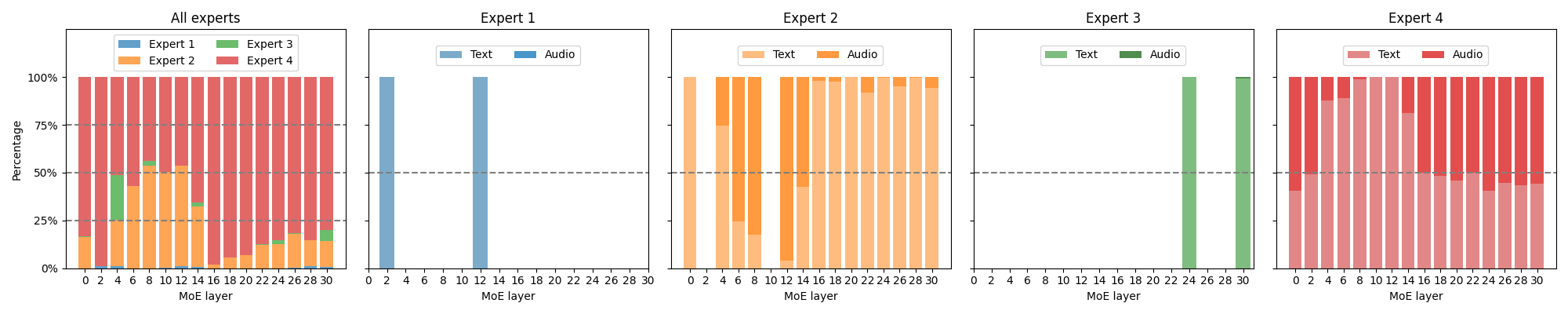}}
\resizebox{0.92\linewidth}{!}{
\includegraphics[width=1.0\linewidth]{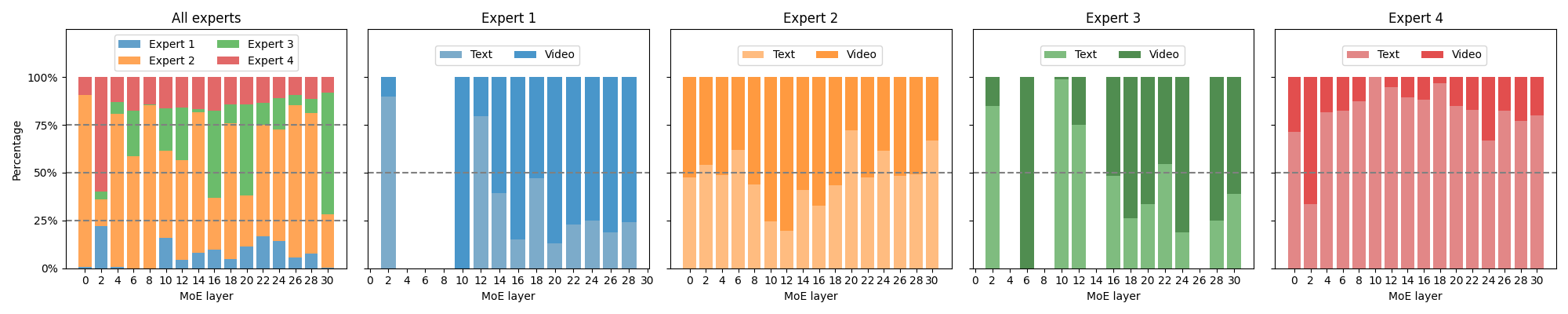}}
\resizebox{0.92\linewidth}{!}{
\includegraphics[width=1.0\linewidth]{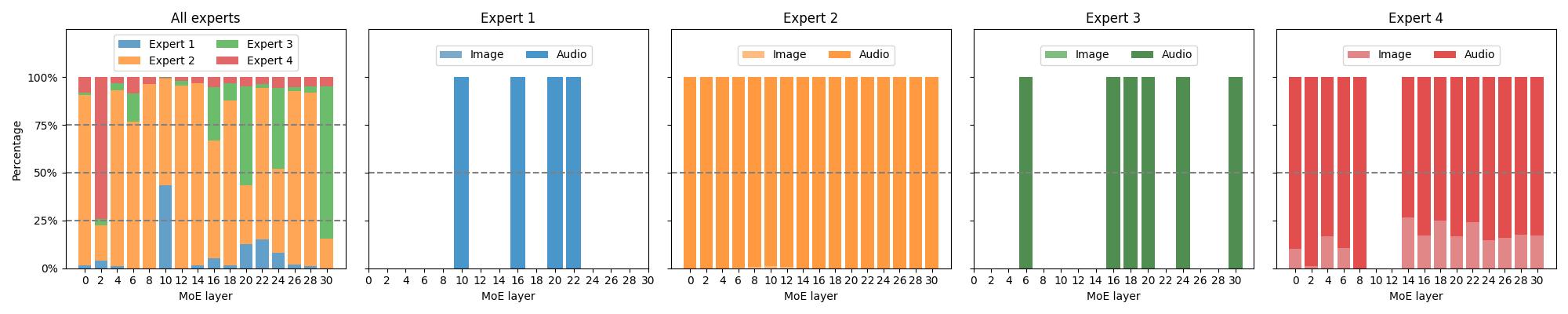}}
\resizebox{0.92\linewidth}{!}{
\includegraphics[width=1.0\linewidth]{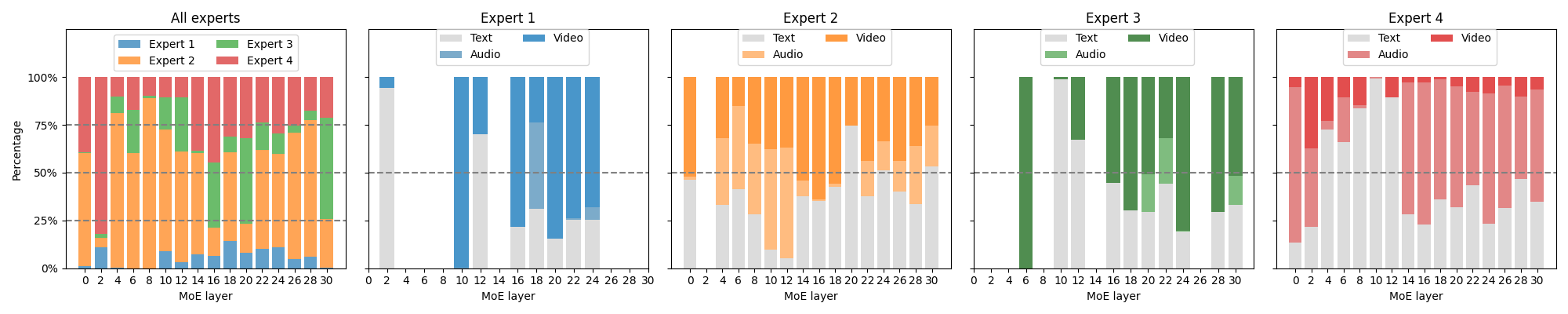}}
 \caption{\textbf{Distribution of expert loading with various cross-modality inputs.} The discontinuous lines represent the distribution of tokens among different experts or modalities. The first figure on the left illustrates the workload among experts, while the remaining four figures depict the preferences of experts towards different modalities. Five layers of figures from the top to down refer to text-image, text-audio, text-video, image-audio, and video-audio-text pair data respectively when being fed to the MoE layers of Uni-MoE trained with MoE-Task3. Each expert focuses on different modal information, which could be compared with Pure-MoE-Task1 in Figure~\ref{fig:experts_distribution_pure}.}
\label{fig:experts_distribution}
\end{figure*}

\textbf{Audio-Text Understanding: } 
In Table~\ref{audio_understanding}, our model significantly surpasses the best performance of baselines by 8.4\% on ClothoV1 and 10\% on ClothoAQA, suggesting a remarkable capability of understanding audio-related questions. 
We reproduced the results of two multimodal baselines X-InstructBLIP and Macaw.
We also observed that Uni-MoE could achieve improvement in the audio captioning and question-answering tasks compared to previous baselines and that with a single expert tuned with the same audio-related data, which sufficiently reveals the advancement of the MoE leveraged for audio and text co-reasoning. Concretely, MoE-Task3 surpasses the best result from the fine-tuning stage of ClothoV1 and ClothoV2 by 2.7\% and 3.5\% respectively.

\begin{figure}[tp]
\centering
\resizebox{0.98\linewidth}{!}{
\includegraphics[width=1.0\linewidth]{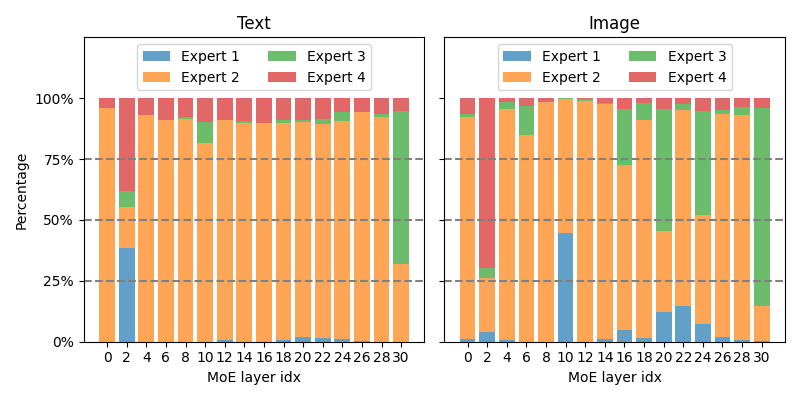}}
\resizebox{0.98\linewidth}{!}{
\includegraphics[width=1.0\linewidth]{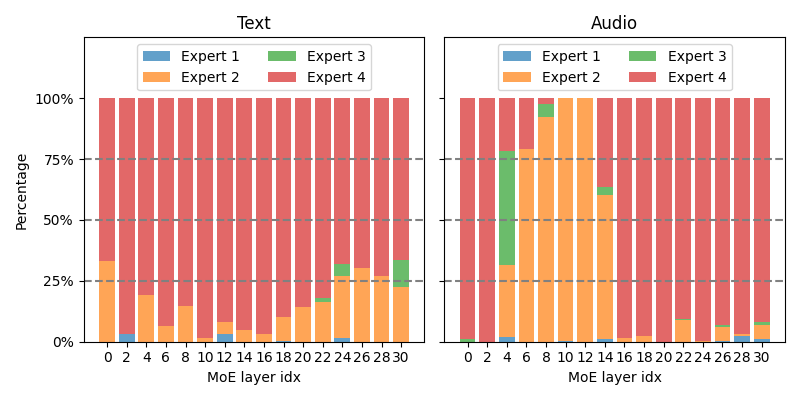}}
\resizebox{0.98\linewidth}{!}{
\includegraphics[width=1.0\linewidth]{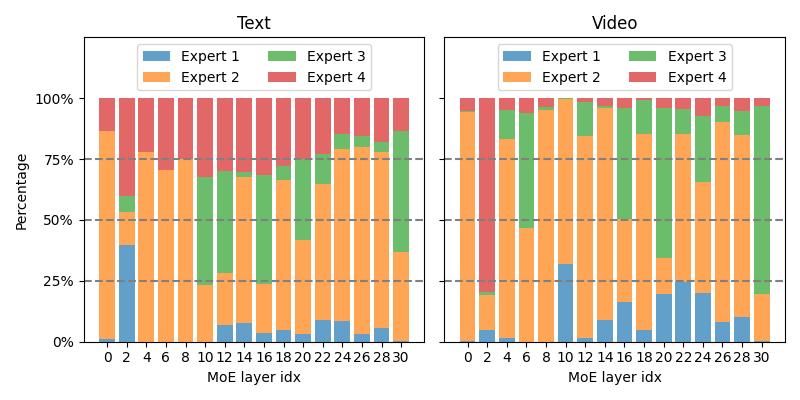}}
\resizebox{0.98\linewidth}{!}{
\includegraphics[width=1.0\linewidth]{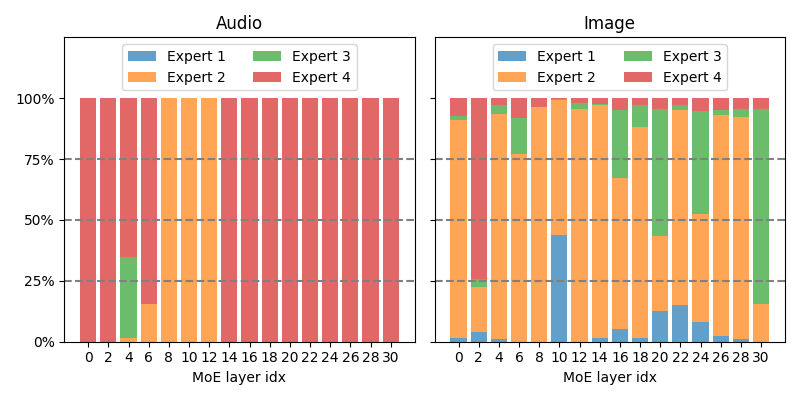}}
\resizebox{0.98\linewidth}{!}{
\includegraphics[width=1.0\linewidth]{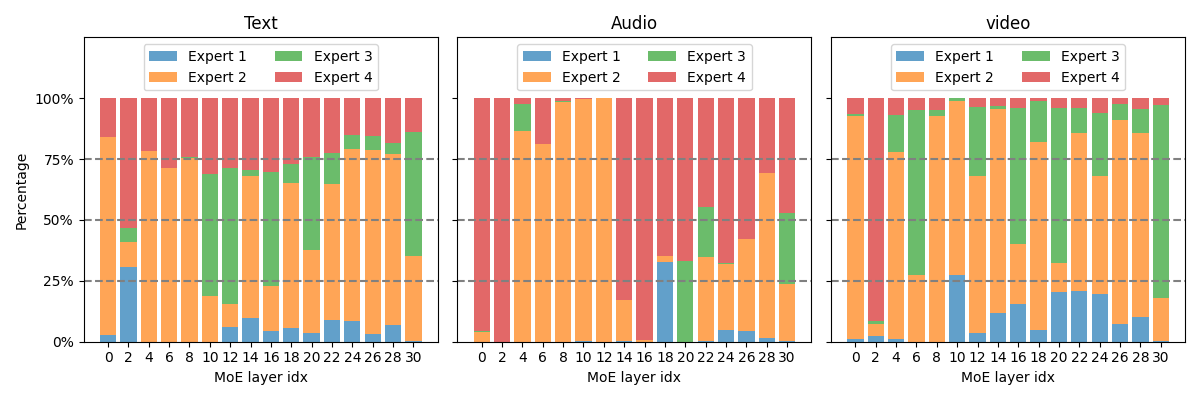}}
 \caption{\textbf{Distribution of modalities across different experts}. The discontinuous lines represent the distribution of tokens.  Five layers of sub-figures from the top to down refer to text-image, text-audio, text-video, image-audio, and video-audio-text pair data respectively when being fed to the MoE layers of Uni-MoE trained with MoE-Task3. Expert 1 refers to the original MLP layer from LLaVA-v1.5. Expert 2 indicates the MLP is optimized through LLaVA-Instruct-150k(T-I) data, i.e., Single-Modality-Expert-Task2. Expert 3 represents the MLP trained with the audio-image LLaVA-Instruct-150k(I-A) dataset, i.e., Single-Modality-Expert-Task3. Expert 4 focuses on audio understanding, i.e., Single-Modality-Expert-Task8. \textit{Training modality-specific experts is useful for expert assignment learning, facilitating the specialization of experts.}}
\label{fig:modalities_distribution}
\end{figure}

\begin{figure}[tp]
\centering
\resizebox{0.91\linewidth}{!}{
\includegraphics[width=0.95\linewidth]{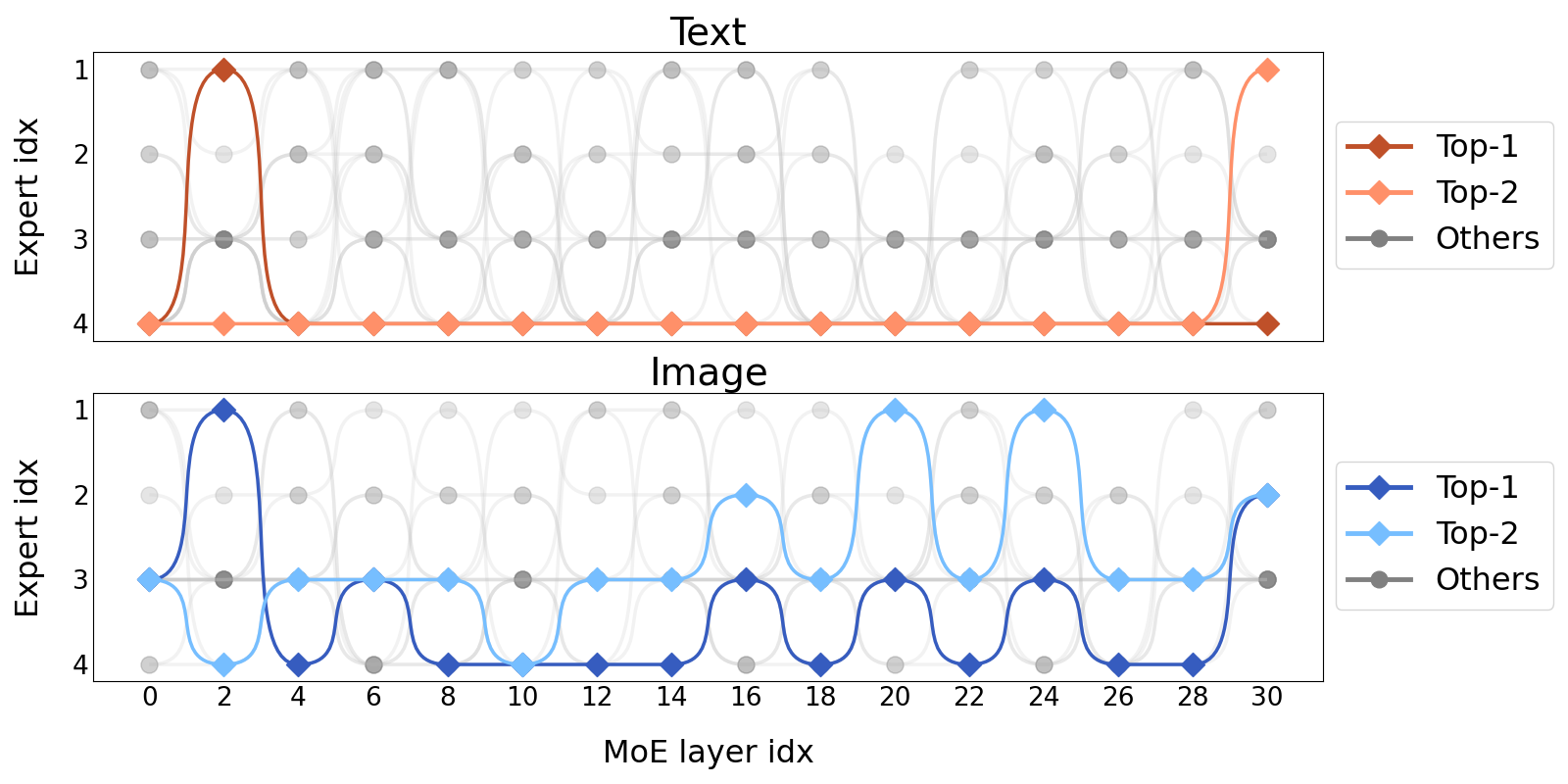}}
\resizebox{0.91\linewidth}{!}{
\includegraphics[width=0.95\linewidth]{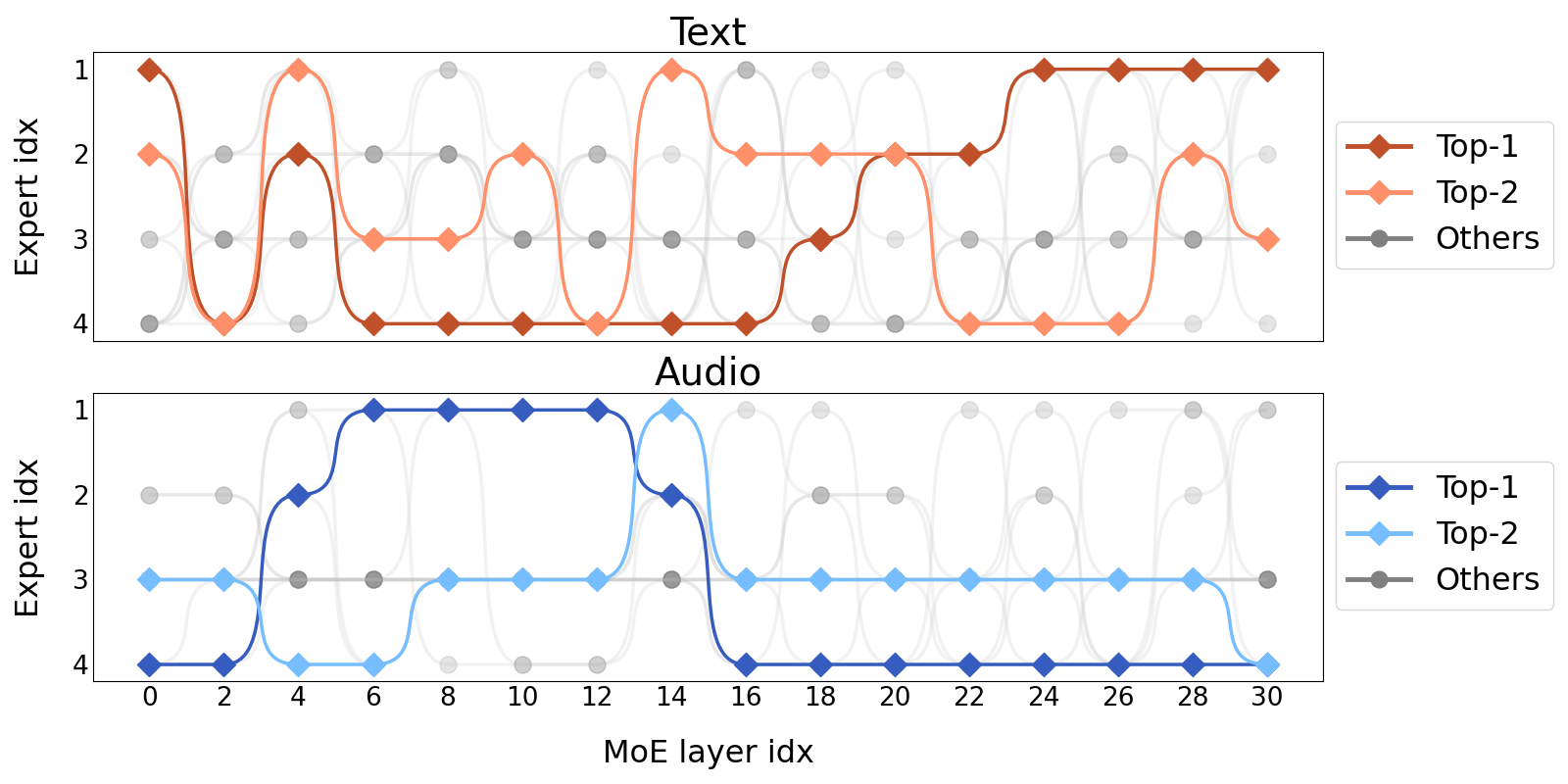}}
\resizebox{0.91\linewidth}{!}{
\includegraphics[width=0.95\linewidth]{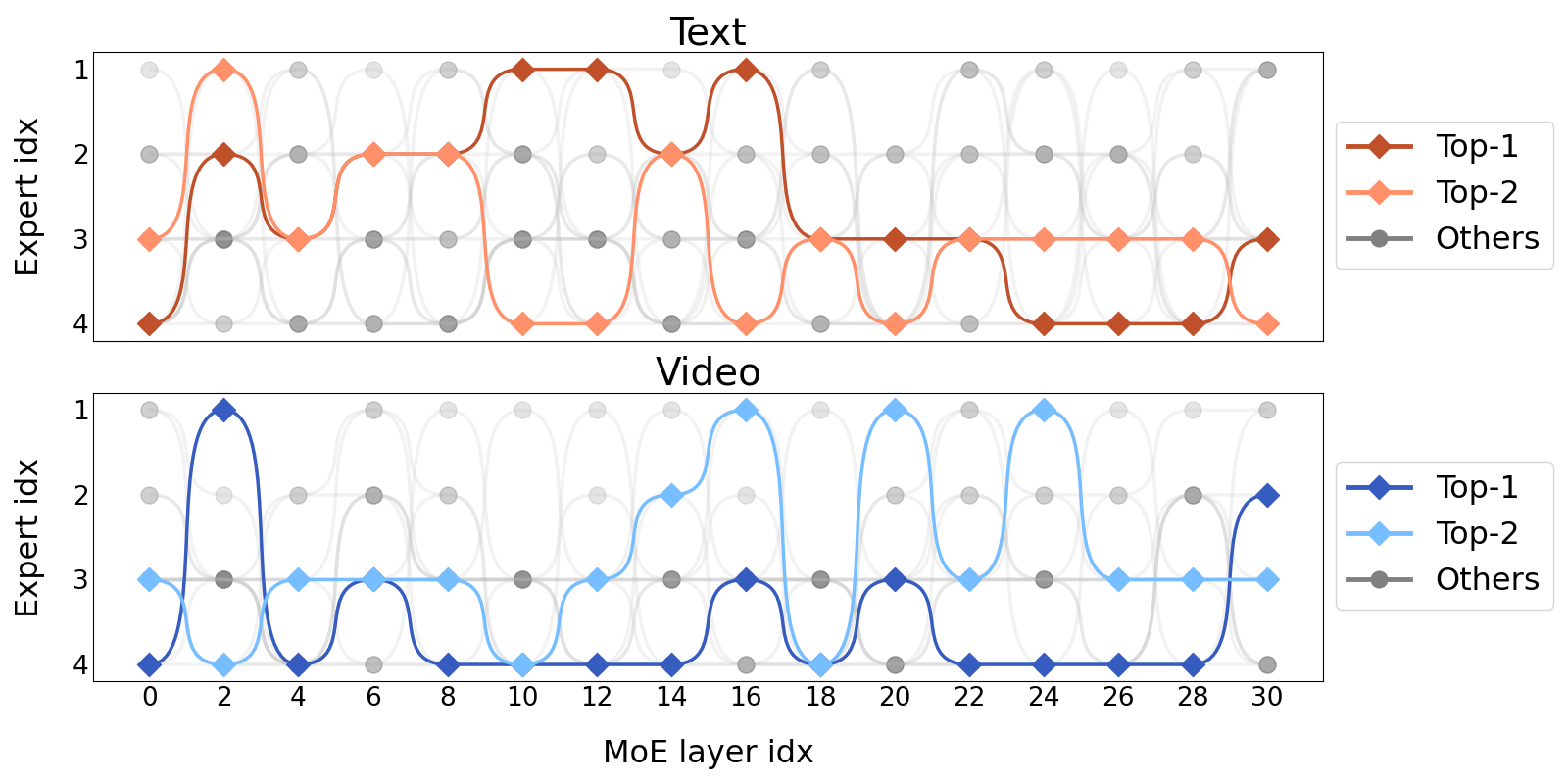}}
\resizebox{0.91\linewidth}{!}{
\includegraphics[width=0.95\linewidth]{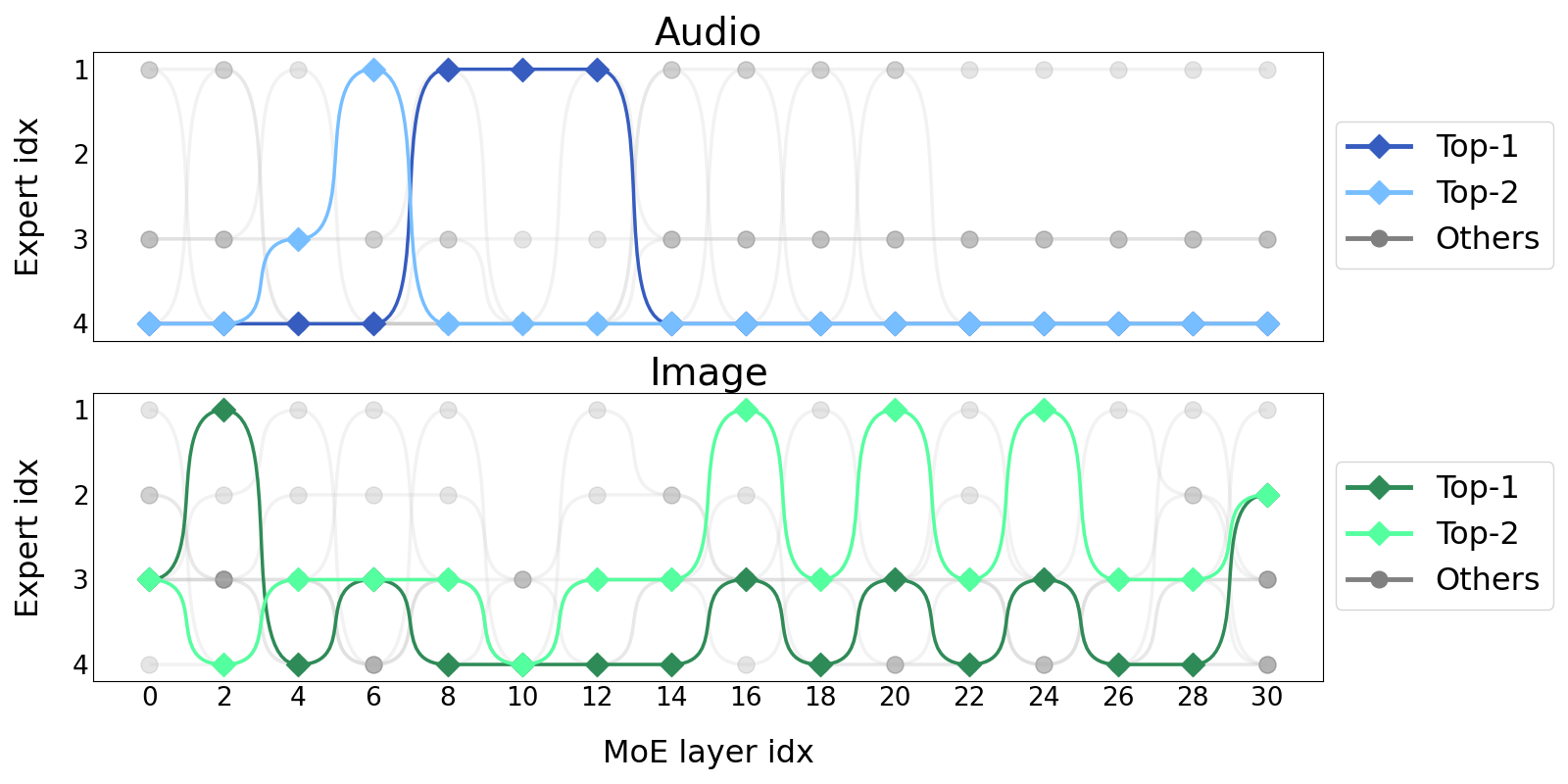}}
\resizebox{0.91\linewidth}{!}{
\includegraphics[width=0.95\linewidth]{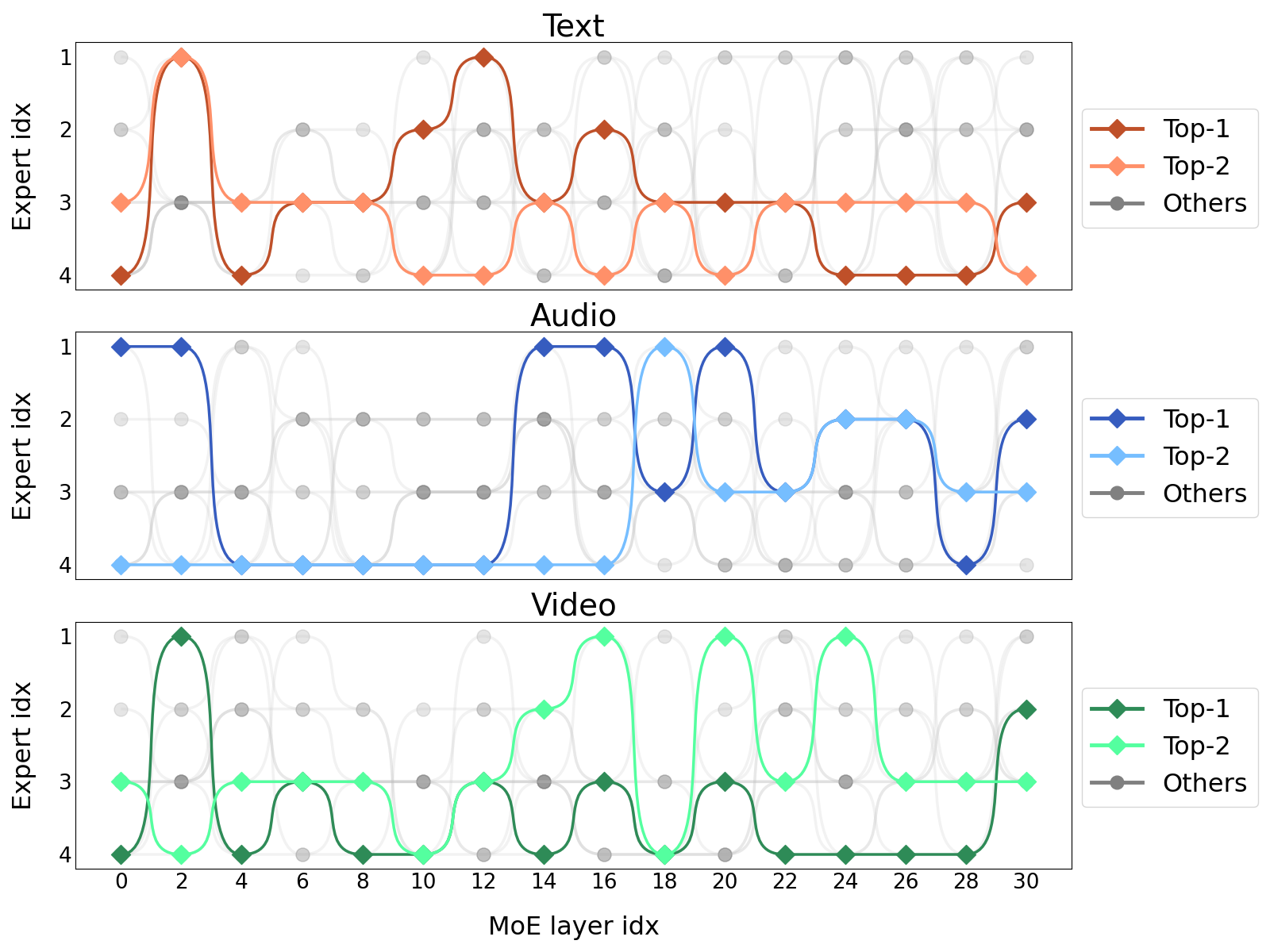}}
 \caption{\textbf{Visualization of activated pathways}. We highlight the top 10 activated pathways. Among them, the non-gray paths represent the top 2 paths, while the gray paths represent the remaining 8 paths. All cross-modality data and the Uni-MoE version are identical to Figures \ref{fig:experts_distribution} and~\ref{fig:modalities_distribution}. Notably, \textit{the expert numbers are not identical to previous experts because we mainly consider the token-level activated path of experts from the first layer to the last one, adopting the PCA to reduce and sort the token-dimension features.} }
\label{fig:activate_ways}
\end{figure}

\textbf{Image-Text Understanding: }
We present experimental results on image-text understanding in Table~\ref{image-text-und} as a sanity check. The best result from our models outperforms all baselines and surpasses the fine-tuning task by an average of 4 points, indicating the enhancement of image and text comprehension from the MoE layer. However, it is also a notable phenomenon that the MoE-Task with all datasets involved shows less capability of image understanding than the one with image data only and the one without long speech data, i.e., MoE-Task1-short and Task2. We hypothesize that this minor decrement in performance is attributed to the mixture of three modality data boosting the understanding of speech with video as well as the image content. Yet, with the presence of long speech data, our models are confused about making the right choice of expert for audio, image, and text modality, which leads to a subtle decline in performance. Overall, \textit{our models show a great ability to interact with other modalities while maintaining the ability of text-image understanding, which makes the method of mixture of experts superior in building multimodal models}. 

\begin{table}[t]
\renewcommand\arraystretch{1.15}
\centering
\caption{\label{video_und}
\textbf{Comparative performance on two zero-shot Video QA benchmarks}. The comparative baselines adopt the same video instruction tuning data, yet employ a more powerful video processing method, e.g., Video-ChatGPT contains more visual tokens by introducing two spatial and temporal tokens. }
\resizebox{\linewidth}{!}{
\begin{tabular}{l|cc|cc}
\hline
\multirow{2}{*}{\textbf{Method}} & \multicolumn{2}{c}{\textbf{ActivityNet-QA}} & \multicolumn{2}{c}{\textbf{MSVD-QA}} \\
\cline{2-5}
    & Accuracy & score & Accuracy & score \\
\hline
{FrozenBiLM } \cite{yang2022zero} & 24.7\% & - & 32.2\% & - \\
{VideoChat } \cite{li2023videochat} & 26.5\% & 2.2 & 56.3\% & 2.8 \\
{LLaMA-Adapter \cite{gao2023llama}} & 34.5\% & 2.7 & 54.9\% & 3.1 \\
{Video-LLaMA } \cite{zhang2023video}& 12.4\% & 1.1 & 51.6\% & 2.5 \\
{Video-ChatGPT \cite{videochatgpt}} & 35.2\% & 2.7 & \textbf{64.9\%} & 3.3 \\
\hline
{Uni-MoE w/ MoE-Task2 epoch1} & 42.7\% & 2.5 & 55.0\% & 3.2 \\
{Uni-MoE w/ MoE-Task2 epoch2} & 42.2\% & 2.5 & 55.3\% & 3.2 \\
{Uni-MoE w/ MoE-Task3 epoch1} & \textbf{42.8\%} & 2.5 & 55.6\% & \textbf{3.4} \\
{Uni-MoE w/ MoE-Task3 epoch2} & 42.7\% & \textbf{2.8} & 56.1\% & 3.4 \\
\hline
\end{tabular}}
\end{table}

\textbf{Video QA: } Table~\ref{video_und} shows the performance of our models on zero-shot tri-modality video QA tasks. To speed up the training and inferring process, we use the simple average pooling representation of randomly extracted 8 frames from a video as the soft video tokens, which often leads to visual information loss. It may achieve lower performance compared to strong video-LLMs such as Video-ChatGPT which focuses more on fine-grained visual representations of a video and its temporal encoding. We also fed LLMs with the audio tokens obtained by the audio or speech encoders and connectors. Experimental results indicate that Uni-MoE trained with MoE-Task3 achieves the best performance on Activity-QA among strong video baselines, surpassing Video-ChatGPT by 7.5\%. Moreover, it also achieves comparative performance on the MSVD-QA dataset. \textit{These findings show that utilizing the MoE approach to integrate the cross-modality reasoning ability, wherein experts concentrate on distinct modalities like audio and image, may effectively improve the overall performance of video understanding.}

\section{Experimental Analysis}

\subsection{Visualization Analysis}

We adopt the method of visualizing the routing distributions and token pathways from MoE-LLaVA to check the specific functions of different experts. For each cross-modality or tri-modality pair, we randomly select 200 samples from the corresponding datasets to draw Figures~\ref{fig:experts_distribution}, \ref{fig:modalities_distribution}, and \ref{fig:activate_ways}. More visualization and analysis can be found in the Appendix.

\textbf{Routing Distributions:} 
In Figure~\ref{fig:experts_distribution}, we present the expert loads (leftmost plot) and the modalities preferences of different experts (four subplots on the right) through MoE-Task3 when encountering all combinations of different modality data. When we fed Uni-MoE with audio-text pairs, experts 2 and 4 almost dominated the workload of almost all tokens, while experts 1 and 3 had barely any loads. A similar trend happens in the case of image-text pairs loading to the MoE layers, expert 2 plays a fairly important role and participates deeply in the processing of image information. It is not until the last layer that expert 3 works more predominantly than other experts, which is not an important case for the analysis. However, when we deliver the video data, the workload of all layers becomes seemingly balanced compared with other circumstances. We also observe distinct behaviours: 1) Expert 1 appears less engaged in token distribution compared to the others, indicating a potential inefficiency. 2) Conversely, Expert 2 demonstrates considerable control over token allocation in the initial layers, which shifts as the model deepens. 3) Expert 4 then begins to increasingly manage the tokens, reflecting a gradual assumption of responsibilities. 4) Expert 3 also contributes to token handling as the model progresses, illustrating collaborative task division among the experts. 
\textit{These behaviours suggest that the experts in Uni-MoE have developed a specific pattern for dividing tasks, particularly notable between Experts 2 and 4. They align with the respective pre-trained specializations in image and audio processing. The provided figures show that Uni-MoE efficiently utilizes expert gates, allocating tokens to the most suitable expert based on their specialized knowledge from fine-tuning tasks. This strategic distribution ensures optimal processing and underscores the model's effective learning and application of task division among experts.}

Moreover, in Figure \ref{fig:modalities_distribution}, we illustrate how different modalities are distributed among the experts in the Uni-MoE model, revealing distinct preferences. Specifically, text tokens are primarily managed by Expert 4 in scenarios involving audio contexts, but shift towards Expert 2 when image contexts are present. In cases where both video and audio data are input, text tokens are more evenly distributed across experts in the latter layers, indicating their role in cross-modality reasoning. For audio and visual tokens, there's a notable pattern of distribution: Expert 4 predominantly handles audio tokens in the initial and final layers, whereas Expert 2 takes over in the middle layers. This alternation highlights their specialized functions in processing different data types. Furthermore, the distribution shifts between text-image and text-audio-video scenarios reveal how image and video tokens are managed. Image tokens are mainly processed by Expert 2, reflecting its image-centric specialization, while video tokens are dispersed across multiple experts, underscoring the complex nature of video data that requires integrated processing from various modalities. In conclusion, \textit{this differentiation in token handling among experts underscores the Uni-MoE model's capacity for strong multimodal interaction and learning, validating its effectiveness in integrating diverse data types including video, speech, text and image.}

\textbf{Token Pathways:} As shown in Figure \ref{fig:activate_ways}, We examine the behaviour of experts at the token level. For all activated pathways, we employ PCA~\cite{Pearson} to obtain the top-10 pathways. Notably, the expert indexes in this figure of token pathways have no strict corresponding relationship with the expert tags in the previous figures of routing distributions. Similar to the result from the routing analysis, the pathway of tokens shows the preference of different modalities in various experts of different MoE layers, which again contributes to a better understanding of the advancement in multi-modal related experts and the behaviour of Uni-MoE in multi-modal learning and inferring. Overall, \textit{Figures \ref{fig:experts_distribution}, \ref{fig:modalities_distribution}, and \ref{fig:activate_ways} present the workflow of our Uni-MoE model at the expert-level, modality-level, and token-level perspectives. The above analysis indicates that Uni-MoE has learned a certain pattern that allows them to divide multiple-modality tasks in a specific manner.}

\begin{table}[t]
\centering
\caption{\textbf{Ablation study about Uni-MoE architectures}. For subtables (a), (b), and (d), we employ the MoE-Task2 to train different Uni-MoE variants with varying expert configurations. In subtable (a), two models include four experts and we set different top-k values; In subtable (b), the top-k value is set to 2 for all model variants, which results in the model with two experts operating as a dense model with the same activated parameter size as that with four experts. The model with one expert is also dense. The identical expert in subtable (c) stems from LLaVA-v1.5-7b and it compares the performance impacts of increasing the number of identical pure experts from four to six. Subtable (d) presents the comparative results of Uni-MoE with various injected ways of MoE layers.}
\label{ablation_architecture}
\text{(a) The value of top-k (Uni-MoE).}
\vspace{2em} 
\resizebox{\linewidth}{!}{
\begin{tabular}{l|cccccc}
\hline
\multirow{2}*{\textbf{Top-k}} & \multirow{2}*{\textbf{A-OKVQA}} & \multirow{2}*{\textbf{OK-VQA}} & \multicolumn{2}{c}{\textbf{ActivityNet-QA}} & \multicolumn{2}{c}{\textbf{RACE-Audio}}  \\
\cline{4-7} 
&&& Accuracy & score & middle & high\\
\hline
1 & 66.46\% & 62.76\% & 42.1\% & 2.5 & 45.13\% & 42.42\% \\
2 & \textbf{66.20\%} & \textbf{63.02\%} & \textbf{42.8}\% & 2.5 & \textbf{46.31}\% & \textbf{43.71\%} \\
\hline
\end{tabular}}

\text{(b) The number of experts (Uni-MoE).}
\vspace{2em} 
\resizebox{\linewidth}{!}{
\begin{tabular}{l|cccccc}
\hline
\multirow{2}*{\textbf{Experts}} & \multirow{2}*{\textbf{A-OKVQA}} & \multirow{2}*{\textbf{OK-VQA}} & \multicolumn{2}{c}{\textbf{ActivityNet-QA}} & \multicolumn{2}{c}{\textbf{RACE-Audio}}  \\
\cline{4-7} 
&&& Accuracy & score & middle & high\\
\hline
1 & 65.68\% & 56.12\% & 41.6\% & 2.8 & 42.59\% & 39.02\% \\
2 & 65.75\% & 62.04\% & \textbf{42.9\%} & 2.5 & 44.64\% & 43.63\%\\
4 & \textbf{66.20\%} & \textbf{63.02\%} & 42.8\% & 2.5 & \textbf{46.31\%}  & \textbf{43.71}\%\\
\hline
\end{tabular}}
\text{(c) The number of identical experts (Pure MoE).}
\vspace{2em}
\resizebox{\linewidth}{!}{
\begin{tabular}{l|ccccc}
\hline
\multirow{2}*{\textbf{Experts}} & \multirow{2}*{\textbf{A-OKVQA}} & \multirow{2}*{\textbf{OK-VQA}} & \multicolumn{2}{c}{\textbf{ActivityNet-QA}} & \multirow{2}*{\textbf{ClothoV2}}  \\
\cline{4-5} 
&&& Accuracy & score &\\
\hline
1 & 65.24\% & \textbf{62.05\%} & 42.5\% & 2.5 & \textbf{23.3\%} \\ 
4 & 64.98\% & 61.67\% & 42.3\% & 2.8 & 21.5\%\\
6 & \textbf{66.81\%} & 61.18\% & \textbf{43.2\%} & 2.6 & 21.8\%\\
\hline
\end{tabular}}
\text{(d) The internal architectures of Uni-MoE.}\\
\resizebox{\linewidth}{!}{
\vspace{1pt}
\begin{tabular}{l|cccccc}
\hline
\multirow{2}*{\textbf{Architecture}} & \multirow{2}*{\textbf{A-OKVQA}} & \multirow{2}*{\textbf{OK-VQA}} & \multicolumn{2}{c}{\textbf{ActivityNet-QA}} & \multicolumn{2}{c}{\textbf{RACE-Audio}}  \\
\cline{4-7} 
&&& Accuracy & score & middle & high\\
\hline
{First-Half} & 65.68\% & 60.96\% & 41.9\% & 2.4 & 38.16\% & 41.14\%\\
{Second-Half} & 63.97\% & 61.33\% & 43.2\% & 2.6 & \textbf{51.39\%} & \textbf{52.69\%}\\
{Interval} & 64.54\% & 61.77\% & \textbf{43.3\%} & 2.5 & 46.17\%  & 46.60\%\\
{All} & \textbf{66.20\%} & \textbf{63.02\%} & 42.8\% & 2.5 & 46.31\%  & 43.71\% \\
\hline
\end{tabular}}
\end{table}

\begin{table*}[t]
\centering

\caption{
\textbf{\label{ablation_training}Ablation study about different training strategies and auxiliary balancing loss~\cite{lepikhin2020gshard}}. All models are trained for one epoch with the same mixed multimodal data from MoE-Task3 or Pure-MoE-Tasks. "mixture(4)" and "pure(4 or 6)" refer to Uni-MoE with four pre-tuned experts from the second training stage and pure MoE with four or six identical MLPs, respectively. The top-k value is set to 2. The ``Source'' represents which specific model the experts (MLP) are from. ``Aux Loss'' refers to the classical balancing loss proposed in GShard~\cite{lepikhin2020gshard}, aiming to encourage giving all experts equal importance. This loss ensures that all experts receive a roughly equal number of training examples. }
\resizebox{0.96\linewidth}{!}{
\begin{tabular}{cccccc|cccccc}
\hline
&\multirow{2}*{\textbf{MoE}} & \multirow{2}*{\textbf{experts}} & \multirow{2}*{\textbf{Source}} & \multirow{2}*{\textbf{Data}} & \multirow{2}*{\textbf{Aux Loss}}  & \multirow{2}*{\textbf{A-OKVQA}} & \multirow{2}*{\textbf{OK-VQA}} & \multicolumn{2}{c}{\textbf{ActivityNet-QA}} & \multirow{2}*{\textbf{ClothoV2}} &  \multirow{2}*{\textbf{Avg.}}\\
\cline{9-10} 
& & & & & & & & Accuracy & score& & \\
\hline
{(a)}&\cmark & mixture(4) & Training Stage 2 & MoE-Task3 & \xmark & 66.20\% & 63.20\% & 42.7\% & 2.5 & 24.7\% & \textbf{49.2\%}  \\
{(a')}&\cmark & mixture(4) & Training Stage 2 & MoE-Task3 & \cmark & 65.23\% & 61.92\% & 42.9\% & 2.5 & 24.3\% & 48.5\%  \\
{(b)}&\cmark & pure(4) & LLaVA-v1.5 & Pure-MoE-Task1 & \xmark & 64.98\% & 61.67\% & 42.1\%& 2.8 & 21.5\% & 47.5\% \\
{(b')}&\cmark & pure(4) & LLaVA-v1.5 & Pure-MoE-Task1 & \cmark & 65.76\% & 61.99\% & 41.9\% & 2.4 & 24.2\% & 48.4\% \\
(c) &\cmark  & pure(6) & LLaVA-v1.5& Pure-MoE-Task1 & \xmark & 66.81\% & 61.18\% & 43.2\% & 2.6 & 21.8\% & 48.2\%\\
(c') &\cmark  & pure(6) & LLaVA-v1.5& Pure-MoE-Task1 & \cmark & 65.24\% & 61.61\% & 42.1\% & 2.7 & 24.5\% & 48.3\%\\
{(d)}&\xmark & single  & LLaVA-v1.5& Pure-MoE-Task1 & \xmark & 65.24\% & 62.05\% & 42.5\% & 2.5 & 23.3\% & 48.2\% \\
\hline
{(e)}&\cmark & pure(4) & LLaMA & Pure-MoE-Task2  & \xmark& 66.55\% & 57.25\% & 41.6\% & 2.8 & 23.8\% & 47.3\% \\
{(f)}&\xmark & single  & LLaMA  & Pure-MoE-Task2 & \xmark & 65.58\% & 56.12\% & 41.3\% & 2.6  & 23.3\% & 46.5\% \\
\hline
\end{tabular}
}
\end{table*}

\begin{table*}[t]
\renewcommand\arraystretch{1.15}
\centering
\caption{\label{more-experts-und}
\textbf{Ablation study of Uni-MoE performances with adding more image-text training data and modality-specific experts}. We only expand the image-text instruction data from LLaVA-150k (MoE-Task 1/2/3) to LLaVA-v1.5-665k (MoE-Task4). For models with \textit{X}B, \textit{X} refers to the size of the language model.  For Uni-MoE with \textit{Y}E, \textit{Y} refers to the number of experts. Names are abbreviated due to space limits. I: Image; T: Text; A: Audio; S: Speech; V: Video. AOK: A-OKVQA \cite{aokvqa}; OK: OK-VQA~\cite{okvqa}; MMB: MMBench~\cite{mmbench}; POPE~\cite{li-etal-2023-evaluating}; SEED:SEED-Bench~\cite{seedBENCH}; MMVet~\cite{yu2023mm}; RAudio: RACE-Audio; AN-QA: ActivityNet-QA~\cite{activitynet}.
``EHSL'' refers to the English High School Listening Test we collected, containing long/short speech types.
$^\ddagger$ indicates that this model employs enormous or unknown data during training.}
\resizebox{1.0\linewidth}{!}{
\begin{tabular}{l|ccccccc|ccc}
\hline
\textbf{Method} & \textbf{AOK} & \textbf{OK} & \textbf{VQAv2} & \textbf{MMB} & \textbf{POPE} & \textbf{SEED} & \textbf{MMVet} & \textbf{RAudio} & \textbf{EHSL} & \textbf{AN-QA} \\
\hline
\textit{Any-Modality Understanding} & & & & & & & & &\\
Macaw-LLM~\cite{lyu2023macaw}  & 1.90\% & 5.70\% & 20.73\% & 3.84\% & - & - & -  & 4.04\%/3.00\% & 0.67\%/2.00\% &  -\\
X-InstructBLIP~\cite{panagopoulou2023x} & 21.52\% & 30.61\% & 37.77\% & 8.96\% & - & - & - & 16.33\%/18.88\% & 0.67\%/2.00\% &-\\
\hline
\textit{Dense Model} & & & & & & & & \\
AnyMAL-70B (I,T,A,V)~\cite{moon2023anymal} & -&42.6\% &64.2\% & -& - & -&-&-&-&-\\
IDEFICS-80B (I,T)~\cite{laurenccon2024obelics} & - & -& 60.0\% & 54.5\% & -  & -& - & - & - &- \\
LLaVA-1.5-7B (I,T)~\cite{liu2023improved} & 70.92\% & 55.09\% & 75.9\% & 72.2\% & 85.9\%  & -& 30.5\% & -& - & -  \\
LLaVA-1.5-13B (I,T)~\cite{liu2023improved} & 73.54\% & 61.93\% & 78.9\% & 73.0\% & 85.9\% & - & \textbf{35.4\%} & -& - & - \\
BLIP-2(FlanT5-xxl)-11B (I,T)~\cite{li2023blip2} & 39.06\% & 53.7\% & 65.0\% & - & 85.3\% & -&-&-&-&-\\
InstructBLIP(Vicuna)-13B (I,T)~\cite{dai2024instructblip} & 58.30\% & 41.02\% & - & - & 50.7\% &25.6\% & - & -& - & -\\
Shikra-13B (I,T)~\cite{chen2023shikra} &- & -& 77.4\% & 58.8\% & - & -& - & -& - & -\\
LLaMA-VID-13B (I,T,V)~\cite{li2023llamavid} & -& -& \textbf{80.0\%} & 66.6\% & 86.0\%  & 62.3\% &- &-&-& \textbf{47.5\%} \\
MiniGPT-4-7B (I,T)~\cite{zhu2023minigpt} & 36.06\% & 29.31\% & - & 23.0\% & -& 42.84\% & 22.1\% & -& - & - \\
LLaMA-Adapter-v2-7B (I,T)~\cite{gao2023llama} &-&-&-&39.5\% & -&-&31.4\% &-&-&-\\
Qwen-VL-7B$^\ddagger$ (I,T)~\cite{qwen} & - & 58.6\% & 78.8\% & 68.2\% & - & 56.3\% & - & -& - & - \\
MobileVLM-3B (I,T)~\cite{chu2023mobilevlm} & - & - & - & 59.6\% & 84.9\%  & - & - & - & - & -\\
LLaVA-Phi-3B (I,T)~\cite{zhu2024llava_phi} & -& -& 71.4\% & 59.8\% & 85.0\% & -&28.9\% & - & - & - \\
{Single-Modality-Expert-Task2} (I,T) & 67.07\% & 62.91\% & 75.18\% & 71.26\% & 84.7\% & 60.63\%& 27.8\% &-&-&- \\
{Single-Modality-Expert-Task5} (I,T,S) & 58.86\% & 56.01\% & 67.35\% & 65.80\% &62.48\%  &53.50\%  &26.9\% &  30.78\%/24.90\% & 9.33\%/12.00\% & -\\
{Single-Modality-Expert-Task6} (I,T,S) & 58.69\% & 57.77\% & 68.74\% & 67.53\%  &65.84\%  &55.21\% &25.6\%  &32.59\%/29.02\% & 18.67\%/8.00\%&-\\
\hline
\textit{Sparse Model}  & & & & & & & & \\
MoE-LLaVA-1.6B×4-Top2~\cite{moeLlava} & 63.8\% & 59.9\% & 74.1\% & 69.1\% & 85.7\% &61.8\% & 28.0\% & - & -&- \\
MoE-LLaVA-2.7B×4-Top2~\cite{moeLlava} & 68.34\% & 62.10\% & 75.4\% & 70.0\% & 85.5\% & 62.3\% & 31.2\% & - & -&- \\

{Uni-MoE w/ MoE-Task1 (4E)} (I,T,S) & 61.22\% & 57.63\% & 68.42\% & 68.15\% &76.67\% & 56.58\% &30.1\% & 47.08\%/47.08\% & 41.33\%/36.00\% & - \\
{Uni-MoE w/ MoE-Task2 (4E)} (I,T,S,V)& 65.07\% & 62.10\% & 73.87\% & 70.50\% & 85.43\%  &60.54\% &28.2\% & 49.65\% /49.37\% & 42.00\%/48.00\% & 42.2\% \\
{Uni-MoE w/ MoE-Task3 (4E)} (I,T,A,V) & 64.28\% & 61.96\% & 73.87\% & 69.82\% & 86.10\%  &59.16\% &31.7\% & -& -& 42.8\% \\
{Uni-MoE w/ MoE-Task4 (4E)} (I,T,S,V) & 70.22\% & 66.02\% & 76.4\% & 73.2\% & 86.0\%& 63.4\% & 32.6\% & \textbf{64.21\%}/\textbf{64.72\%} & 48.00\%/\textbf{58.67\%} & 45.6\%\\
+ Aux\_loss & 69.61\% & 66.13\%& 76.0\%& 72.6\%&  85.0\%  & 63.3\% & 31.7\% & 63.86\%/64.24\%& \textbf{50.00\%}/54.00\% & 45.2\%\\
{Uni-MoE w/ MoE-Task4 (8E)} (I,T,S,V) &  70.0\% & 66.2\% & 76.6\% & 73.0\% & 86.2\% & 63.3\% & 32.5\% & 63.75\%/61.56\% & 48.67\%/46.0\% & 46.0\% \\
+ Aux\_loss & \textbf{70.7\%} & \textbf{66.4\%} & 76.7\% & \textbf{73.2\%} & \textbf{86.3\%} & \textbf{63.4\%} & 32.8\% & 62.33\%/64.18\% & 42.00\%/50.00\% & 46.4\%\\
\hline
\end{tabular}}

\end{table*}

\subsection{Ablation Study}
\label{ablation_study}

\textbf{Comparative Analysis of Uni-MoE and Dense Models}.
In previous Tables~\ref{automatic_speech_image}-\ref{video_und}, we compare the performances of dense models (Single-Expert-Tasks) and Uni-MoE (w/ MoE-Tasks). The experimental results show that \textit{1) The performance of Uni-MoE is consistently better than dense models on almost all evaluation benchmarks.} By comparing the performances of Single-Expert-Modality-Task6 and Uni-MoE w/ MoE-Task1 on speech-image, long speech, and image-text performances, where they use the same type of training datasets, we can see that Uni-MoE achieves better performances on all evaluation benchmarks, especially on the long speech understanding tasks. \textit{2) After training on unbalanced mixed-modality data, Uni-MoE exhibits less performance bias.} For instance, compared to the larger performance drop of Single-Modality-Expert-Task6, Uni-MoE trained with MoE-Task1 shows less performance degradation on short speech-image (in Table \ref{automatic_speech_image}) and text-image understanding (in Table~\ref{image-text-und}) tasks. It also improves the performances on the long speech understanding benchmarks including RACE-Audio and English High School Listening
Test, wherein long speech training data accounts for a small proportion of the training data. 
\textit{3) Uni-MoE shows better generalization than dense models for out-domain inputs when they are trained on the same types of mixed cross-modality data.} Compared to Single-Expert-Modality-Task7, Task6 introduces the mixed data of text-image instruction and short speech-image during training. Introducing more types of multimodal data for dense models lowers the performance on long speech and does not enhance the performance on out-domain evaluation benchmark MMBench-Audio. However, when more data types are introduced, Uni-MoE not only maintains the performance in long speech understanding but also improves its performance in extrapolated three-modal input scenarios. \textit{Hence, these finding suggests the better ability of Uni-MoE to combine generalization on complex multimodal data than dense models.}

\textbf{Impact of the Value of Top-k:} Our ablation study, detailed in Table \ref{ablation_architecture} (a), investigates the effect of varying the number of activated experts (Top-k) while we set the total number of experts to be identical. We observed that increasing the number of activated experts from one to two enhances model performance, indicating that activating more experts can significantly improve the efficiency of Uni-MoE. The increasing performance also suggests the collaboration ability of modality-specific experts in our model. It is identical to the visualization analysis of Uni-MoE in the bottom part of Figure~\ref{fig:experts_distribution}, where it uses more experts at each layer to handle video content.  Consequently, we have set the optimal number of activated experts at two to maximize performance across various cross-modality benchmarks in Tables~\ref{automatic_speech_image}- \ref{video_und}.

\textbf{Scaling up the Number of Experts:} We investigate variations in expert numbers while maintaining a constant count of activated experts, detailed in Table \ref{ablation_architecture} (b) and (c). The results illustrate that Uni-MoE utilizing a greater number of sparse experts surpass the performance of the dense expert configurations, particularly excelling in long speech-text scenarios. This enhancement is attributed to the employment of two specialized experts, specifically trained on Single-Modality-Tasks 2 and 3, showcasing significant advancements in visual and speech tasks as demonstrated in Tables \ref{automatic_speech_image} and \ref{image-text-und}. Analysis of routing distributions further confirms the critical role of these single-modality trained experts in their respective fields. For Uni-MoE, we find that scaling number of experts to four can achieve better comprehensive performance on different modality.
\textit{Conversely, as indicated in Table \ref{ablation_architecture} (c), employing more standard experts from previous configurations without increasing the number of active experts leads to marginal improvements in overall performance. This observation underscores the necessity of strategic expert selection and the effectiveness of sparse expert configurations.}

\begin{figure*}[t]
    \centering
    \includegraphics[width=0.96\textwidth]{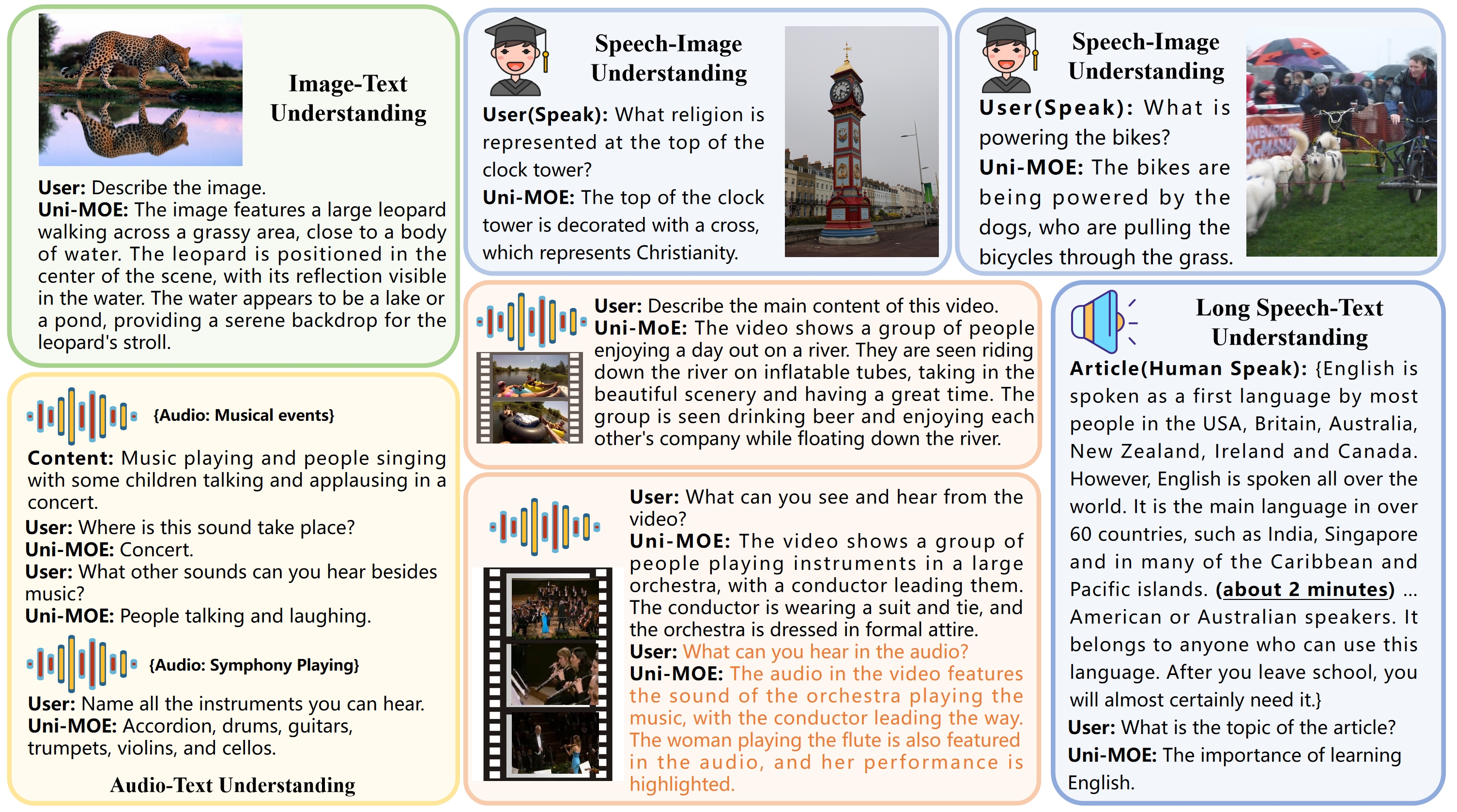}
    \caption{\textbf{An illustration of various cases generated by Uni-MoE}. Interestingly, Uni-MoE trained with MoE-Task3 could understand the audio content from the video, while the instruction tuning data almost does not contain related data. It also could understand long speech from real English listening tests of high school students.}
    \label{fig:case_figure}
\end{figure*}

\textbf{Analyzing the Architectures of Uni-MoE:} 
In Table \ref{ablation_architecture} (d), we evaluate four configurations of MoE architecture within the Uni-MoE framework. The "First-Half" configuration applies MoE layers exclusively to the initial segment of the model, maintaining a conventional dense structure in the latter half. Conversely, the "Second-Half" setup incorporates MoE layers in the latter segment while preserving a dense architecture in the initial segment. The "Interval" configuration intersperses MoE and dense layers throughout the model. Lastly, the "All" configuration converts all layers to sparse MoE layers. We observe  that fully converting to MoE ("All") does not exactly lead to superior performance and additionally incurs extended training durations when compared with other configurations. Notably, the "Interval" layout demonstrates the highest average efficacy across all tasks, establishing itself as the most effective architecture among those tested. Furthermore, positioning MoE layers in the latter half of the model significantly enhances the model's capacity for understanding lengthy speech segments, outperforming the second-best configuration by 5\% and 6\% than middle and high complexity categories, respectively. \textit{The above analysis presents that we still need to explore more robust and efficient MoE architectures in building larger MLLMs.}

\textbf{Effectiveness of Training Strategy:} 
We examine the impact of different training strategies on model performance by presenting three distinct model variants in Table \ref{ablation_training}. The comparative analysis between model variant (a) and variants (b), (c) and (d) first reveals that the tri-phase training approach employed in the model (a) facilitates noticeable enhancements across a range of multimodal benchmarks. This underscores the benefit of incorporating specialized training phases for cross-modality data, affirming the strategic advantage of engaging training experts in the process. Further analysis shows that training MoE with identical configurations (see pure models in Table~\ref{ablation_training}) could result in a negligible performance increase compared to a single-expert approach, as evidenced by models (b), (c), (d), (e), and (f). 
Interestingly, the performance of models can vary when the number of expert sources is increased, according to the unstable performances of adding identical experts from LLaMA or LLaVA. Experts trained on multimodal data perform better when experts work together. However, integrating an auxiliary balancing loss is a potential solution to mitigate these inconsistencies and stabilize performance. Moreover,
visual illustration in Figures \ref{fig:experts_distribution_pure} and \ref{fig:modalities_distribution_pure} (presented in the Appendix) highlight the homogeneity among the experts in model (b), suggesting a need to improve task allocation diversity and expert differentiation. 
\textit{Overall, initiating single-modality training proves advantageous for transitioning models from dense to sparse structures, as demonstrated by our approach. This strategy enhances initial model efficiency, facilitating a more effective and rapid adaptation in subsequent training phases, thereby validating our training strategy's effectiveness}.

\textbf{Analysis of Balancing auxiliary Loss:} The classical balancing loss introduced in Gshard~\cite{lepikhin2020gshard} encourages giving all experts equal importance. This loss ensures that all experts receive a roughly equal number of training examples. In this paper, we also explore the effect of auxiliary balancing loss on model performance. As the experimental results are shown in Tables~\ref{ablation_training} and \ref{more-experts-und}, our findings indicate that: \textit{1) Employing an auxiliary loss consistently enhances both the synergy among experts and the overall performance of the model across various modalities, when applied to the model with identical (pure) expert; 2) In our Uni-MoE model, as the number of experts expands to eight, the auxiliary loss shows its effectiveness to facilitate expert collaboration. This improvement is primarily attributed to the expanded routing search space resulting from the increased number of experts. The introduction of auxiliary loss at this stage plays a role in optimizing the selection of expert combinations, thereby fully activating the capabilities of the experts}.

\textbf{Comparisons of Uni-MoE and Dense Large Visual-Language Models (LVLMs)}. The results presented in Table~\ref{more-experts-und} indicate that the larger LVLMs, LLaVA-v1.5-13B and LLaMA-VID-13B, outperform Uni-MoE on image-text benchmarks. This superior performance can be attributed to two primary factors. First, these models focus exclusively on visual and language data during training, and they activate more parameters (13B) during inference compared to Uni-MoE's 11B, enhancing their effectiveness in image-text tasks. Second, LLaMA-VID benefits from the inclusion of 703K video data points used in pre-training video-to-language connections, a dataset not utilized by Uni-MoE, giving it an edge in evaluations like ActivityNet-QA. Additionally, for LLaMA-VID, the number of sampling frames is larger than Uni-MoE. Despite these differences, Uni-MoE still excels in image-text comprehension over similar-sized MLLMs when the same image-text instruction data is incorporated. Moreover, it surpasses well-known unified multimodal models such as X-InstructBLIP and Macaw-LLM in other modal capabilities. \textit{Interestingly, for Uni-MoE, adding image-text data enhances its performance in video understanding, which inspires us to further enhance the performance of video-LLMs by introducing additional image-text data}.

\subsection{Case Study}
In our analysis depicted in Figure~\ref{fig:case_figure}, we present the performance of Uni-MoE trained with MoE-Task3 on different modalities. We can see that Uni-MoE could understand any cross-modality inputs and recognize the real long speech produced by humans and speech content in the video outside the training data, as the bottom examples are shown in Figure~\ref{fig:case_figure}. Combining previous quantitative evaluation results and generated cases, we conclude that Uni-MoE shows its power to handle various modalities, trained with small yet diverse mixed multimodal data. Interestingly, Uni-MoE trained with MoE-Task3 could understand the audio content from the video, while the instruction tuning data almost does not contain related data. It indicates the robustness and generalization of utilizing MoE to handle various modalities compared to previous dense baselines such as X-InstructBLIP, which was trained on multiple modalities of data yet achieves inferior performance on speech-image or video understanding tasks.


\section{Conclusion}
In this research, we ventured into expanding the capabilities of large multimodal models through the integration of Mixture of Experts (MoE) architecture, resulting in the development of Uni-MoE. We devised and implemented a novel three-phase training strategy specifically tailored to enhance both the stability and generalization performance of Uni-MoE across a broad spectrum of multimodal scenarios. This approach was rigorously tested against a variety of challenging benchmarks, encompassing both cross-modality comprehension and long-form speech and video reasoning tasks. Our findings reveal that Uni-MoE not only surpasses existing benchmarks in cross-modality and mixed-modality frameworks but also outperforms conventional MoE models equipped with identical experts. Furthermore, our detailed ablation studies validate the significance of our tailored three-phase training strategy, highlighting its critical role in enhancing the robustness and adaptability of MoE-based MLLMs when confronted with diverse multimodal datasets. We hope our work could spark the research of utilizing the MoE architecture to scale up MLLMs.


%



\ifCLASSOPTIONcompsoc
  \section*{Acknowledgments}
\else
  \section*{Acknowledgment}
\fi

The authors would like to thank the efforts of editors and reviewers for checking our work.





\bibliographystyle{IEEEtran}
\bibliography{paper}

\begin{thebibliography}{10}
\bibitem{tts}
Microsoft, ``Text to speech an ai speech feature that converts text to lifelike speech.''
\end{thebibliography}




%




\vspace{-1cm}

\begin{IEEEbiography}[{\includegraphics[width=1in,height=1.25in,keepaspectratio]{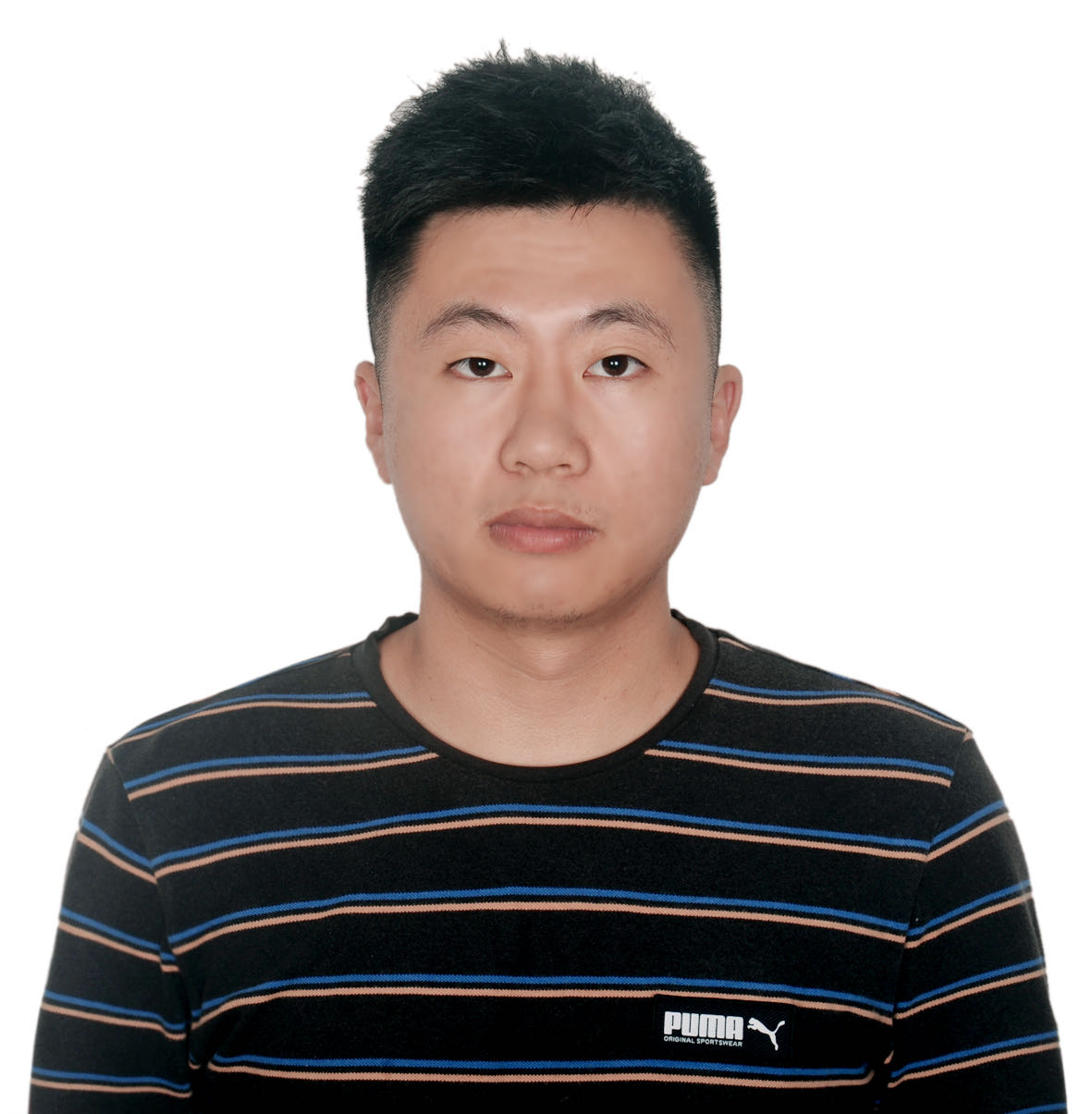}}]{Yunxin Li}
received the M.S. and B.S. degrees from the Harbin Institute of Technology, in 2019 and 2022. He is currently pursuing the Ph.D. degree with the School of Computer Science, Harbin Institute of Technology, Shenzhen. His research interests include large multimodal models, cross-modal reasoning, and natural language generation.
\end{IEEEbiography}

\vspace{-1cm}

\begin{IEEEbiography}[{\includegraphics[width=1in,height=1.25in,keepaspectratio]{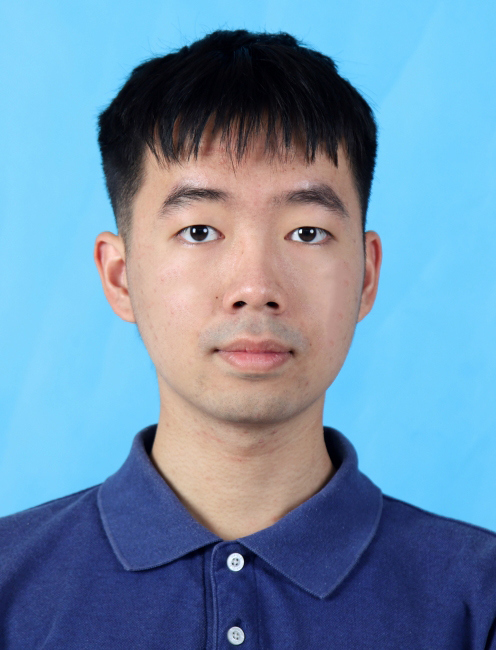}}]{Shenyuan Jiang} earned a Bachelor of Science degree in Computer Science from Harbin Institute of Technology, Shenzhen campus, in 2022. He is now pursuing a Master's degree at the School of Computer Science in Harbin Institute of Technology, Shenzhen. Jiang's research interests are centered around the development and enhancement of multimodal large language models.
\end{IEEEbiography}

\vspace{-1cm}

\begin{IEEEbiography}[{\includegraphics[width=1in,height=1.25in,clip,keepaspectratio]{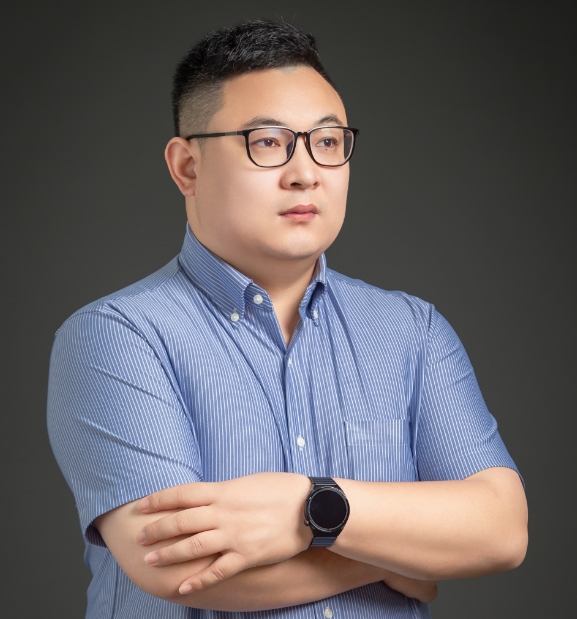}}]{Baotian Hu}
received the M.S. and Ph.D. degrees in computer science from the Shenzhen Graduate School of Harbin Institute of Technology, Shenzhen, China, in 2012 and 2016, respectively. He is currently an Associate Professor with the School of Computer Science and Technology, Harbin Institute of Technology, Shenzhen. His current research interests include deep learning and its application in
natural language processing and vision-language information processing.
\end{IEEEbiography}

\vspace{-1cm}

\begin{IEEEbiography}[{\includegraphics[width=1in,height=1in,clip,keepaspectratio]{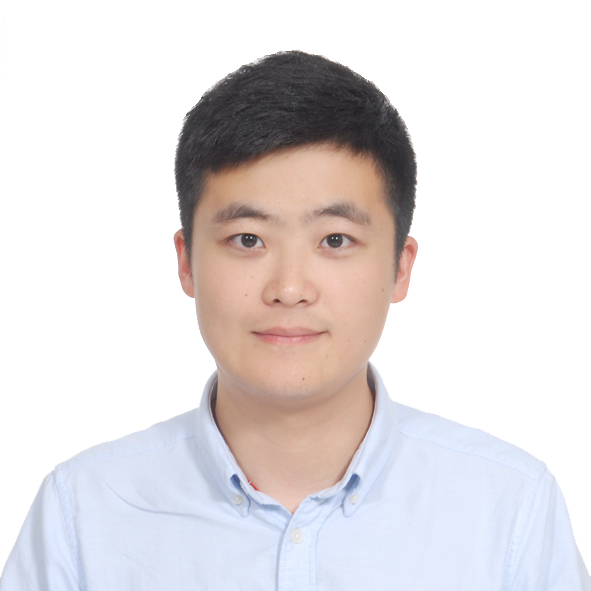}}]{Longyue Wang}
is a Senior Researcher with the Tencent AI Lab, Shenzhen, China. He received his Ph.D. degree from Dublin City University in 2018. Longyue has studied and practiced in a broad field of Artificial Intelligence especially on Large Language Model, Multimodal, Language Agent, Natural Language Processing, Machine Translation, Deep Learning and AI for Science. He has published 60 papers in leading NLP journals and conferences.
\end{IEEEbiography}

\vspace{-1cm}
\begin{IEEEbiography}[{\includegraphics[width=1in,height=1.25in,clip,keepaspectratio]{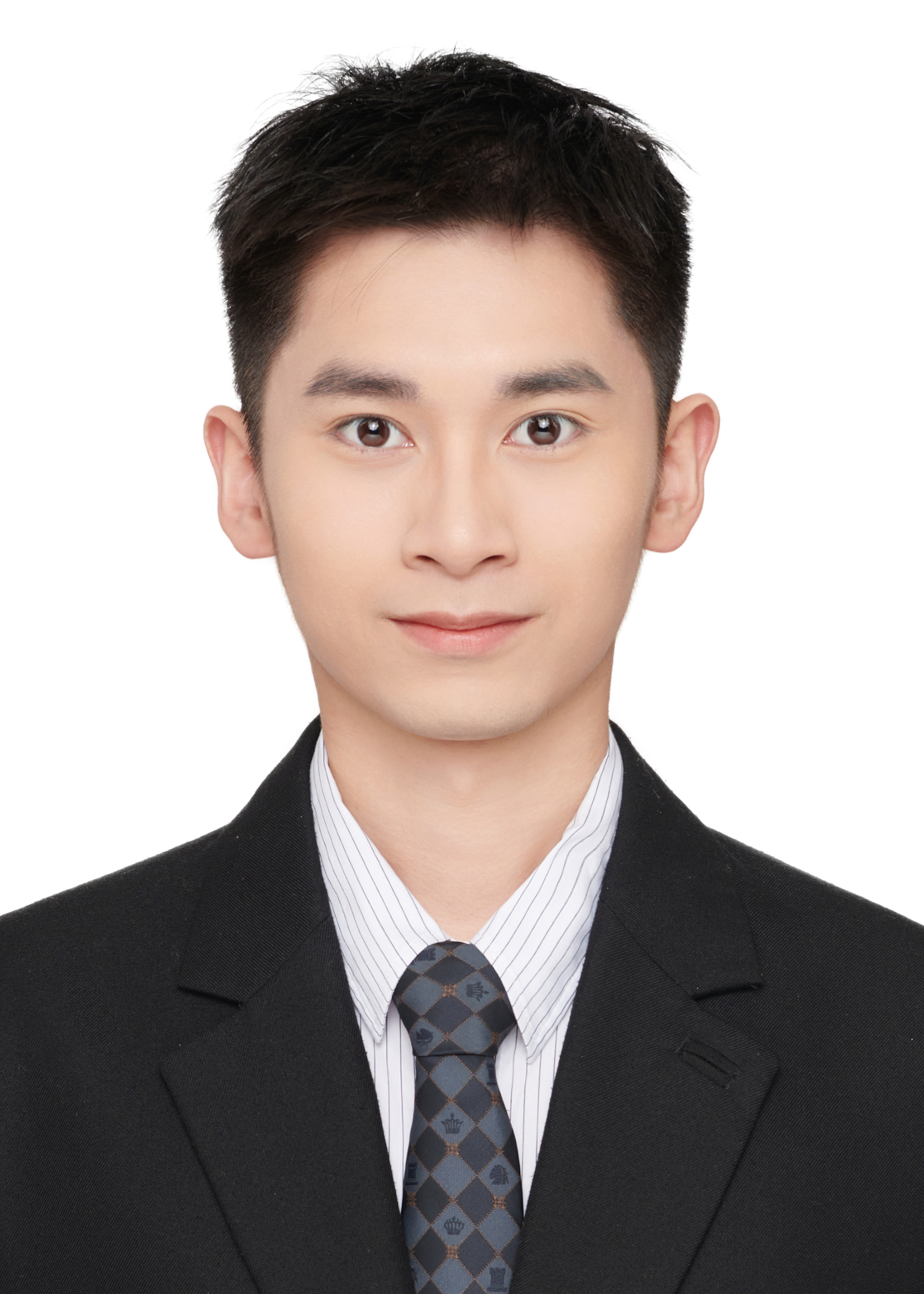}}]{Wanqi Zhong}
is an undergraduate student majoring in Computer Science and Technology at Harbin Institute of Technology, China. His current research interests include computer vision and deep learning, particularly focusing on vision-central general perception.
\end{IEEEbiography}

\vspace{-1cm}
\begin{IEEEbiography}[{\includegraphics[width=1in,height=1.10in,clip,keepaspectratio]{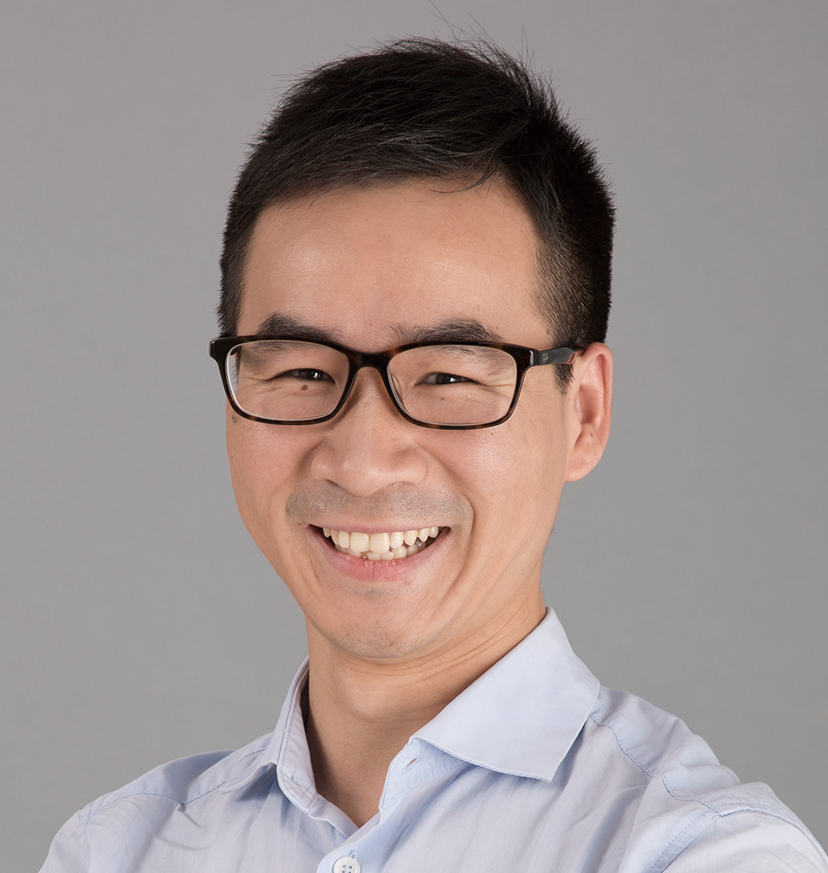}}]
{Wenhan Luo} (Senior Member, IEEE) is currently an Associate Professor at the Hong Kong University of Science and Technology. Prior to that, he worked as an Associate Professor at Sun Yat-sen University and as a research scientist for Tencent and Amazon. He has published over 80 papers in top conferences and leading journals. He also has served as a reviewer, senior PC member, and Guest Editor for several prestigious journals and conferences. He received Ph.D. degree from Imperial College London in 2016, M.E. degree from the Institute of Automation, Chinese Academy of Sciences in 2012, and B.E. degree from Huazhong University of Science and Technology in 2009.
\end{IEEEbiography}

\vspace{-1cm}
\begin{IEEEbiography}[{\includegraphics[width=1in,height=1.25in,keepaspectratio]{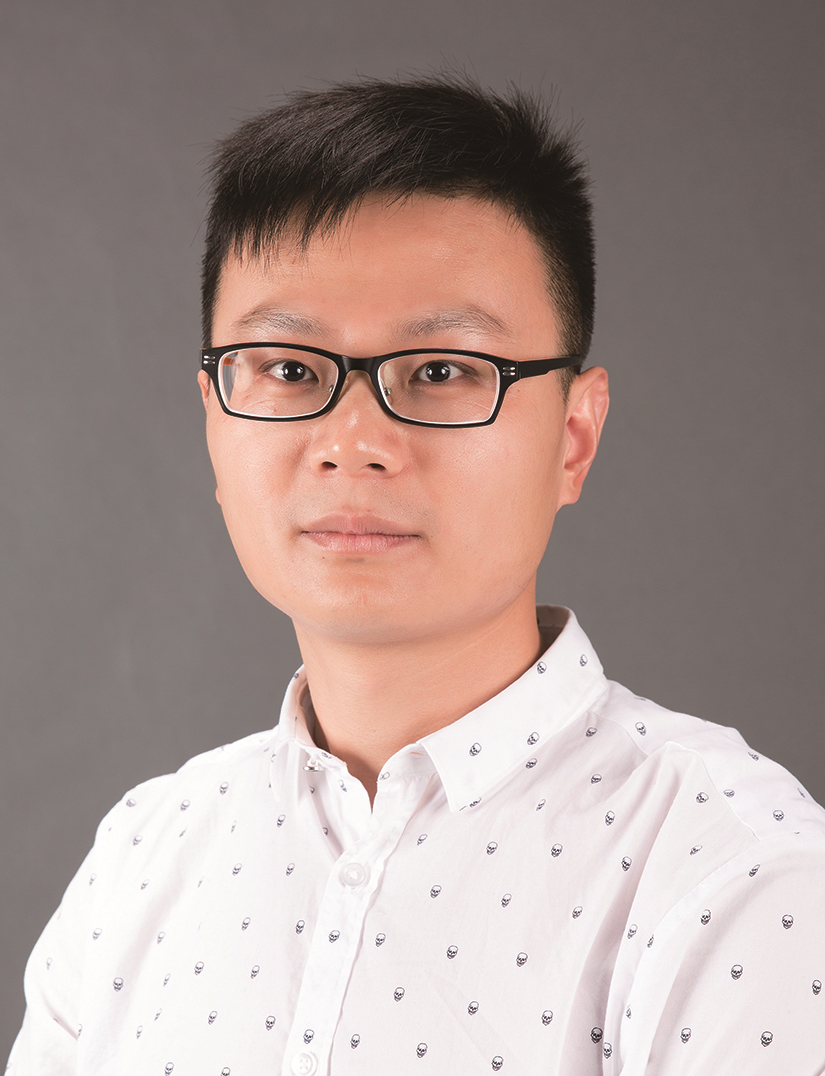}}]{Lin Ma}
received the B. E., and M. E. degrees from Harbin Institute of Technology, Harbin, China, in 2006 and 2008, respectively, both in computer science. He received his Ph.D. degree in the Department of Electronic Engineering at the Chinese University of Hong Kong (CUHK) in 2013. He is now a Researcher with Meituan, Beijing, China. His current research interests lie in the areas of deep learning and multi-modal learning, specifically for image and language.
\end{IEEEbiography}

\vspace{-1cm}
\begin{IEEEbiography}[{\includegraphics[width=1in,height=1.25in,keepaspectratio]{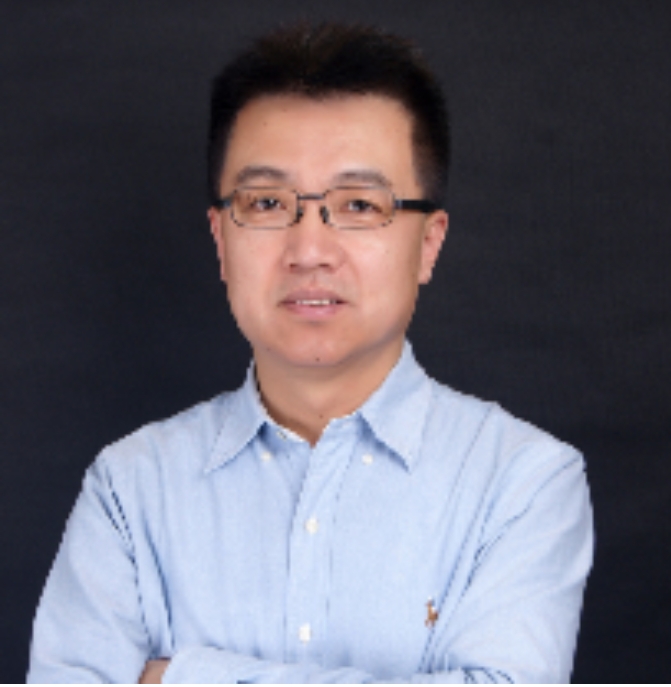}}]{Min Zhang}
received the Ph.D., M.S., and B.S. degrees in Computer Science and Technology at the Harbin Institute of Technology, in 1991, 1994, and 1997, respectively.
He is currently a Professor at the School of Computer Science and Technology, Harbin Institute of Technology, Shenzhen. His research interests include multimodal reasoning, large models, and natural language generation.
\end{IEEEbiography}






\appendices
\section*{Appendix}
\label{sec:appendix}

\subsection*{Comparative Visualization Analysis of Uni-MoE (4 experts) trained with MoE-Task2 and Pure-Task1}
In this section, we present the routing distributions and token pathways of MoE-Task2 and Pure-MoE-Task1 on five selected combinations of multi-modal data(Image-Text, Audio-Text, Video-Text, Image-Audio, Video-Audio-Text), each figure is shown utilizing 200 of data pair samples. These routing distributions are based on the training up to one epoch.

\begin{figure*}[t]
\centering
\includegraphics[width=1.0\linewidth]{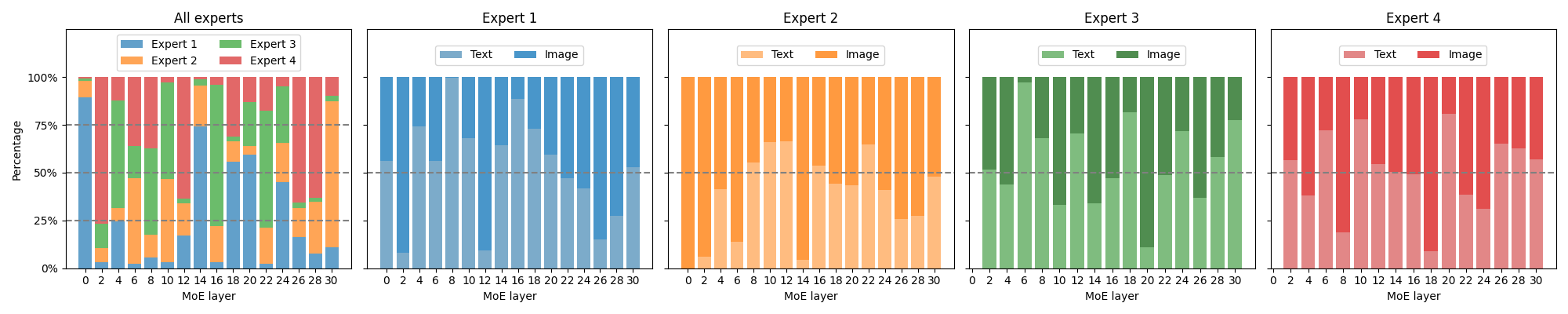}
\includegraphics[width=1.0\linewidth]{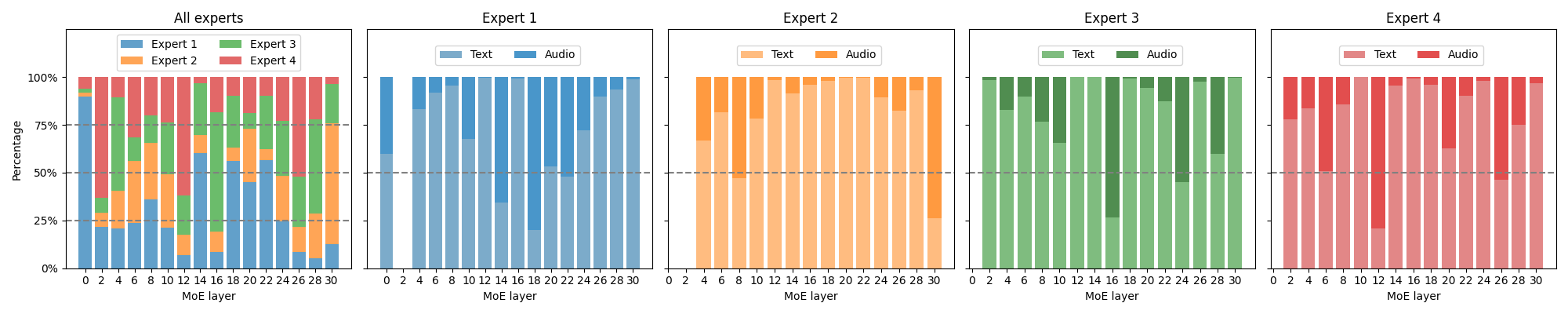}
\vspace{2pt}
\includegraphics[width=1.0\linewidth]{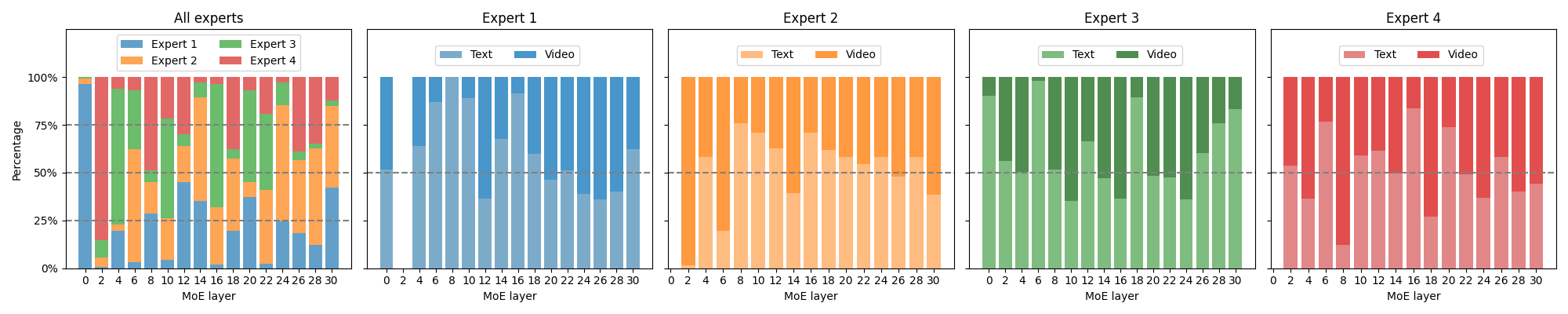}
\includegraphics[width=1.0\linewidth]{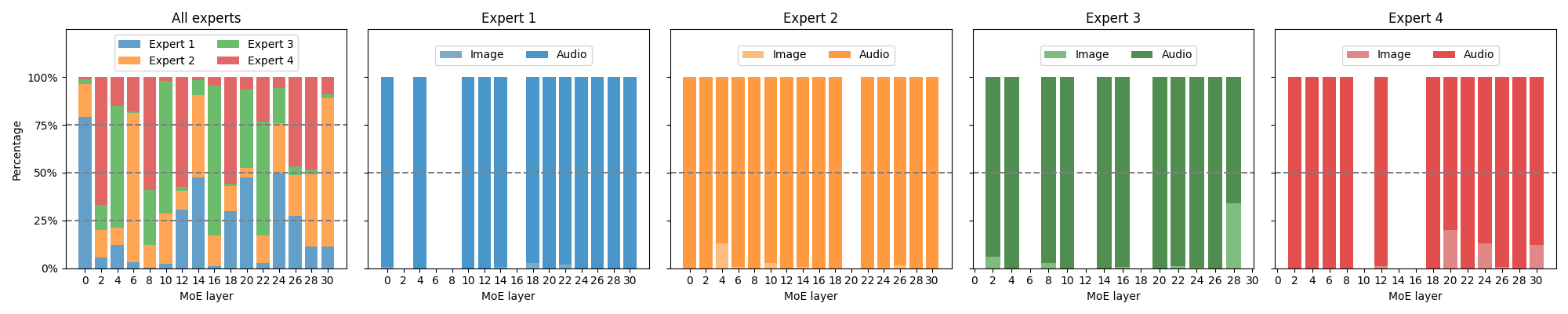}
\vspace{2pt}
\includegraphics[width=1.0\linewidth]{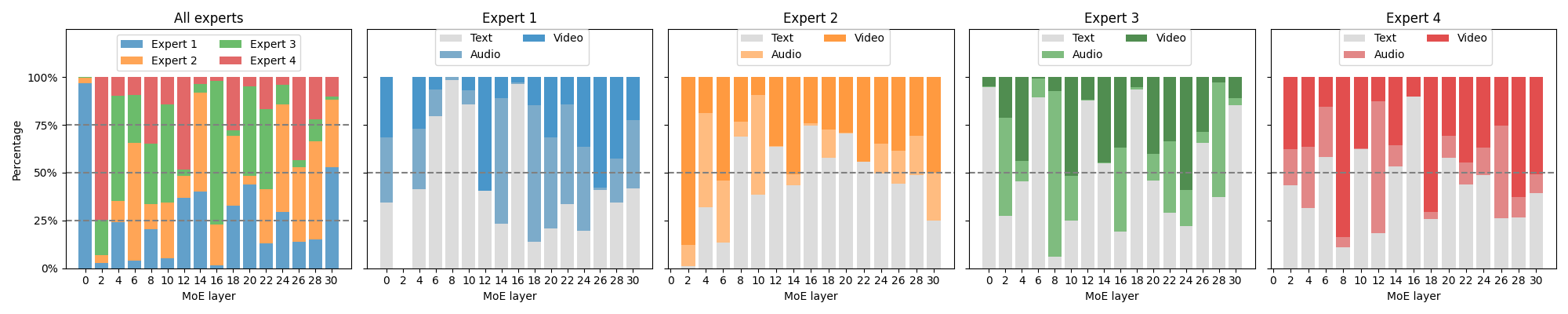}
\caption{Distribution of expert loadings and expert preferences on \textbf{Pure-MoE-Task1.} For different cross-modality data pairs, different experts from different layers have a high degree of consistency.  we fail to observe the modalities for which different experts are primarily responsible. This may be attributed to the fact that experts are initially identical.}
\label{fig:experts_distribution_pure}
\end{figure*}

\begin{figure*}[t]
\centering
\includegraphics[width=1.0\linewidth]{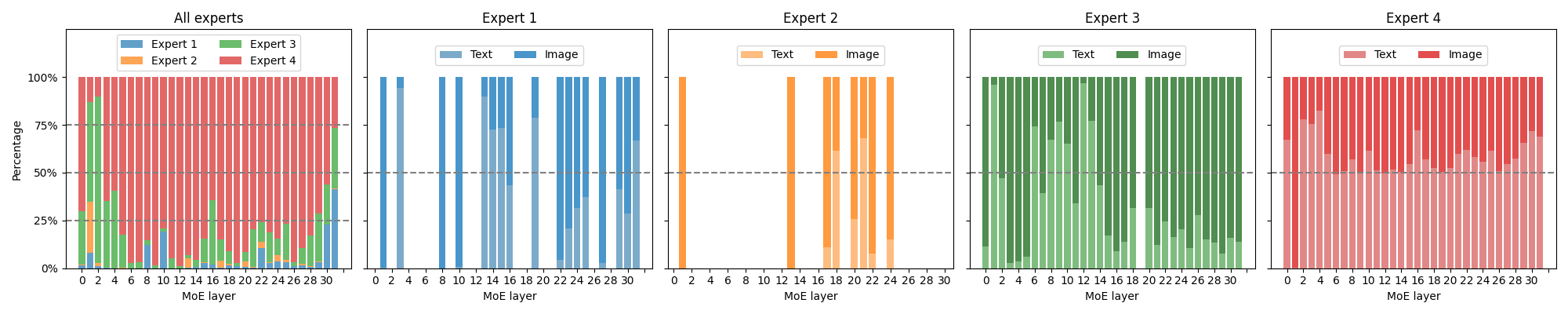}
\vspace{2pt}
\includegraphics[width=1.0\linewidth]{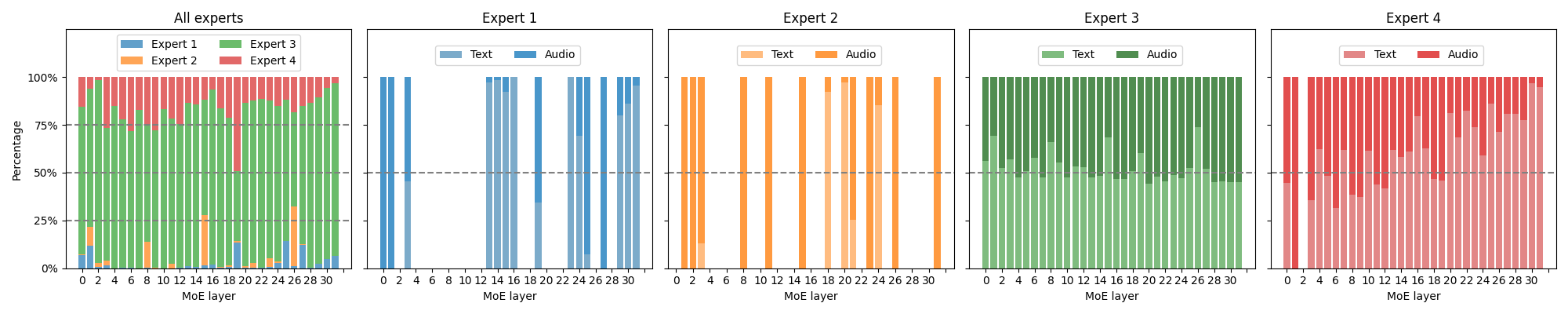}  
\vspace{2pt}
\includegraphics[width=1.0\linewidth]{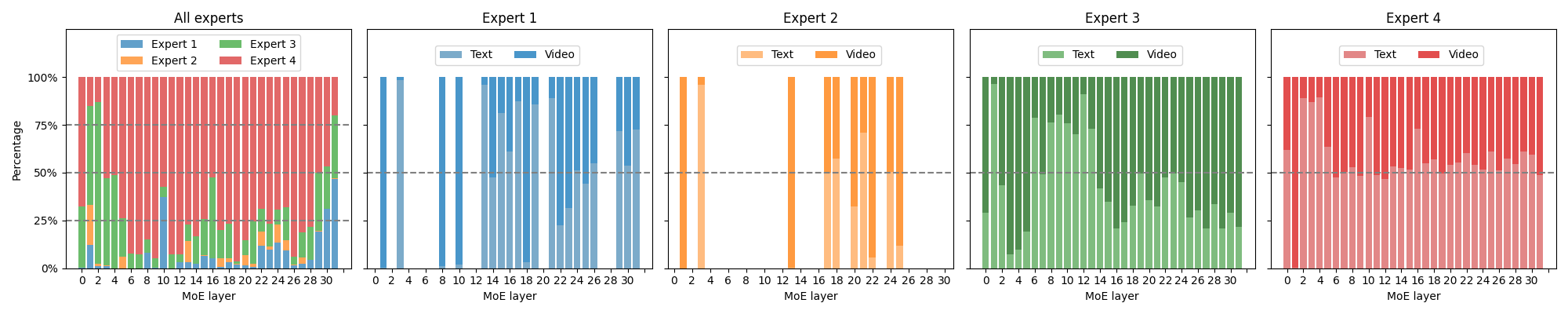}
\vspace{2pt}
\includegraphics[width=1.0\linewidth]{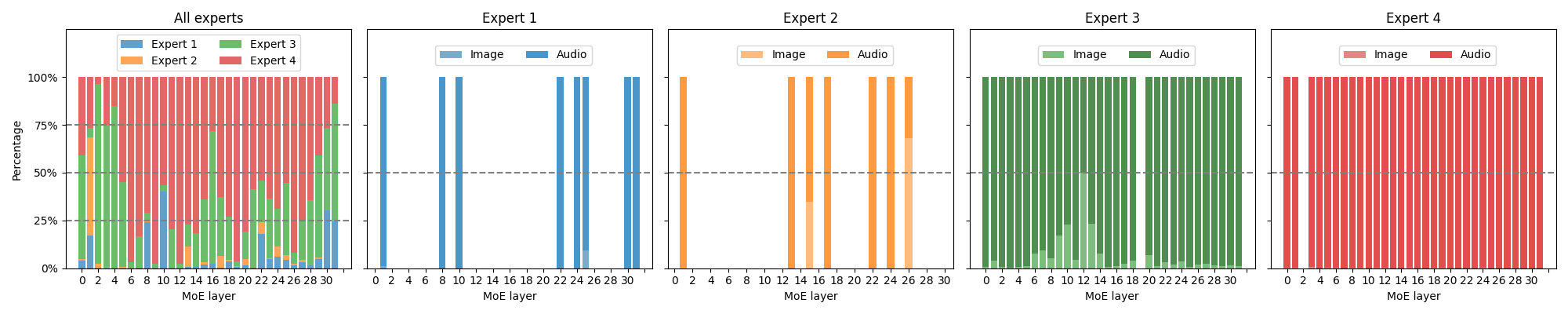}
\vspace{2pt}
\includegraphics[width=1.0\linewidth]{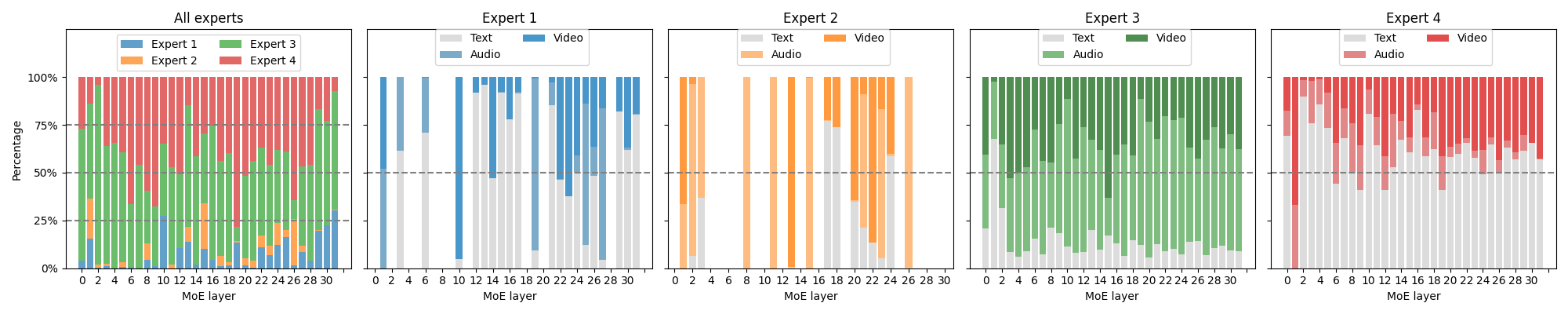}
\caption{Distribution of expert loadings and expert preferences on \textbf{MoE-Task2.}}
\label{fig:experts_distribution_speech}
\end{figure*}

\textbf{Routing Distributions}.
In our study, we conducted an ablation analysis by training a model called Pure-MoE-Task1. This model's performance did not meet the levels exhibited by other Uni-MoEs. The routing distribution for Pure-MoE-Task1, as illustrated in Figure~\ref{fig:experts_distribution_pure}, shows notable differences compared to MoE-Task3 presented in Figure \ref{fig:experts_distribution}. Specifically, Pure-MoE-Task1 demonstrated a relatively balanced distribution in terms of both expert loads and their preferences for different modalities.
Conversely, MoE-Task3 and other Uni-MoE models, such as MoE-Task2 (illustrated in Figure \ref{fig:experts_distribution_speech}), exhibited distinct preferences among the experts that were fine-tuned during the single-modality optimization phase. For example, MoE-Task2 includes four experts, each fine-tuned for a different purpose: Expert 1 was trained using an audio-relevant dataset derived from an image dataset, Expert 2 was adapted from the fine-tuned LLaVA model's MLP layers, Expert 3 was developed for long speech training tasks, and Expert 4 was optimized for image-related tasks with textual information. The data suggest a strong relationship between the preference of an expert and the single-expert training stage. Specifically, during scenarios involving audio features, the workload of Expert 3, which was fine-tuned for long speech tasks, significantly increases. Similarly, with image inputs, Expert 4's workload exceeds that of all other experts. When handling video files containing both visual and audio content, the workload is almost evenly distributed between Experts 3 and 4, highlighting their significant roles in MoE training for task-specific contexts.

Overall, our findings suggest that the strategy of fine-tuning individual experts effectively transfers the capabilities of single modalities to enhance the performance of sparse Large Language Models (LLMs) across various tasks. While utilizing identical experts across all modules does not distinctly separate their functions, it inadvertently reveals unique patterns for each expert, which in some cases, proves to be effective. This characteristic pattern underscores the innovative approach of using Mixture of Experts in the Multimodal Large Language Model (MLLM) field. Future research should aim to further leverage the potential of MoE within MLLM to enhance its application and effectiveness.

\begin{figure}[p]
\centering
\resizebox{\linewidth}{!}{
\includegraphics[width=0.95\linewidth]{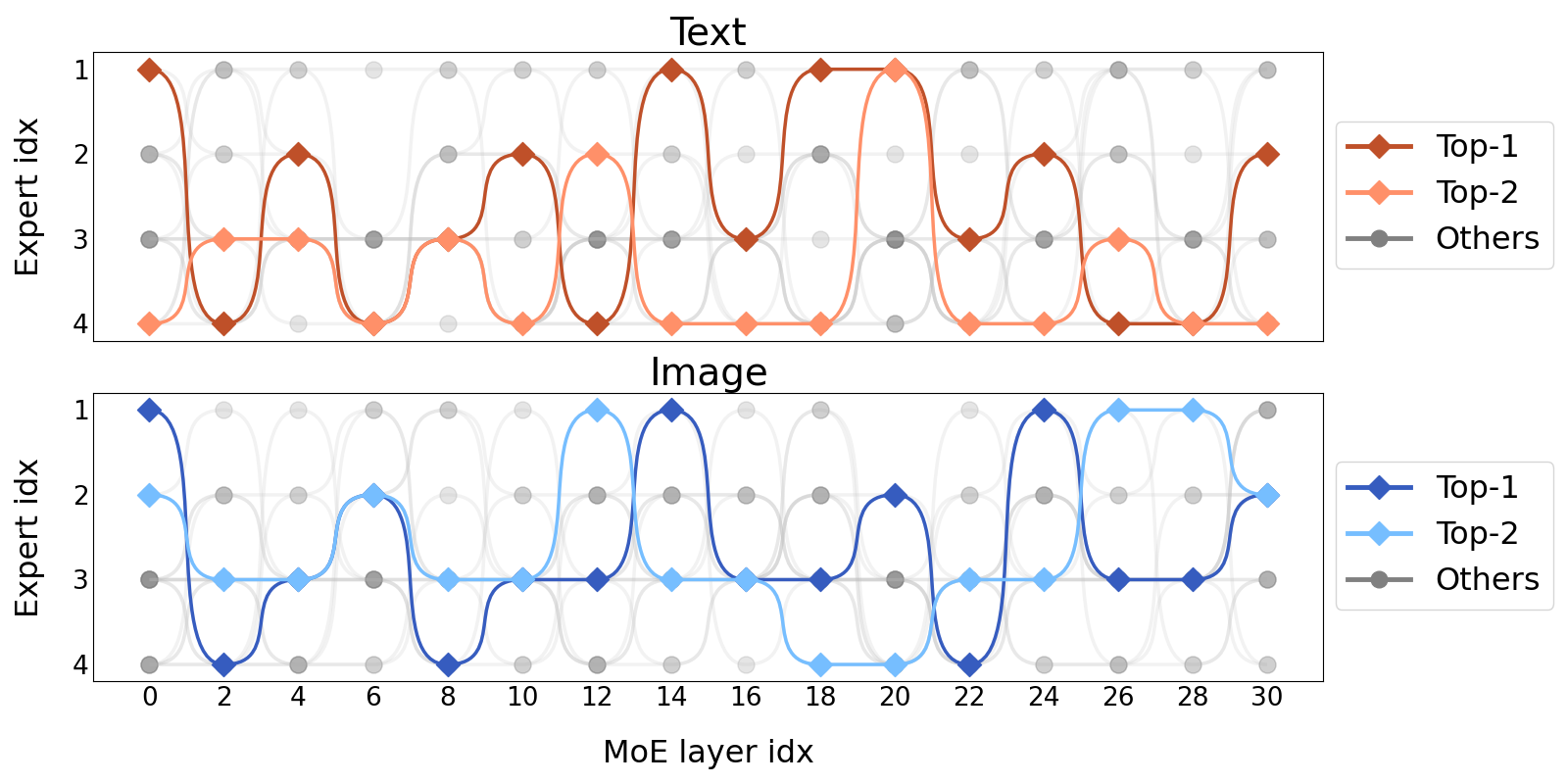}}
\resizebox{\linewidth}{!}{
\includegraphics[width=0.95\linewidth]{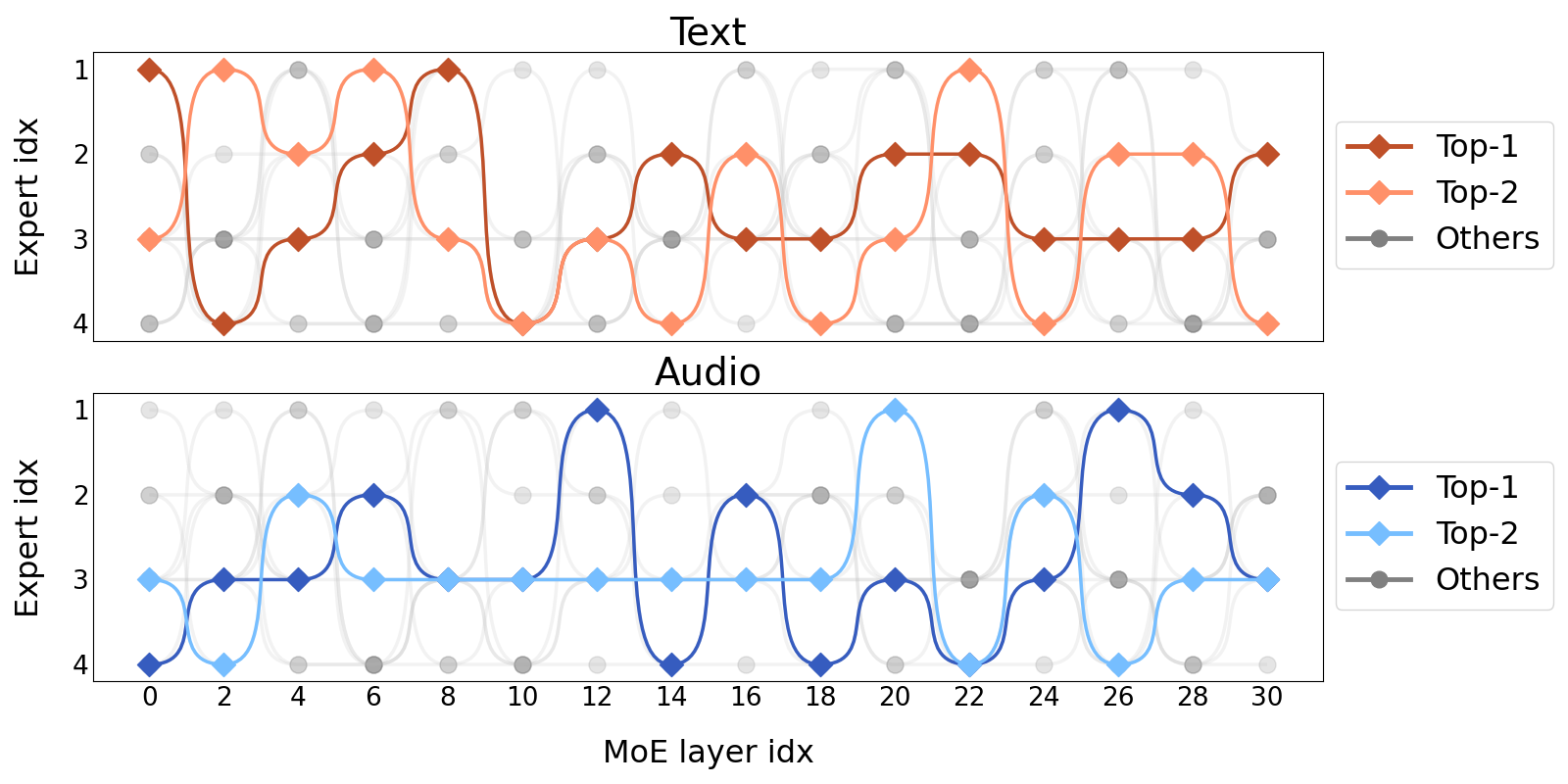}}
\resizebox{\linewidth}{!}{
\includegraphics[width=0.95\linewidth]{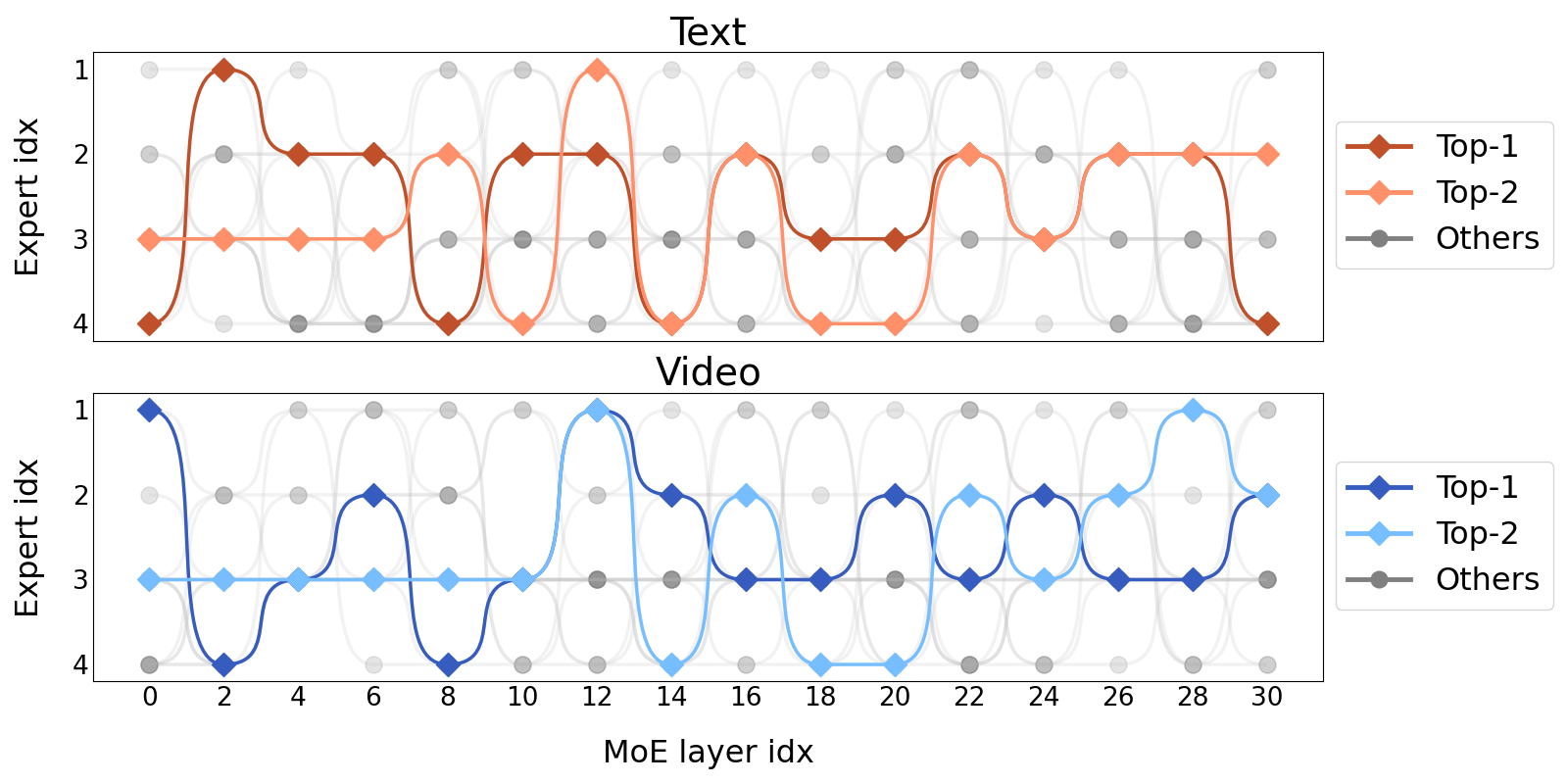}}
\resizebox{\linewidth}{!}{
\includegraphics[width=0.95\linewidth]{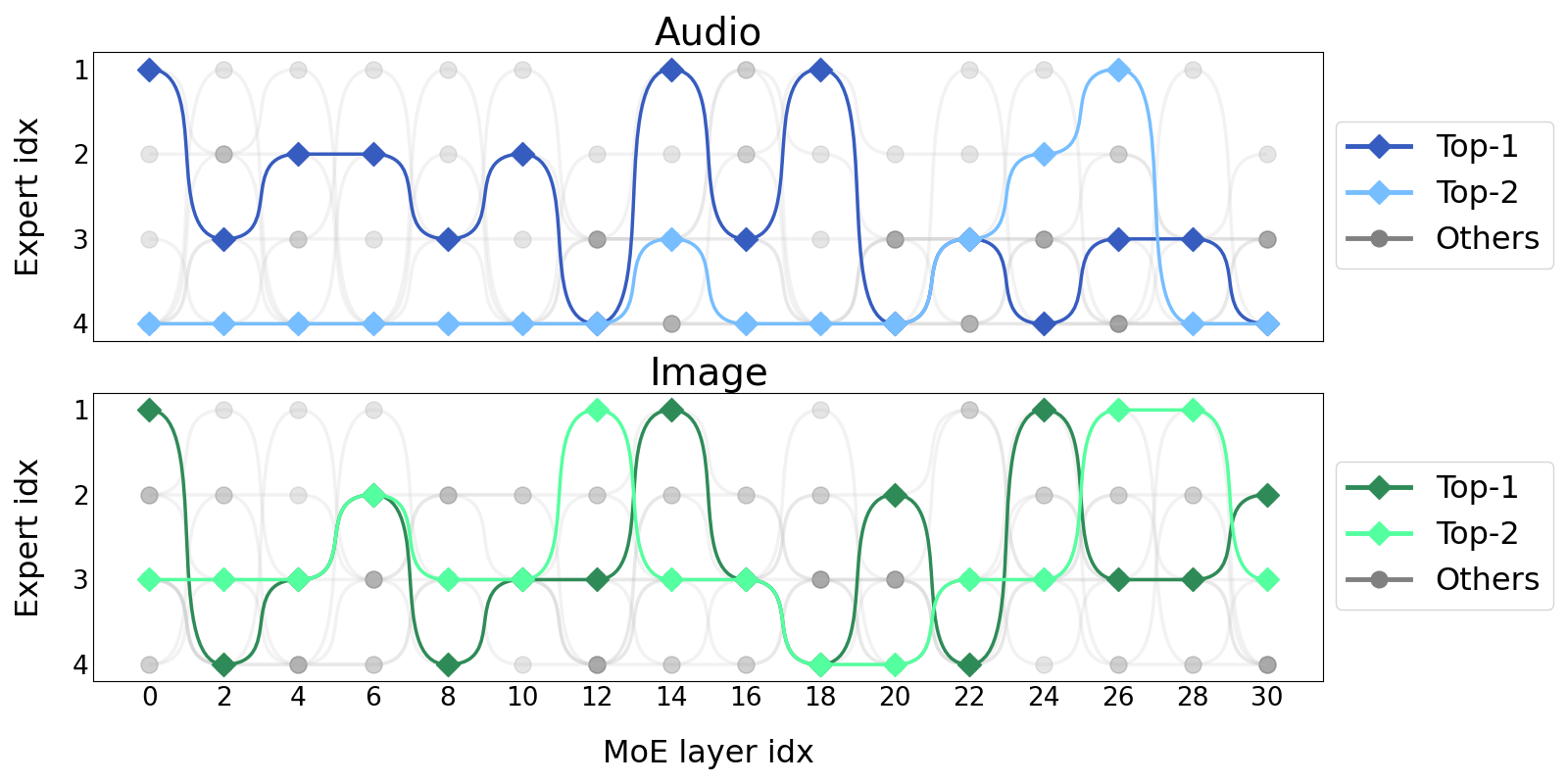}}
\resizebox{\linewidth}{!}{
\includegraphics[width=0.95\linewidth]{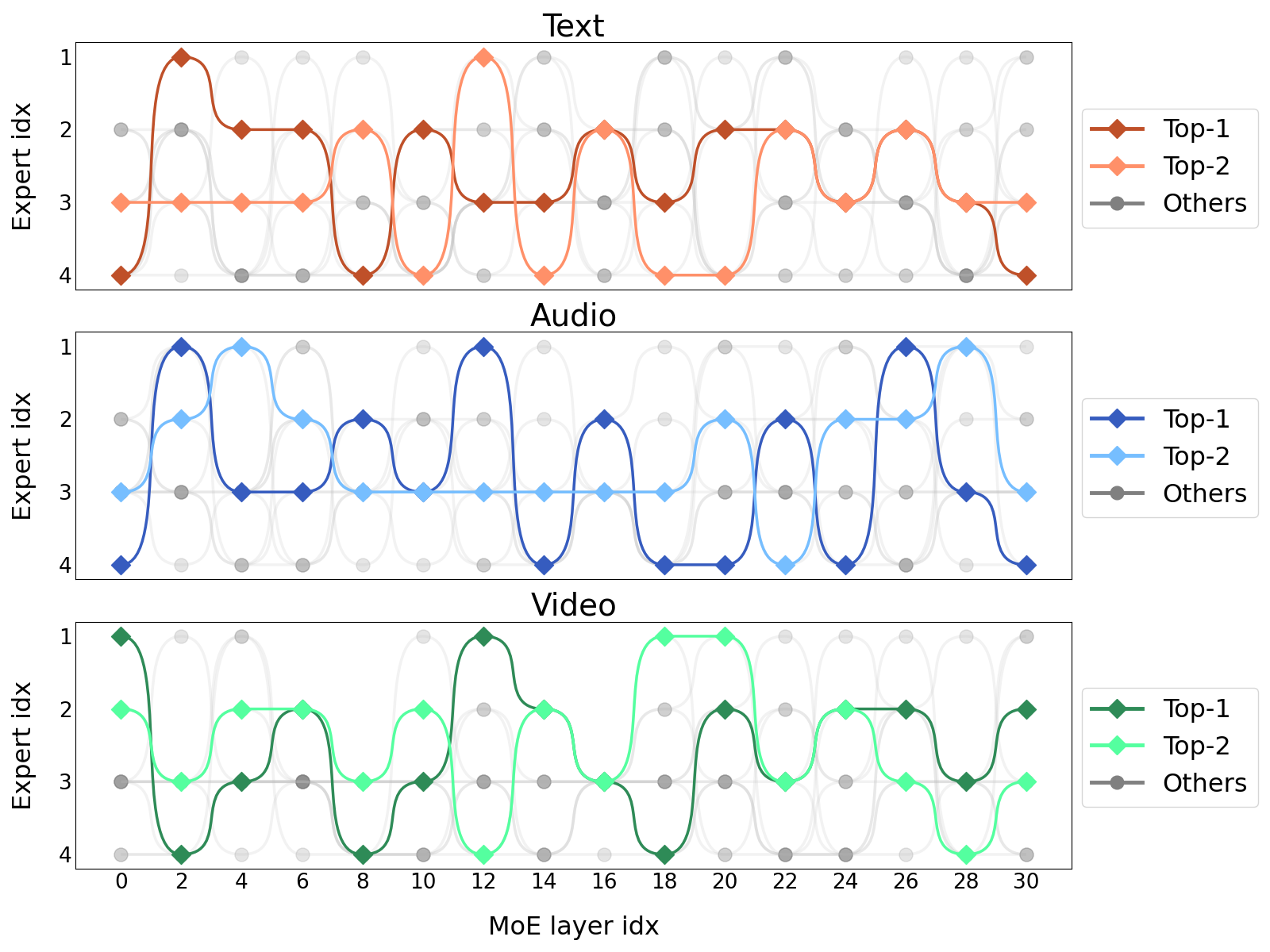}}
 \caption{Visualization of activated pathways on \textbf{Pure-MoE-Task1}.}
\label{fig:activate_ways_pure}
\end{figure}

\begin{figure}[p]
\centering
\resizebox{\linewidth}{!}{
\includegraphics[width=0.95\linewidth]{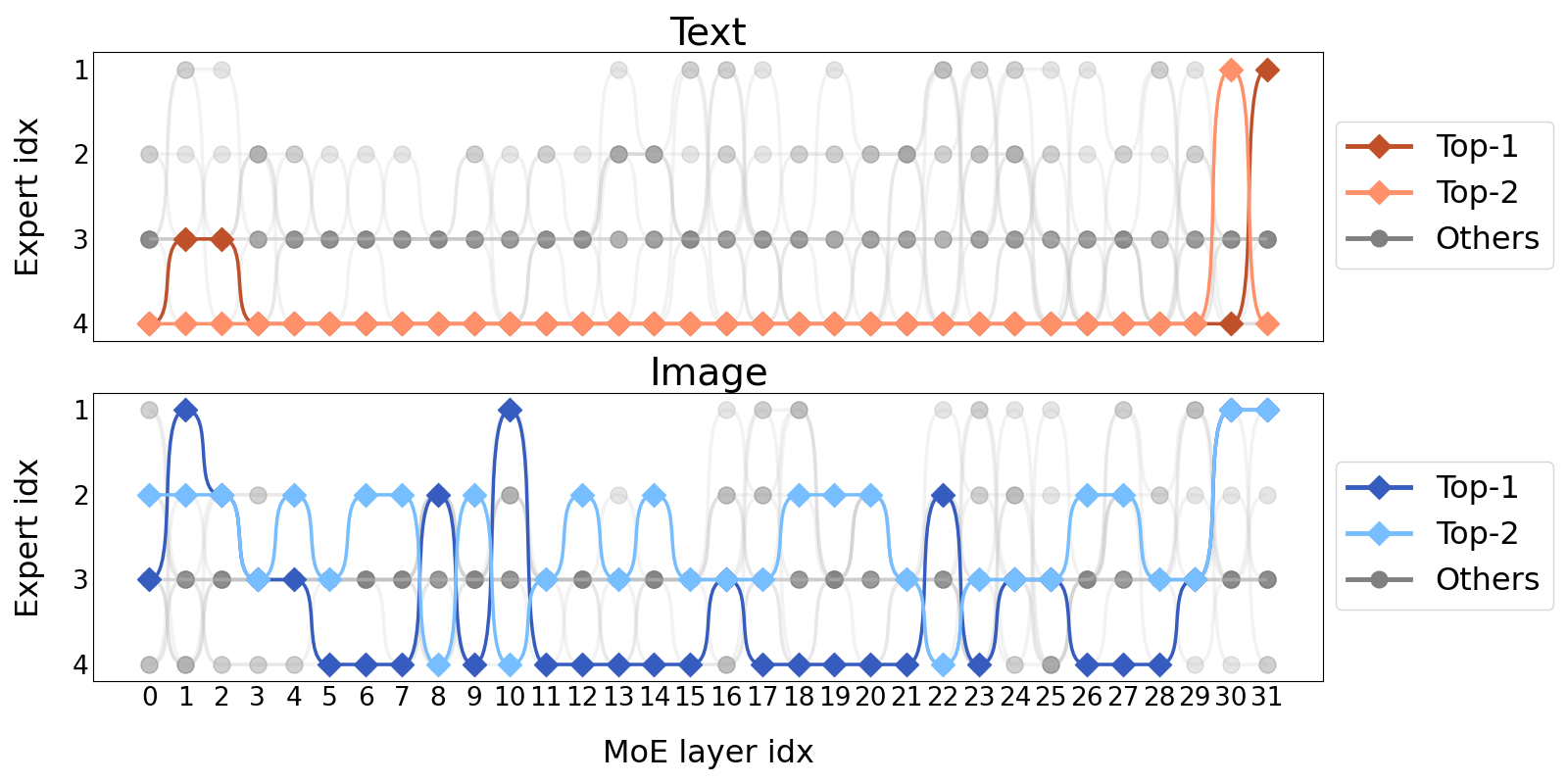}}
\resizebox{\linewidth}{!}{
\includegraphics[width=0.95\linewidth]{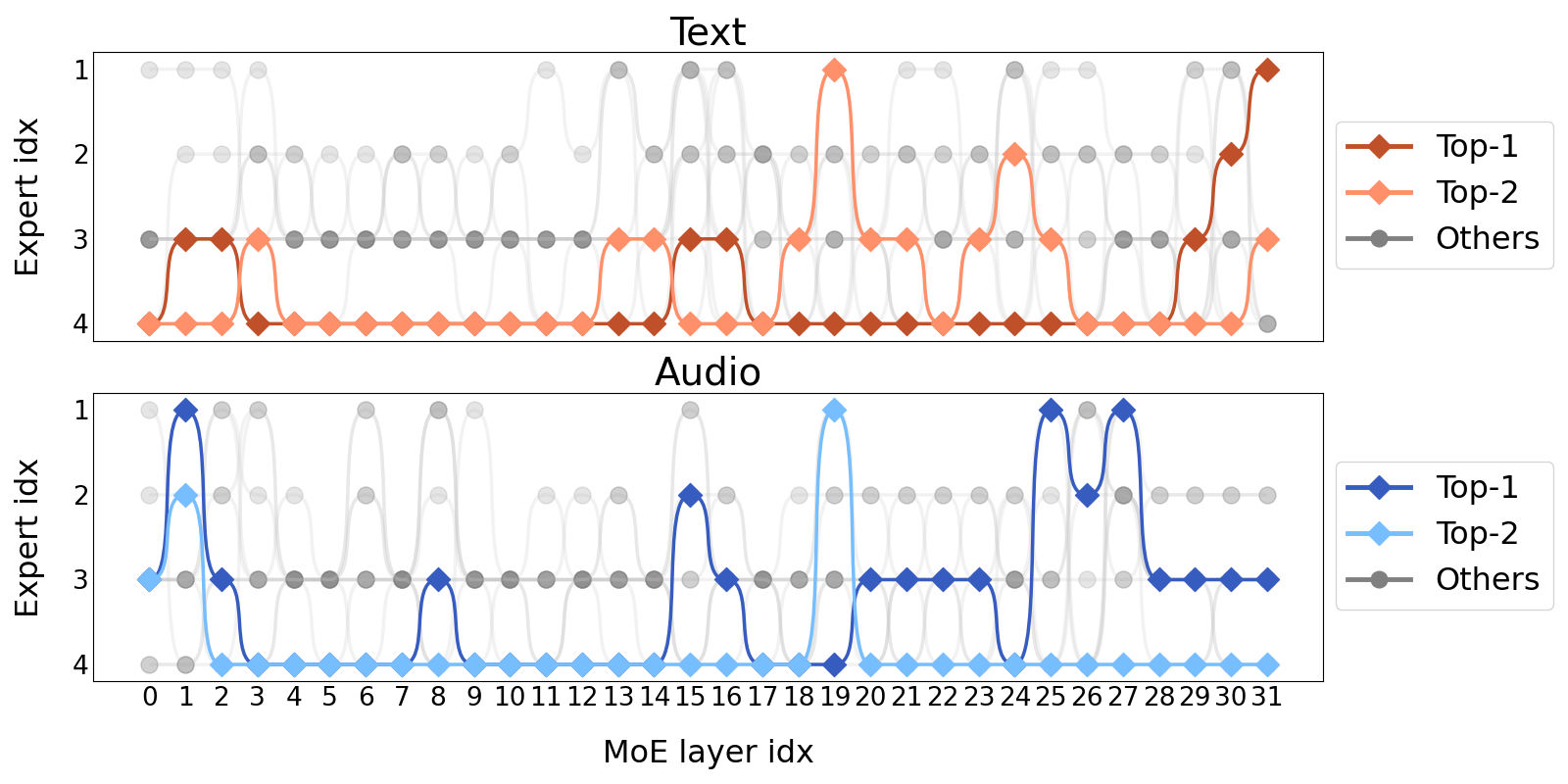}}
\resizebox{\linewidth}{!}{
\includegraphics[width=0.95\linewidth]{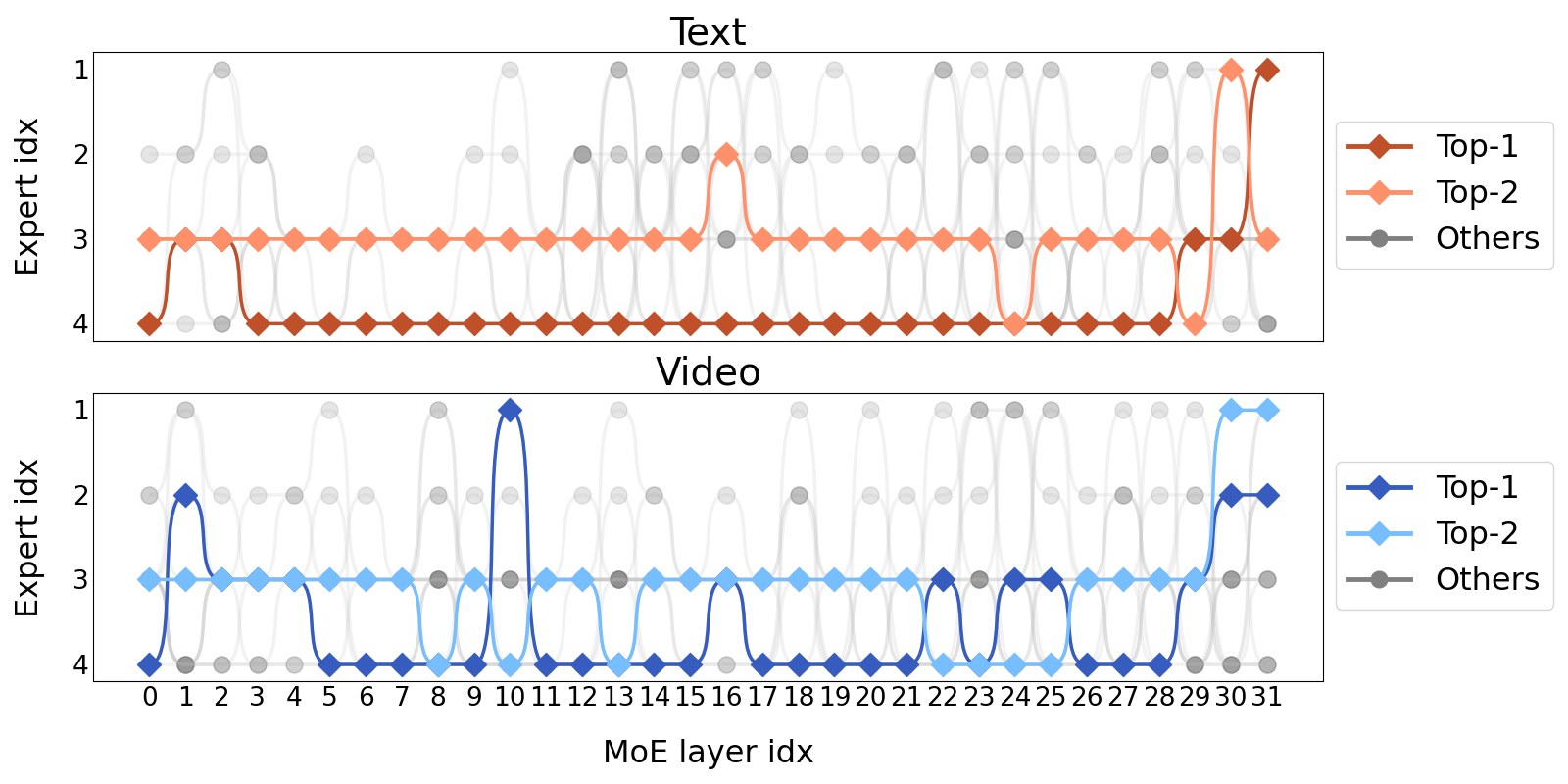}}
\resizebox{\linewidth}{!}{
\includegraphics[width=0.95\linewidth]{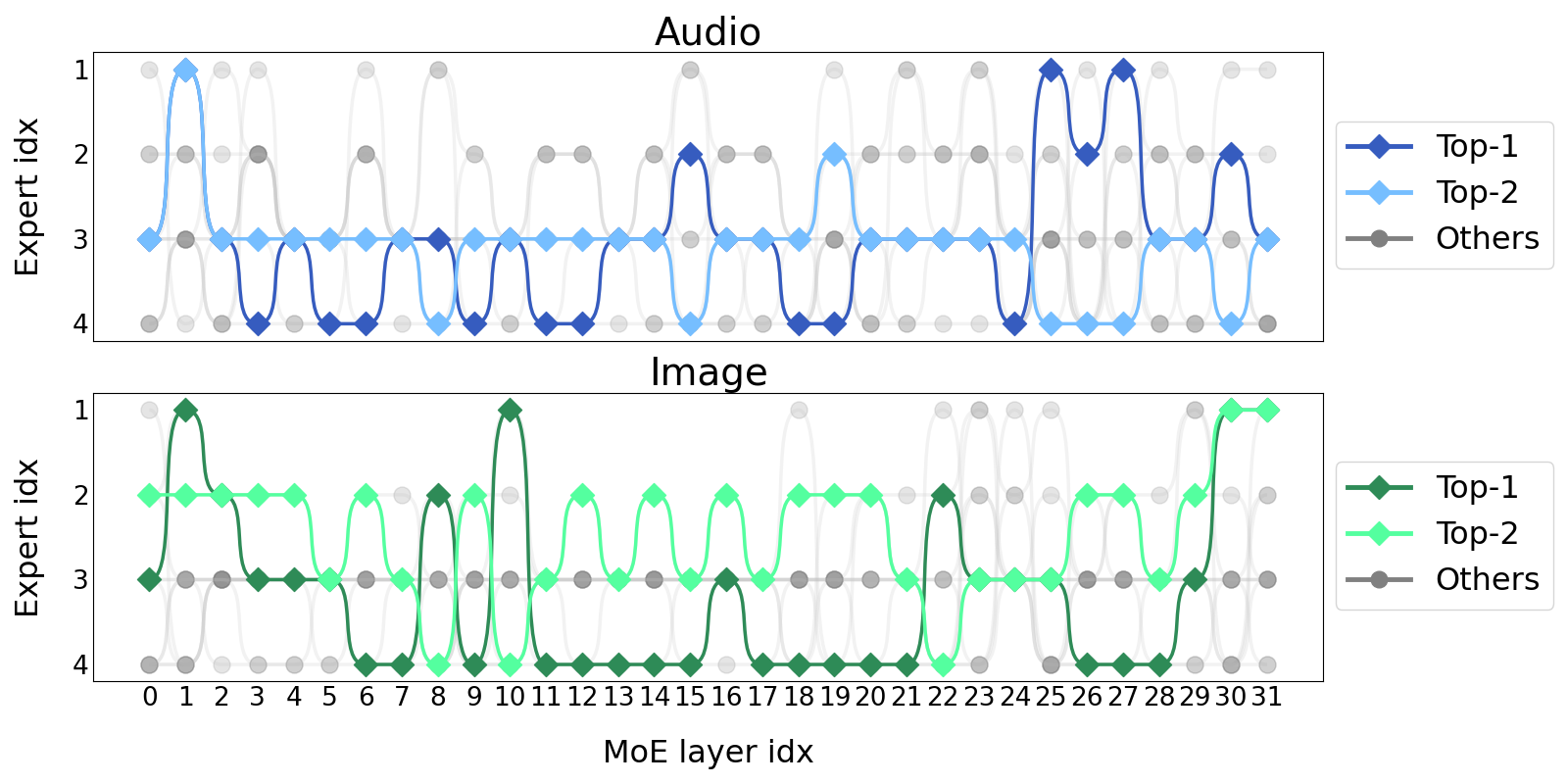}}
\resizebox{\linewidth}{!}{
\includegraphics[width=0.95\linewidth]{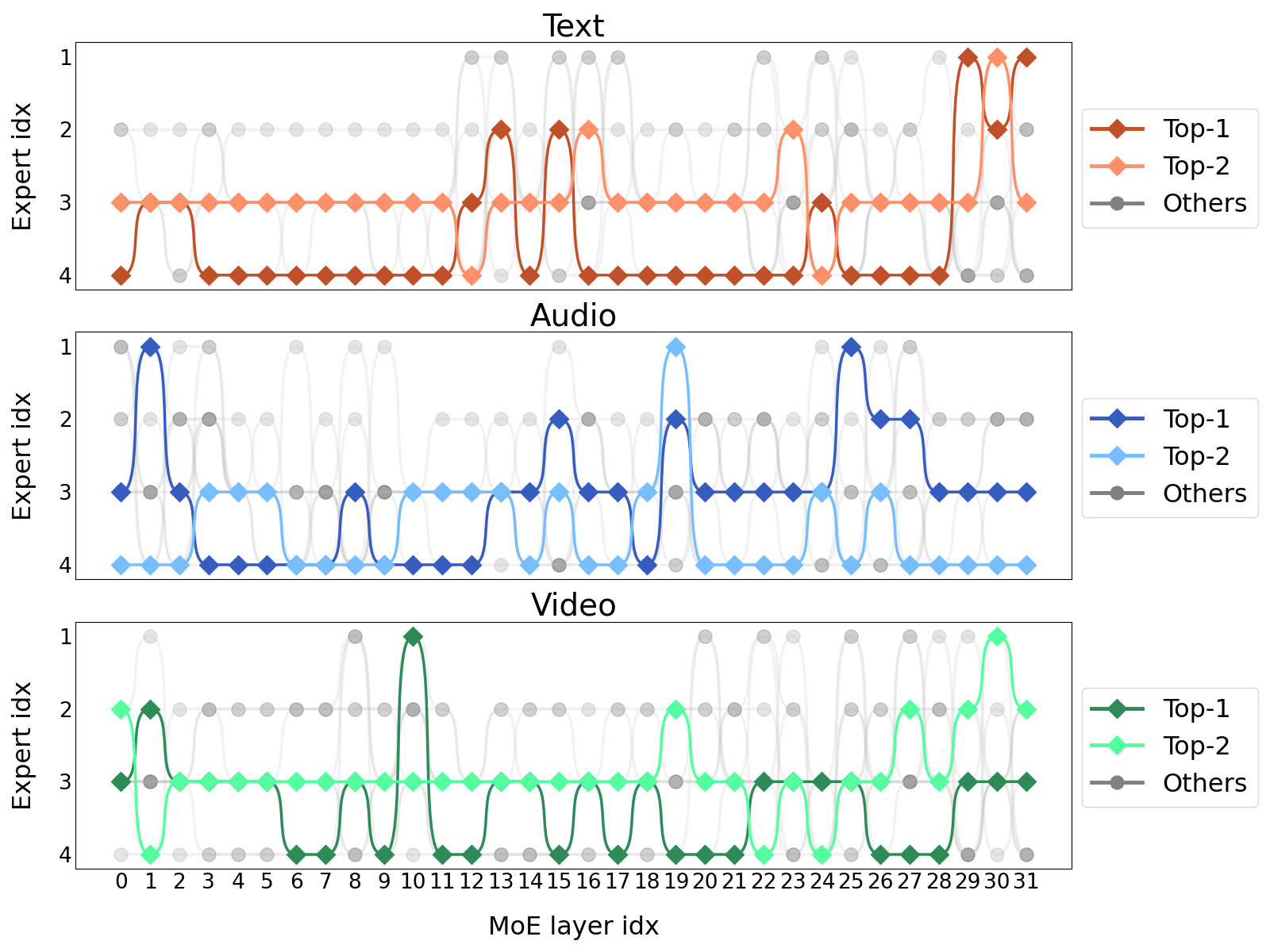}}
 \caption{Visualization of activated pathways on \textbf{MoE-Task2}.}
\label{fig:activate_ways_speech}
\end{figure}

\textbf{Token Pathways}. 
In Figure~\ref{fig:activate_ways_pure} and Figure ~\ref{fig:activate_ways_speech}, we track the paths of each token for Pure-MoE-Task1 and MoE-Task2, respectively. In general, the overall trends of the token paths align with the analysis in the above routing distributions. The paths of Pure-MoE-Task1 appear more disorderly and diverse, which is attributed to a more balanced expert assignment. On the other hand, MoE-Task2 shows its unique preference for experts.

\begin{figure}[p]
\centering
\includegraphics[width=1.0\linewidth]{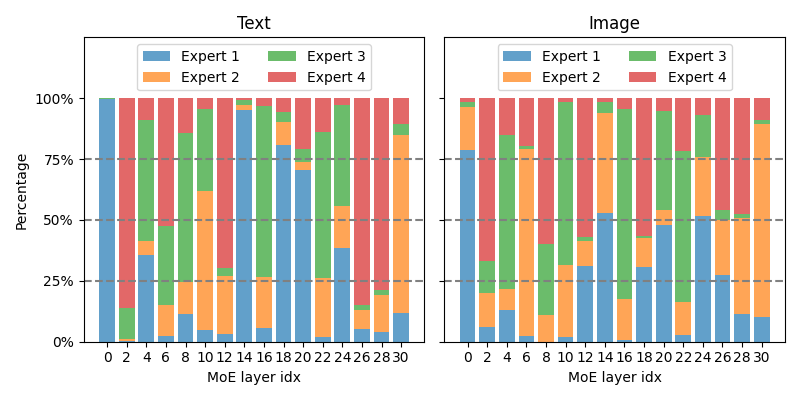}
\vspace{5pt}
\includegraphics[width=1.0\linewidth]{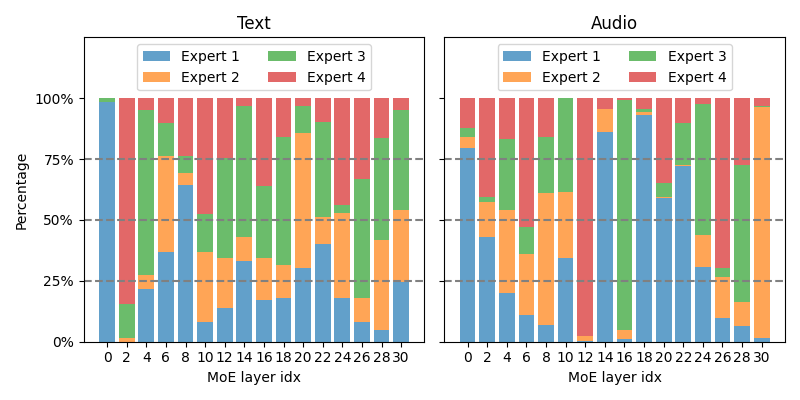}
\vspace{5pt}
\includegraphics[width=1.0\linewidth]{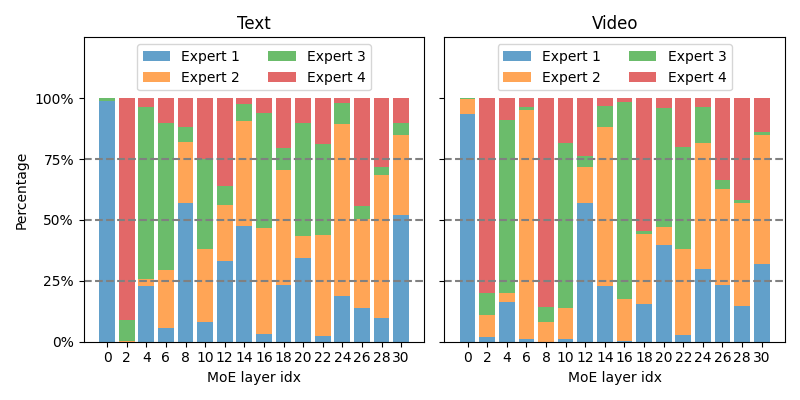}
\vspace{5pt}
\includegraphics[width=1.0\linewidth]{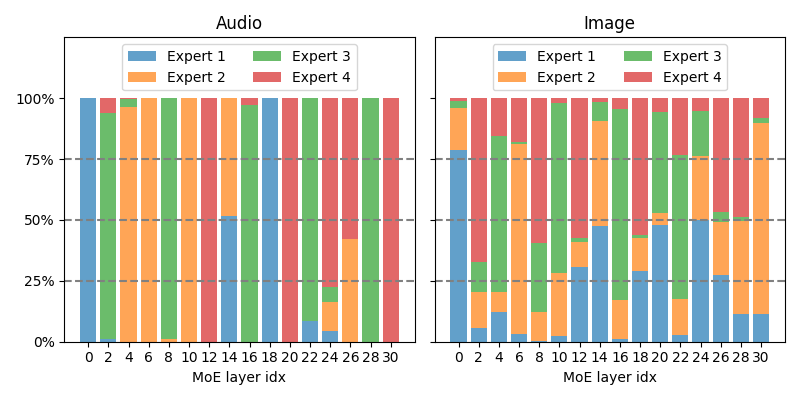}
\vspace{5pt}
\includegraphics[width=1.0\linewidth]{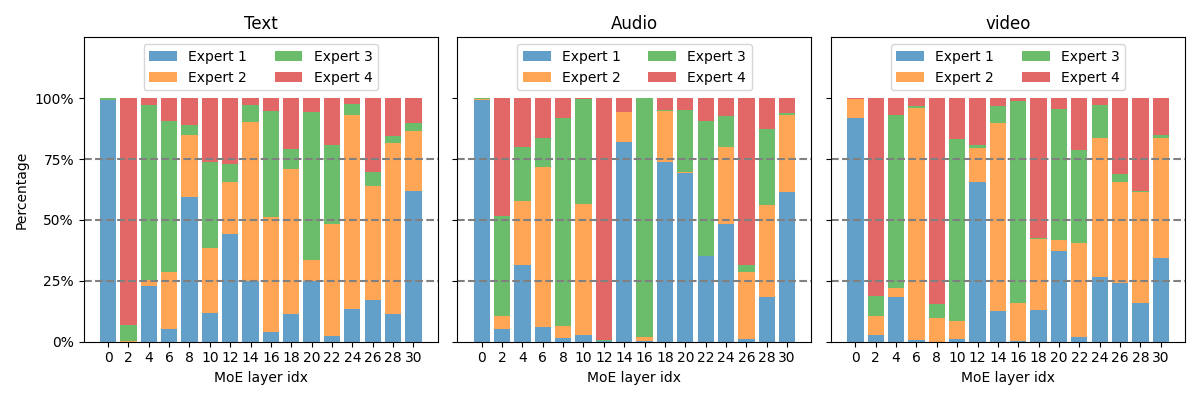}
 \caption{Distribution of modalities across different experts on \textbf{Pure-MoE-Task1}.}
\label{fig:modalities_distribution_pure}
\end{figure}

\begin{figure}[p]
\centering
\includegraphics[width=1.0\linewidth]{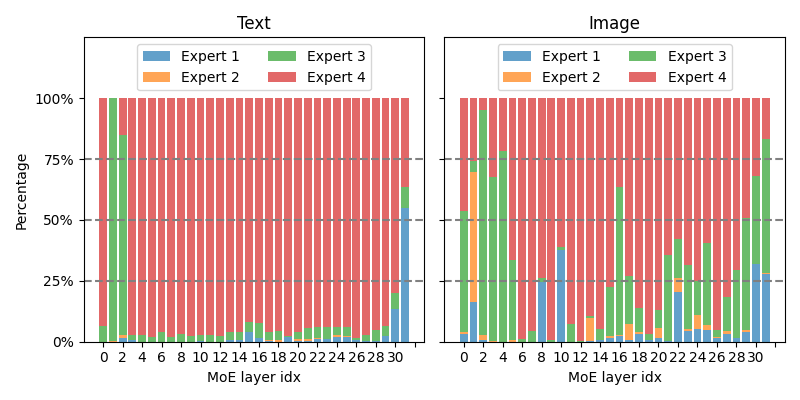}
\vspace{5pt}
\includegraphics[width=1.0\linewidth]{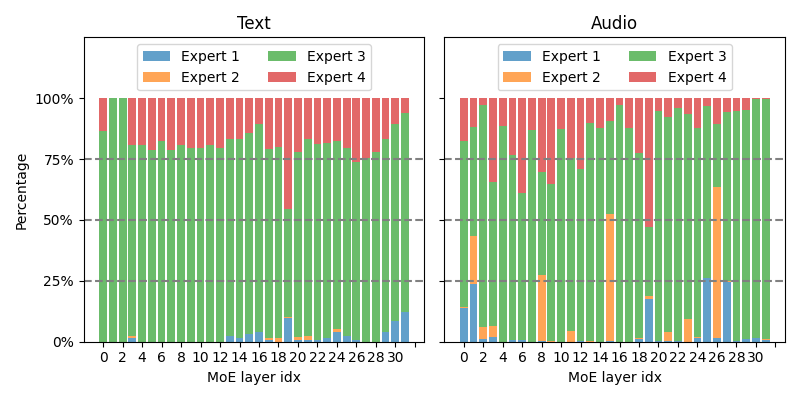}
\vspace{5pt}
\includegraphics[width=1.0\linewidth]{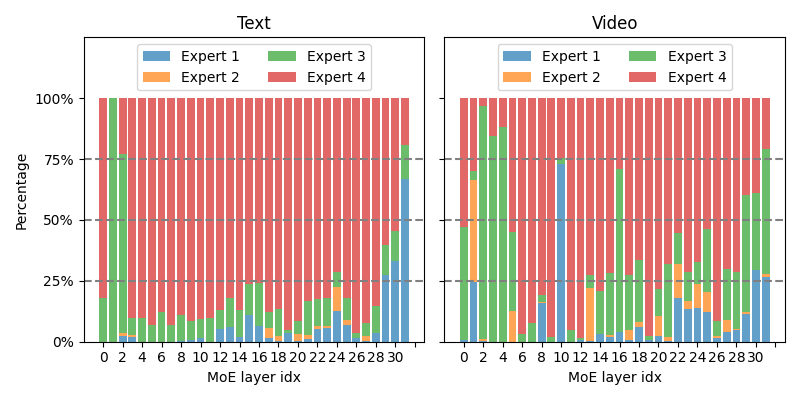}
\vspace{5pt}
\includegraphics[width=1.0\linewidth]{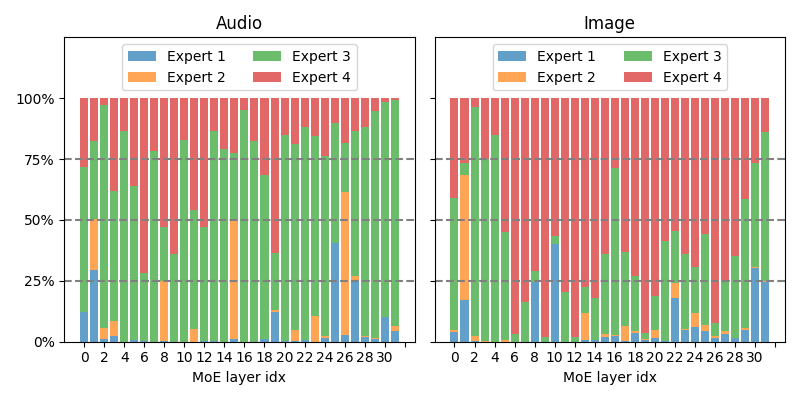}
\vspace{5pt}
\includegraphics[width=1.0\linewidth]{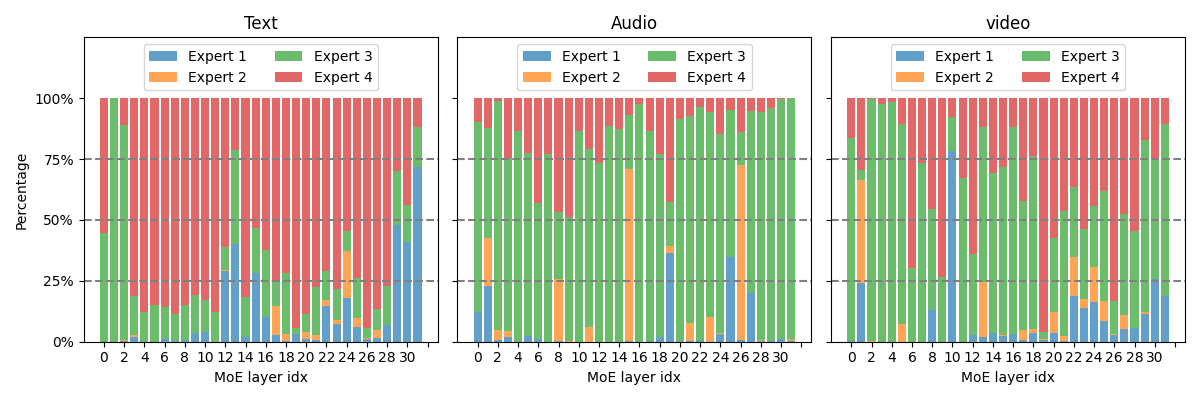}
 \caption{Distribution of modalities across different experts on \textbf{MoE-Task2}.}
\label{fig:modalities_distribution_speech}
\end{figure}

\end{document}